# UNIVERSITY OF SOUTHAMPTON

## Faculty of Physical Science and Engineering

## School of Electronics and Computer Science

Ontology Learning Using Formal Concept Analysis and WordNet

By

BRYAR A. HASSAN (bah1g12)

6$^{th}$ of September 2013

A dissertation submitted in partial fulfilment of the degree of

MSc Software Engineering

By examination and dissertation

**Supervisor:** Dr Nicholas Gibbins

**Examiner:** Dr Su White

# Abstract

A manual construction of ontology is usually a time-consuming and resource demanding process and it involves some domain experts as well. Therefore, it would be beneficial to support a part of this process in order to be automated or semi-automated. In that sense, this project and dissertation aim to propose a framework for learning concept hierarchies from free texts using a Formal Concept Analysis and WordNet. The overall process includes a set of steps. Firstly, the document is Part-Of-Speech tagged, and then parsed to yield a parse tree for each sentence. Next, dependencies in the form of verb/noun are extracted from the parse trees. Afterwards, the pairs are lemmatized and the word pairs should be pruned and filtered, and then the formal context is constructed based on the word pairs. However, there may be some erroneous and uninteresting pairs in the formal context because the output of the parser can be erroneous and no all the derived pairs are interesting and the size of formal context might be substantially large due to constructing the formal context from a large free text corpus. Meanwhile, deriving the concept lattice from the formal context might be a time consuming processing and it depends on the size and complexity of the formal context data. Hence, reducing the size of formal context might result in eliminating the erroneous and uninterested pairs and consuming less time for deriving the concept lattice. On that basis, two techniques are investigated by conducting an experiment, which are WordNet-based technique and Frequency-based technique. Finally, lattice of formal concepts is computed and results in a traditional concept hierarchy. The results are evaluated by comparing the reduced concept lattice with the original one. Even though the system has a few limitations and inconsistencies of its components that may hinder to draw a logical conclusion, the following results still indicate that suggesting concept hierarchies in this project and dissertation are relatively potential. Firstly, there are similarities between the concepts of the reduced concept lattice and the original one. Secondly, the size of formal context can be reduced differently by using any of linguistic resources or/and statistical approaches. Lastly, the impact of using WordNet-based technique and Frequency-based technique are different from each other in terms of reducing the size of formal context as well as the order of applying these techniques is investigated in reducing the size of formal context in an efficient way. Thus, this project and dissertation could provide a baseline of basic engineering principles for deriving concept hierarchies in order to take steps ahead towards the goal of finding more principles and answering further research questions in the future.

# Acknowledgements

This project and dissertation would not have been created if it were not for the sincere encouragements and support of my supervisor, Dr Nicholas Gibbins. His supports and recommendations have tremendously assisted me while the project was on progress. Thus, I am truly indebted and thankful to him for his inspiration and guidance through my project.

I would also like to thank my examiner Dr Su White greatly for his steadfast patience in revising my work.

I should also take the opportunity to thank Kurdistan Regional Government (KRG) who offered me this precious scholarship, which I regard as the most precious gift I have ever been given in my life.

# Table of Contents

# 1. Chapter One: Introduction

The Semantic Web is a web of data that can be processed directly or indirectly by machines [1]. The Semantic Web as an extension of the current Web can bring structure to the content of the Web and thereby support automated services based on the description of semantics. In the meantime, the Semantic Web relies on formal ontologies that structure underlying data to enable comprehensive and transportable machine understanding [2]. Ontologies are widely used in information systems and are an essential part of the Semantic Web and also the proliferation of ontologies requires that the engineering of ontologies be completed easily and quickly to result in the success of the Semantic Web. However, manual building of ontologies still remains a tedious, cumbersome task and the main problem in building ontologies manually is the bottleneck of knowledge acquisition and time-consuming construction, difficulty of maintenance, and integration of various ontologies for various applications and domains. To resolve this problem, ontology learning is a solution to the bottleneck of knowledge acquisition and time-consuming construction of ontologies. Ontology learning (ontology extraction, ontology generation, or ontology acquisition) is a subtask of information extraction and its goal is to semi-automatically or automatically extract relevant concepts and relations from a given corpus or other kinds of data sets to form ontology. In that context, this project and dissertation basically aim to propose an approach for the automatic induction of concept hierarchies from free text using Formal Concept Analysis and WordNet. This process includes a set of steps. One of the most essential step of this process is how the phrase dependencies or word pairs are extracted, particularly the needed and interested pairs because not all of the pairs are correct or interesting. To eliminate the erroneous and uninterested pairs, and reduce the size of formal context to consume less time for deriving the concept lattice, we will investigate two techniques for eliminating the erroneous and uninteresting pairs and their terms by comparing the pairs with the WordNet, and then the pairs are weighted based on some statistical measure and thereby only the pairs over a certain threshold are transformed into a formal context to which Formal Concept Analysis is applied.

## 1.1 Research Questions

The main problems in building ontologies manually are the bottleneck of knowledge acquisition and time-consuming construction, difficulty of maintenance, and integration of various ontologies for various applications and domains. In addition, manual building of ontologies still remains a tedious, cumbersome task. Hence, ontology learning is a solution to the bottleneck of knowledge acquisition and time-consuming construction of ontologies. Recently, a lot of research works, particularly about building concepts, concept hierarchies, and relationships have been conducted to design appropriate methods and frameworks for ontology learning, but some of them suffer from some common shortcomings that prevent wide production and usage of ontology that prevent wide production and usage of ontology. Therefore, a question needs to ask is how could a framework be proposed for learning concept hierarchies from domain-specific text by using Formal Concept Analysis and WordNet. This process for automating concept hierarchies includes Natural Language Processing parser to produce a parse for each sentence, and then syntactic dependencies in the form of verb–noun are extracted to construct formal context. However, there may be still some erroneous and uninteresting pairs in the formal context because the output of the parser can be erroneous and no all the derived pairs are interesting and the size of formal context might be substantially large. In the meantime, deriving the concept lattice from the formal context might be a time consuming processing and it depends on the size and complexity of the formal context data. Hence, this experiment might result in eliminating uninterested and erroneous pairs and reducing the size of formal context to consume less time for deriving the concept lattice. In turn, the following questions need to be asked:

- Does reducing the size of formal context require linguistic resources, such as WordNet or/and statistical approaches?
- Are linguistic resources more effective than statistical approaches in reducing the size of formal context or vice versa?
- If both linguistic resources and statistical approaches need to reduce the size of formal context? In which order should they apply on the formal context to be more effective?
- Can the size of formal context be reduced without affecting the quality of the result?

## 1.2 Objectives

To answer these above research questions and to deal with the issues of formal context, this project and dissertation aims the following below points:

1. It is to propose a framework for learning concept hierarchies from domain-specific texts by using Formal Concept Analysis and WordNet.

2. It is to propose a novel approach for conducting an experiment by using and investigating two distinct techniques which are WordNet-based and Frequency-based techniques to eliminate the erroneous and uninteresting pairs in the formal context and reduce the size of formal context to result in less time consuming when concept lattice derives from it.

3. The experiment of this project and dissertation also aims to investigate whether a linguistic resource-based technique, such as WordNet or/and statistical-based technique needed for the purpose of reduction the size of formal context?

4. The experiment is to investigate the impact of using each and both techniques for reducing the size of formal context.

5. Conducting an experiment is to investigate how the quality of concept lattice results is before and after conducting the experiment?

6. It is also to evaluate the experiment by comparing the resulting concept lattice with the concept lattice before applying WordNet-based and Frequency-based techniques.

7. It is to evaluate the usability of Stanford Natural Language Processing as a framework and the Concept Explorer as open source software for implementing the basic ideas needed for the research and study of Formal Concept Analysis and lattice compaction.

### 1.3 Scoping Assumptions

Based on the problem description and objectives, the following assumptions are considered:

1. Necessary tools, such as the Concept Explorer that the Formal Concept Analysis can be built upon should be provided.

2. Necessary tools and packages, such as Stanford Natural Language Processing framework as a set of natural language analysis tool that raw English language text can be analysed respectively should be provided.

3. Both the Concept Explorer and Stanford Natural Language Processing framework are written by Java. This project and dissertation could be integrated with Stanford Natural Language Processing framework as well as the Concept Explorer. Accordingly, Java programming language should be used in this project and dissertation.

4. Extensible Mark-up Language (XML) as a mark-up language and CEX file as an XML-based file format is used as a standard format to link the components of this project and dissertation.

### 1.4 Thesis Structure

This project and dissertation are divided into six chapters. As it has presented, the first chapter is mainly about an overview of ontology learning, research questions, objectives, and scoping assumptions.

In chapter two, a theory, background, and the state of art of ontology and ontology learning are introduced and the relevant techniques of natural language processing and information extraction are presented. First, a detail is given regarding ontologies, and then automatic ontology generation lifecycle is explained. Second, ontology learning including ontology learning input sources and methodologies and ontology learning approaches is presented. Last but not least, an overview is given in regard with information extraction and text mining that composes of Natural Language Processing, Stanford Natural Language Processing framework, and WordNet. Last, the basic idea of Formal Concept Analysis and lattice compaction for the purpose of research and study of this project and dissertation is introduced with giving details about formal contexts and concepts, concept lattice, first example of Formal Concept Analysis, and the Concept Explorer as subsections.

Chapter three elaborates the design and overall process of the project and dissertation. This chapter describes design phase of the overall concept hierarchy process and a method for discovering concepts and concept hierarchies in a free text by using Formal Concept Analysis and WordNet. At the first glance, it begins with explaining the overall processing about how to learn concepts and concept hierarchies from free text by ten steps. Then, Formal Concept Analysis as one of the ontology learning techniques is defined and detailed by focusing on formal contexts, concepts and concept lattices, and giving an example about applying Formal Concept Analysis on free text. Finally, WordNet-based and Frequency-based techniques as two distinct techniques that can be applied on formal context are demonstrated and detailed via examples

Chapter four further describes the implementation and testing of the baseline design from chapter three. At the first glance, an implementation of this project is introduced in chapter four. First and foremost, a novel approach is proposed to introduce a framework architecture for the overall process of this project. Second, the modules of this framework are implemented in order. It starts from Natural Language Processing parser, and then the word pairs are extracted and thereby the formal context can be constructed. Correspondingly, an experiment is conducted by using WordNet-based and frequency-based techniques that might result in eliminating uninterested and erroneous pairs and reducing the size of formal context to consume less time for deriving the concept lattice. Afterwards, miscellaneous module as a method of communication among the other modules is detailed. Next, formal concept and lattice compaction are applied by using the Concept Explorer software. Final, an example is applied to clarify the implementation steps with the experiment in detail. Later in this chapter, the project is tested by using JUnit testing and Big Bang integrated testing.

Chapter five is basically about evaluating the used frameworks and tools, and the resulted concept lattices as well as discussing the results and achievements from the described method of evaluation for the concept lattices found by the system. It begins with evaluating the used frameworks and tools in this project and dissertation including Stanford Natural Language Processing frame, the Concept Explorer, and WordNet. Afterwards, the methods of evaluating concept lattices are presented and discussed, and then the evaluation strategy that composes of a case study and its specification are presented. At the end of this chapter, the results, achievements, and limitations from the described method of evaluation for the concept lattices found by the system are discussed.

Chapter six presents the concluding remarks for the work done and discusses possible directions for further work in the future.

The structure of the appendices is as follows:

- Appendix A: The Project Work Plan and Milestones
- Appendix B: Acronyms and Abbreviations
- Appendix C: Digital Appendix
- Appendix D: Case Study
- Appendix E: The Used Tools and Frameworks
- Appendix F: Additional Results
- Appendix G: Bing Bang Integrated Testing

# 2. Chapter Two: Theory and Background

Over the last decades, substantial works have been done in the field of Semantic Web technologies. One of the on-going research areas in the Semantic Web filed is ontology engineering and constructing it automatically. Ontology engineering is described as a time consuming and resource demanding processing with involving some domain experts to construct an ontology manually [3]. Therefore, it would be beneficial to partially or completely support this process for constructing ontology semi-automatically or automatically. Ontology learning is defined as a subtask of information extraction to extract relevant concepts and relations semi-automatically from a given corpus or other types of data sources to form an ontology. In this chapter, a background and theory about the field of ontology learning and its related areas are introduced and the relevant techniques of natural language processing and information extraction are presented. First, a detail is given regarding ontologies, and then automatic ontology generation lifecycle is explained. Second, ontology learning is overviewed. Last but not least, an overview is given in regard with information extraction and text mining. Last, the basic idea of Formal Concept Analysis and lattice compaction for the purpose of research and study of this project and dissertation is introduced.

## 2.1 Ontologies

Ontology is a body of formally represented knowledge and a specification of a conceptualisation as a set of concepts within a domain. In other words, it could represent objects, concepts and other entities that exist in an area of interest and the relationship among them in order to model domain support reasoning about concepts. In philosophical perspective, on the other hand, ontology means the representation of what exist in the world. There are two common reasons for developing ontology information systems [4]. The first reason is to establish a shared understanding of a domain and facilitate sharing information accordingly. The second reason is to enable the construction of intelligent or semantic applications. The latter reason can be seen in the context of the Semantic Web where is necessarily used for representing domain knowledge that allows machine to perform reasoning tasks on documents. Further, Ontology consists of concepts, taxonomic relationships that define a concept hierarchy, and non-taxonomic relationships between them, and concept instances and assertions between them. Ontology should be machine readable and formal. In that sense, ontology could provide a common vocabulary between different applications. This knowledge representation structure often composed of a set of frame-based classes that hierarchically organised for describe a domain and provides a skeleton of the knowledge base. The represented concepts for describing a dynamic reality are used by the knowledge base and it changes in accordance with the changes in the reality. Nevertheless, if there are changes in only the described domain, the ontology changes. Formally, ontology can be defined as a tuple of ontology concepts, a set of taxonomic and non-taxonomic relationships, a set of axioms, and instances of concepts and relationships [5].

## 2.2 Automatic Ontology Generation Lifecycle

Ontology engineering mainly studies the methodologies and methods for building ontologies. There are several methodologies for building ontologies. OTK, METHODOLOGY, and DILIGENT are examples of ontology methodology. All of these methodologies target on ontology engineer rather than machines, but, the dynamic machine interpretation of input and output data of applications is more needed. Hypothetically, ontology engineering methodologies can be automatically processed by defining automatic ontology generation lifecycle.

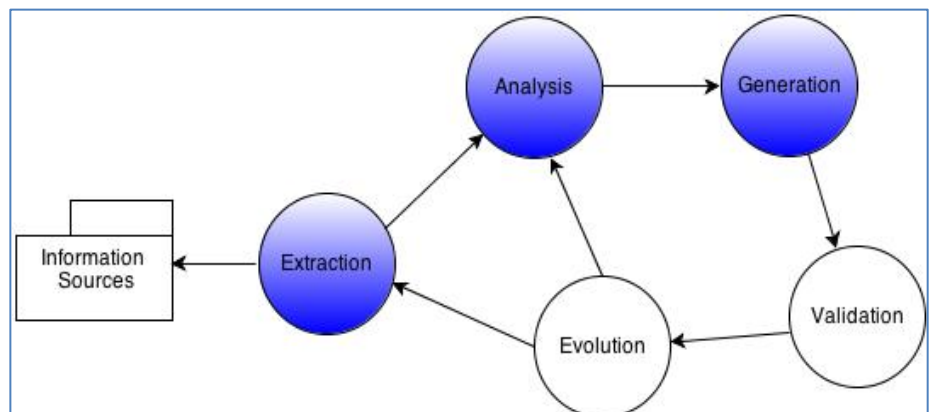

Figure 2.1: Automatic ontology generation process

Automatic ontology generation lifecycle is defined as a process composed of five main steps that can be considered necessary to achieve automatic ontology generation. According to [6], the main task of this process is represented by these steps for building ontologies starting from an existing corpus source. The process is presented in Figure 2.1 and the details of the five steps are:

**1. Extraction:** This provides the needed information acquisition to generate the ontology (concepts, attributes, relationships, and axioms) by starting from an existing corpus source. Input resources of ontology learning can be from structured data, Semi-structured data, or unstructured data [7]. The details of these input types are given in the next section.

**2. Analysis:** This stage focuses on alignment and/or the matching of retrieved information of two or more existing ontologies. This step needs: the used techniques in the first step; a semantic analysis to detect homonyms, synonyms, and other relations of this type; an analysis of concepts' structure to find hierarchical relationships and identify common attributes; techniques based upon reasoners in order to detect inconsistencies and induced relations.

**3. Generation:** This step is about the ontology merging, and the formalisation of the meta-model used by the tool in a more general formalism which is interpretable by other applications, such as RDF(S) and OWL.

**4. Validation:** All the three previous steps might introduce incorrect concepts and relationships. Thus, an automated validation phase of the result is required. On the other hand, at the end of each previous step, a validation task can be introduced. This step is usually done by hand, but validation can be automated in some cases.

**5. Evolution:** Ontology is a dynamic description of a domain and thereby it may need some changes in quality and number with the evolution of the applications, such as the number of concepts, properties, and relationships. This operation is taken as an addition of new requirements and it could be followed by a new step of information extraction

The main task of this project and dissertation can be represented by the first steps of automatic generation lifecycle. These three steps start from building ontologies from an existing corpus source.

**2.3 Ontology Learning**

Ontology learning refers to the semi-automatic or automatic support for constructing an ontology when the semi-automatic and automatic support for the instantiation of a given ontology is referred to as ontology population [8]. Ontology learning concerns with knowledge discovery in different types of data sources and with its representation via an approach for automating or semi-automating the knowledge acquisition process. Further, ontology learning development consists a set of layers[1, 9]. Such layers are shown in Figure 2.2. Ontology learning layer cake consists of concepts, relationships between them, and axioms. It is necessary to identify the natural language terms that refer to them in order to identify the concepts of a domain. This is particularly essential for ontology learning from a domain of free text. Similarly, identifying Synonym helps to avoid redundant concepts when two or more natural language terms can represent the same concept. In the future, this identification helps to uniquely identify their respective concepts. Identifying the taxonomic relationships (generalisation and specialisation) between the concepts is the next step. Non taxonomic relations are also necessary to extract.

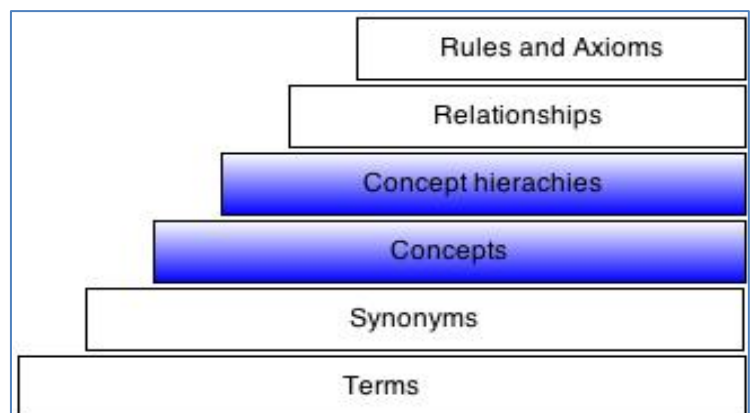

Figure 2.2: Ontology learning layer cake (adapted from [1, 9])

These layers and their related concepts are detailed below:

**1. Terms:** Terms is a pre-requisite for all aspects of ontology learning from text and they can be defined as linguistic realisations of domain-specific concepts and thereby central to further complex tasks [1, 9, 10].

**2. Synonyms:** Synonyms addresses the acquisition of semantic term. The use synonyms is provided by WordNet and related lexical resources.

**3. Concepts:** Concepts can be anything about which something is said [10]. They can be concrete or abstract, composite or elementary, fictitious or real, or the description of a task, action, function, strategy, or a reasoning process. Concepts are represented by nodes in the ontology graphs.

**4. Conceptual Relations:** Relations can be studies in two ways [1, 10]. The first way is a node in the ontology that is a concept and may be learned as the other concepts. The second way is to relate two or more concepts and thereby it should be learned as a subset of a product of n concepts (for n > 1). In the other words, relations may be learned intentionally independent of the relations of concepts or extensionally considering relations between concepts. For example, the *part-of* binary relation is a concept under the super concept "relation" and has its own features and may be related to other concepts by some other relations as well. Alternatively, it relates the concepts, for example, "door" and "house" or "hand" and "human" can be shown as (part-of door house) or (part-of door house). In addition, conceptual hierarchy may be non-taxonomic or taxonomic relations. Non-taxonomic relations are any relations between concepts with the exception of ISA relations, such as synonymy, antonymy, meronymy, attribute-of, possession, causality and other relations. On the other hand, taxonomic relations are used for organising ontological knowledge by using generalisation or specialisation relationships through which simple or multiple inheritances can be applied.

**5. Axioms:** Axioms are used to model sentences that are always true [9, 10]. In ontology, they aim constraining the information contained in the ontology, deducing new information or verifying its correctness.

**6. Rules:** In spite of learning ontological knowledge by systems, there are several systems that may learn how to extract and learn ontological knowledge. Some systems may learn met-knowledge, such as rules for extracting relations, instances, and specific fields from the Web.

**2.3.1 Ontology Learning Input Sources and Methodologies**

Input resources mainly can be of three types: unstructured, semi-structured, and structured. There are also different types of information retrieval and extraction, such as Natural Language Processing techniques, clustering, machine learning, semantics, morphological or lexical or combination of them. The learning process can be performed from the scratch or some prior knowledge can be used. The first aspect is about the availability of prior knowledge. Prior knowledge is used in the construction of a first version of ontology and such prior knowledge can demand little effort to be transformed into the first version of ontology. Then, this version can be extended automatically via learning procedures and by a knowledge engineer manually. The second aspect is about the type of input used by the ontology process. The three different types of input sources are detailed below [7]:

**1. Ontology Learning from Structured Data:** These ontology learning procedures extract the parts of ontology using the available structured information. Database schemas, existing ontologies, and knowledge base are examples of structured information sources. The main issue in learning ontology with structured information sources is to determine which pieces of structured information can provide suitable and relevant knowledge [11]. For example, a database schema might be exploited for identifying concepts and their relationships as well.

**2. Ontology Learning from Semi-structured Data:** The result quality of ontology learning procedures using structural information is usually better than the ones using unstructured input information completely. Dictionaries are example of semi-structured data, such as WordNet. WordNet is one of the manually compiled electronic dictionaries that is not restricted to any specific domain and covers most English nouns, verbs, adjectives, and adverbs [12]. WordNet is the most successful product among its similar products because it offers researchers an ideal, many of which were not initially envisaged by the authors, for disambiguation of meaning, information retrieval and semantic tagging. Along with this, it is well-documented open source and cost-free.

**3. Ontology Learning from Unstructured Data:** Learning ontology from unstructured data are those methods that do not rely upon structural information rather than unstructured data. Unstructured data is an important source of data for learning ontology because unstructured data is the most available format for ontology learning input. Unstructured data composes of natural texts, such as PDF and Word documents, or webpages.

**2.3.2 Ontology Learning Approaches**

In fact, different approaches and methods have been proposed to point out the problem of automatically or semi-automatically learning ontology processes and particularly deriving concept hierarchies from different sources of inputs. They can be classified based on their domain and features as well. The most commonly used methods on free text are clustering, Formal Concept Analysis (FCA), Association rules, and frequency-based. Table 2.1 gives a brief classification of ontology learning methods based on their input sources, featured used and purpose.

| Domain | Method | Featured used | Purpose | Papers |
|---|---|---|---|---|
| Free text | Clustering | Syntax | Extract | [15][17] |
| | Inductive logic-programming | Syntax, Logic representation | Extract | [13] |
| | Association rules | Syntax, Tokens | Extract | [20][21] |
| | Frequently-based | Syntax | Prune | [14] |
| | Pattern-matching | | Extract | [14] |
| | Classification | Syntax, semantics | Refine | [20] |
| | Formal Concept Analysis | Concept syntax, semantics | Refine | [15][17][18] |
| Dictionary | Information extraction | Relations | Extract | [15] |
| | Page rank | Tokens | | [16] |
| Knowledge base | Concept Induction, A-Box mining | Relations | Extract | [17] |
| Relational schema | Data correlation | Relations | Reverse engineering | [18] |

Table 2.1: Classification of ontology learning approaches (adapted from [19])

For extracting knowledge from unstructured data, there are some methods to be used, such as statistical and linguistic approaches. Statistical approaches usually reply on word co-occurrence and word frequencies, while linguistic approaches make use of the techniques of natural language processing for extracting information from text [20]. There are also pattern based approaches that can search for a certain patterns to indicate some kind of relations from free text. Clustering algorithms are one of the ontology learning approaches in which terms can be organised into a hierarchy to be directly transformed into an ontology prototype. Most importantly, a novel comparison of different ontology learning approaches in [21] is presented in regard with the task of automatically or semi-automatically taxonomies from textual data. It concludes that the examined approaches have a comparable performance regarding effectiveness, efficiency, and traceability. In comparison to the other methods, Formal Concept Analysis has relatively a better effectiveness, efficiency, and traceability because it not only produces clusters, but also intentional descriptions of these clusters thus fascinating their understanding.

According to [21], different methods have been proposed to address the issue of deriving a concept hierarchy automatically or semi-automatically from text via clustering analysis. These approaches can be classified into two groups. The first group is the similarity-based approaches and the second group is Set-theoretical approaches. Both approaches adopt a vector-space model and represent a word or term as a vector containing attributes or features derived from a certain corpus. The first type of approaches is characterised by using a distance or similarity measure so as to compute the pairwise distance or similarity between vectors corresponding to terms. This has to be done in order to decide whether they are semantically similar or not. If they are similar, they should be clustered. Further, the similarity-based clustering are grouped into divisive (top-down) and agglomerative (bottom-up). On the other hand, set-theoretic approaches order the objects partially in accordance with the inclusion relationship between their feature sets. The current solutions and approaches of automatic ontology building system are sufficient to understand that there are a lot of problems still need solving. Currently, several states of art in ontology learning are available, but research papers focusing on automatic ontology generation are rare. The authors of [10] provide a framework which classifies software and techniques for building ontologies in six main dimensions. It is relatively detailed and fascinating classification, but it focuses only on the learning methods and classifying and comparing the ontology learning systems by introducing a new framework. It also gives an overview of the ontology learning systems in accordance with the framework.

In [22], the authors present a useful and comprehensive tutorial about learning ontology from text, yet the considered corpus source does not sit fit with the use case of this project and dissertation. In addition, a novel approach presents in [21] about Formal Concept Analysis that it has the advantage that it not only produces clusters, but also intentional descriptions of these clusters and thereby facilitates their understanding. Most significantly, a novel comparison of different ontology learning approaches is presented with regard of the task of automatically or semi-automatically taxonomies from textual data. It concludes that the examined approaches have a comparable performance regarding effectiveness, efficiency, and traceability. In comparison to the other methods, Formal Concept Analysis has relatively a better effectiveness, efficiency, and traceability. Irrespective of performance, every approach has its own benefits. It is also presented that Formal Concept Analysis not only produces clusters, but also intentional descriptions of these clusters thus fascinating their understanding. Likewise, another novel approach is presented in [23] to acquire hierarchies automatically from domain-specific texts by using Formal Concept Analysis. This approach is evaluated by comparing it with hierarchical agglomerative clustering algorithm as well as with Bi-Section-KMeans and found that this approach produces better results on the two datasets considered. Nevertheless, this evaluation may not be sufficiently efficient as it is not based on computing how many of the super-concept relations in the automatically learned ontology are correct or computing the similarity of the automatically learned concept hierarchy with a given hierarchy for the domain in question. Further, the impact of a smoothing technique is analysed in order to deal with data sparseness and found that it does not improve the results of Formal Concept Analysis based approach. Through the lens of Formal Concept Analysis, a method of studying semantics of concept hierarchy in database system based on the theories of Formal Concept Analysis is proposed in [24]. To generate semantic rules in database systems, partial order relation of attribute based on partial order collection of attribute can be formal rules. As a consequence, this research states that Formal Concept analysis is an effective method and is beneficial supplement of the current methods. Moreover, concept hierarchies learning for getting more concise concept representation method has proposed by using Formal Concept Analysis with incorporation with fuzzy logic to deal with vague and uncertain information in practice. Also, concept hierarchies based classifier has been produced [25]. As a result of an experiment, the classification based on concept hierarchy has almost the same classification accuracy with that from original concept lattice. Despite of applying Formal Concept Analysis, some other approaches have been used in ontology learning and concept hierarchies, such as association rules and clustering algorithms yet their results have not been more effective and concise than the Formal Concept Analysis [26, 27]. Meanwhile, a lot of researches have been conducted with the use of WordNet for a number of different purposes, such as automatic text classification, and word sense disambiguation. Further, determining the similarity between words is another prominent example of the use of WordNet. Several algorithms have been proposed including the distance between the conceptual types of words, and the hierarchical structure of the WordNet ontology. In that sense, measuring similarity of words in a generic and hierarchical corpus have been presented by [28, 29]. Nevertheless, no research has been done in regard with learning concept hierarchies from free text by using both Formal Concept Analysis and WordNet.

Thus, a lot of researches have been conducted on different aspects of ontology learning. Nevertheless, there are still some challenging open problems that may need more attention. Much work has been done on extracting taxonomic relations, whereas less work has been done on discovering non-taxonomic and much less work has been conducted on axiom learning. Most of the works, on the other hand, conducted are domain-specific and have built ontologies for a specific domain. As a result, building ontologies automatically or semi-automatically still needs more work. To do so, some of the open problems to be considered for improvement in the field are concepts, concept hierarchies, relationships, and axiom learning. In that sense, a framework is presented in this project and dissertation for learning concept hierarchies from domain-specific texts by using Formal Concept Analysis and WordNet, and an experiment is conducted on that basis. The framework is based on Formal Concept Analysis, a method of deriving a concept hierarchy automatically or formal ontology from a collection of objects and their properties.

In brief, Formal Concept Analysis is applied on free text with the help of WordNet in this project and dissertation to automate concept hierarchies of ontology engineering process, to evaluate the results in comparison with constructing concept hierarchies manually, and to find an efficient and accurate way to do so between them.

## 2.4 Information Extraction and Text Mining

Several methods from a diverse spectrum of fields are used by ontology learning [30], such as artificial intelligence and machine learning, knowledge acquisition, natural language processing, information retrieval, and reasoning and database management. Information extraction is to extract structured information from unstructured data, whereas text mining is related to information extraction as it is used to find relevant pattern and trends. Text mining usually involves the process of deriving linguistic features from the text and removal of unimportant words for mining. The categorisation of text is a typical task of text mining.

**2.4.1 Natural Language Processing**

The main goal of research about Natural Language Processing (NLP) is to understand and parse language. Natural Language Processing is a subfield of linguistics that consists of understanding of natural human language and automatic generation. Natural Language Processing systems analyse text to find group words and phrase structure in accordance with their semantic and syntactic type. Text mining and information extraction are usually dependent of NLP. For instance, text mining could be used in finding relations between nouns in a text. As a result, text is pre-processed by a Part of Speech Tagger (POST) that groups the words in grammatical types. The most important components of Natural Language Processing that are used in this project and dissertation are tokenisation, sentence splitter, Part-of-Speech Tagger, and lemmatisation.

**2.4.2 Stanford Natural Language Processing Framework**

The Stanford Natural Language Processing Group is a team of faculty, postdocs, research scientists, students and programmers who work together on algorithms that allows computer to process and understand human languages, such as English and Kurdish languages. Stanford Natural Language Processing tools ranges from basic research in computational linguistics to key applications in human language technology and covers areas, such as sentence understanding, probabilistic parsing and tagging, machine translation, biomedical information extraction, word sense disambiguation, grammar induction, and automatic question answering. One of the set of natural language analysis tools of this group is Stanford Natural Language Processing framework. It is an integrated framework, and a suite of open-source and natural language processing tools for English language that is written in Java [31]. In addition, Stanford Natural Language Processing framework provides a set of natural language analysis tools that can take free English language text as an input and give the based forms of words and their parts of speeches. Furthermore, it includes tokenisation, part-of-speech tagging, named entity recognition, coreference, and parsing. On the other hand, the Stanford Natural Language Processing framework API is used by Annotation and Annotator classes. Both of them form the backbone of the Stanford Natural Language Processing package. Annotations are the data structure that can holds the result values of annotations, which are basically maps, such as part-of-speech, parse, or name entity tags. On the other hand, Annotators are relatively like functions that they do things, such as parse and tokenise, but they operate over Annotations instead of Objects. Both Annotations and Annotators are integrated by AnnotationPipelines that create sequences of generic Annotators. Stanford Natural Language Processing framework inherits from the AnnotationPipeline class and is customised with Natural Language Processing Annotators. Table 2.2 describes the needed Annotations in this project that are currently supported and the Annotations that they are generated by Stanford Natural Language Processing.

| Property name | Generated Annotation | Description |
| --- | --- | --- |
| Tokenise | List of tokens, and chracterOffsetBeginAnnotation, characterOffsetEndAnnotation, TextAnnotation | Tokenises the text. This component started as a PTB-style tokeniser, but was extended since then to handle noisy and web text. |
| Split | SentencesAnnotation | Splits a sequence of tokens into sentences |
| Pos | PartOfSpeechAnnotation | Labels tokens with their POS tag |
| Lemma | LemmaAnnotation | Generates the word lemma for all tokens in the corpus |
| Ner | NamedEntityTagAnnotation and NormalisedNamedEntityTagAnnotation | Recognises named (PERSON, LOCATION, ORGANISATION, MISC) and numerical entities (DATE, TIME, MONEY, NUMBER) |
| Regexner | NamedEntityTagAnnotation | Implements rule-based NER simply over taken sequences using Java regular expressions |
| Truecase | TrueCaseAnnotation and TrueCaseTextAnnotation | Recognises the true case of tokens in the text where this information was lost. For example, all lower case text |
| Parse |  | Provides syntactic analysis fully using both the constituent and the dependency representations |

Table 2.2: Summarisation of the needed Annotators in this project and dissertation

**2.4.3 WordNet**

WordNet is a lexical electronic database that is considered to be the most important resource available to researchers in computational linguistics [32, 33]. The purpose of WordNet is twofold: to support automatic text analysis and artificial intelligence applications. It is also known as one of the most used manually compiled electronic dictionaries that is not restricted to any specific domain and covers almost all English verbs, nouns, adjectives, and adverbs. Even though there are similar products, such as CYC, Cycorp, and Roget's International Thesaurus, WordNet is the most successful and growing one and it has been used in several applications over the last ten years [12]. Furthermore, knowledge structure and database contents of WordNet are focused below:

**1. Knowledge Structure in WordNet:** In WordNet, both verbs and nouns are structured into a hierarchy which are defined by hypernym or IS-A relationships [33]. For example, the first sense of the word cat has the following hyponym hierarchy as shown in figure 2.3. The words at the same level are synonyms of each other. For instance, some sense of cat is synonymous with some other senses of domestic cat and wildcats and so on. Also, each set of synonyms or synset has a unique index which shares its properties, such as a dictionary or gloss definition.

```
cat, true cat -- (feline mammal usually having thick soft fur and being unable to roar; domestic cats; wildcats)
    => feline, felid -- (any of various lithe-bodied round-headed fissiped mammals many with retractile claws)
        => carnivore -- (terrestrial or aquatic flesh-eating mammal; terrestrial carnivores have four or five clawed digits on each limb)
            => placental, placental mammal, eutherian, eutherian mammal -- (mammals having a placenta; all mammals except monotremes and marsupials)
                => mammal -- (any warm-blooded vertebrate having the skin more or less covered with hair; young are born alive except for the small subclass of monotremes and nourished with milk)
                    => vertebrate, craniate -- (animals having a bony or cartilaginous skeleton with a segmented spinal column and a large brain enclosed in a skull or cranium)
                        => chordate -- (any animal of the phylum Chordata having a notochord or spinal column)
                            => animal, animate being, beast, brute, creature, fauna -- (a living organism characterized by voluntary movement)
                                => organism, being -- (a living thing that has (or can develop) the ability to act or function independently)
                                    => living thing, animate thing -- (a living (or once living) entity)
                                        => object, physical object -- (a tangible and visible entity; an entity that can cast a shadow; "it was full of rackets, balls and other objects")
                                            => entity -- (that which is perceived or known or inferred to have its own distinct existence (living or nonliving))
```

Figure 2.3: WordNet knowledge structure

**2. The Database Contents of WordNet:** WordNet could distinguish between verbs, nouns, adjectives, and adverbs as they follow different grammatical rules, but it does not include determiners and prepositions [32]. Every synset includes a group of collocations or synonymous words and various senses of a word are in various synsets. Glosses are further clarified by defining the meaning of synsets. The here below is an example of synset with gloss:

Good, right, ripe – the right or most suitable for a particular purpose
"A good time to plant tomatoes"
"The right time to act"
"The time is ripe for great sociological changes"

Almost all synonym sets are connected to other synsets through a number of relations. These relations may vary based on the word families, such as noun, verb, adjective, and adverb. Table 2.4 presents the semantic relations between the synsets and word families in the WordNet where 1 represents an existing relation and 0 represents no existing relationship.

|                     | Nouns | Verbs | Adjectives | Adverbs |
|---------------------|-------|-------|------------|---------|
| Hypernyms           | 1     | 1     | 0          | 0       |
| Hyponyms            | 1     | 0     | 0          | 0       |
| Coordinate terms    | 1     | 1     | 0          | 0       |
| Holonym             | 1     | 0     | 0          | 0       |
| Meronym             | 1     | 0     | 0          | 0       |
| Troponym            | 0     | 1     | 0          | 0       |
| Entailment          | 0     | 1     | 0          | 0       |
| Related nouns       | 0     | 0     | 1          | 0       |
| Similar to          | 0     | 0     | 1          | 0       |
| Participate of verb | 0     | 0     | 1          | 0       |
| Root adjectives     | 0     | 0     | 0          | 1       |

Table 2.4: Relation between synsets and word families in the WordNet

Because semantic relations can apply to all members of a synset and share a meaning, and are mutually synonyms, words can be connected to other words via relations including synonyms and hypernyms. Hypernyms can be defined as if Y is hypernym of X if every X is a kind of Y. For example, canine is a hypernym of dog. Alternatively, Hyponyms can be defined as if Y is a hyponym of X if every Y is a kind of X. For instance, dog is a hyponym of canine. Likewise, synonym can be defined as if a word X has the same meaning as word Y. For example, pretty has the same meaning as beautiful in the WordNet dictionary for the English language.

Incidentally, WordNet-based technique can make use of hypernyms and synonyms to reduce the size of formal context for both objects and attributes that are represented by nouns and verbs respectively.

## 2.5 Formal Concept Analysis

Formal Concept Analysis (FCA) is one of methods of ontology learning that basically used for data analysis and deriving relations between objects and attributes [34]. FCA aims deriving formal ontology or a concept hierarchy from a collection of objects and their attributes. In a concept hierarchy, each concept is represented by a set of objects sharing the same values for a certain set of attributes, whereas each sub-concept comprises of a subset of the objects in the concepts above it. In terms of lattice theory, FCA is a mathematical theory modelling the concept of "concepts" that starts with a formal concept so as to allow a mathematical description of instructions and extensions and deal with how objects can be hierarchically grouped together in accordance with their common attributes. Attribute logic as the study of possible attribute combination is one of the common aspects of FCA. Concept lattice is an ordered set in which any two elements have infirmum and suremum and can be defined as a set of all formal concepts of a context K together with the order relation $\leq$ is often a complete lattice which is called the concept lattice of K and can be denoted by B(K) [35]. Moreover, FCA is a method which is mainly used for data analysis, such as deriving implicit relationships between objects described by a set of attributes on the one hand and these attributes on the other hand. The main aim and meaning of Formal Concept Analysis is to support the relational communications of humans by mathematically developing appropriate conceptual structures that can be logically activated [36]. Formal Concept Analysis uses order theory for analysing the correlations between objects, G, and their measures M. it identifies from a data description called formal context K as a set of features $B \subseteq M$ being correlated with its set of objects $A \subseteq G$. This correlated pair is called a formal concept (A, B). Meanwhile, formal concepts are partially ordered by $(A, B) \leq (A, B) \leftrightarrow A_1 A_2$ or, which is equivalent, $B_2 \subseteq B_1$. Objects $o_1, o_2 \in G$ are clustered iff $\{o_1, o_2\} \subseteq A$, where (A, B) is a formal concept in K.

### 2.5.1 Formal Contexts and Concepts

Formal concept can be represented by a set of objects sharing the same values for a certain set of attributes, whereas each sub-concept comprises of a subset of the objects in the concepts above it. According to [35], formal context can be formally defined as a triple K := (G, M, I), where G is a set whose elements are called object, M is a set whose elements are called attributes, and I is a binary relation between G and M. That means, $I \subseteq G \times M$ and $(g, m) \in I$ is read "object g has attribute m". For the purpose of this project and dissertation, formal context can be described in a table because it is the simplest format for writing down a formal context. The objects can be corresponding to the rows of the table, while the attributes can be corresponding to the columns of the table. The cells (x, y) of the table represent Booleans values wherever object x has value y. Table 2.5 gives a table format of formal context.

|  | Latin America | Europe | Canada | Middle East | United States |
|---|---|---|---|---|---|
| **Air Canada** | 1 | 1 | 1 | 0 | 0 |
| **British Airways** | 1 | 1 | 1 | 0 | 1 |
| **Scandinavian Airline** | 1 | 1 | 1 | 1 | 0 |
| **Turkish Airline** | 0 | 1 | 0 | 1 | 1 |
| **United Airlines** | 0 | 1 | 0 | 1 | 1 |
| **Iraqi Airways** | 0 | 1 | 0 | 1 | 0 |

Table 2.5: A formal context about the destinations of airlines and destinations

### 2.5.2 Concept Lattice

Formal Concept Analysis deals with how objects can be hierarchically grouped together in accordance with their common attributes. Attribute logic as the study of possible attribute combination is one of the common aspects of FCA. Concept lattice can be defined as a set of all formal concepts of a context K together with the order relation $\leq$ is often a complete lattice which is called the concept lattice of K and can be denoted by B(K) [35]. That is, for each subset of concepts, there is always a unique least super-concept and a unique greatest common sub-concept as well. Figure 3.3 depicts the concept lattice of the formal context in Figure 2.4 by the Concept Explorer. In a Concept Explorer, each node represents a formal concept. For instance, a concept c2 is a super-concept of a concept c1 if and only if there is a path of descending edges from the node representing c2 to the node representing c1. Alternatively, a concept c3 is a sub-concept of a concept c4 if and only if there is a path of descending edges from the node representing c4 to the node representing c3.

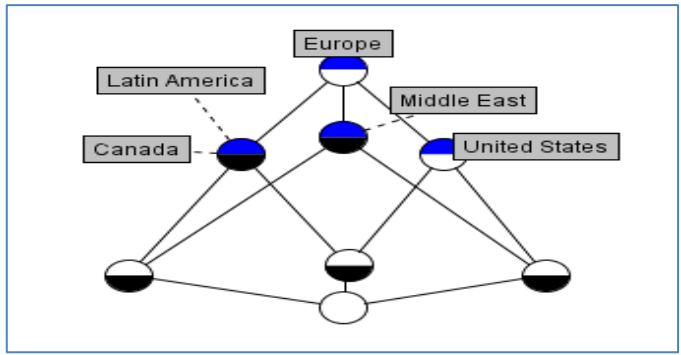

Figure 2.4: The concept lattice of the formal context in Figure 2.1

In spite of the formal context and lattice diagram, the implications between attributes valid in a context can be examined. The here below is the implications sets of the formal context of Table 2.6

*1 < 0 > building allude ==> reference;*
*2 < 0 > reference ==> building allude;*

**2.5.3 First Example**

As already mentioned earlier, syntactic dependencies between the verbs and subject, object, and prepositional in the text collection as a form of verb-noun so as to derive context attributes describing the interested terms. Generally, these dependencies lead a better result than any subsets of them. Stanford Natural Language Processing is used to extract these dependencies automatically. Then, these dependencies can be represented in formal contexts. For example, let us take the following below two sentences:

*The museum houses an impressive collection of medieval and modern art. The building combines geometric abstraction with classical references that allude to the Roman influence on the region.*

After parsing these two sentences and applying all NLP Stanford, the following below interested dependencies are extracted:

museum-house, collection-house, modern-medieval, building-combine, abstraction-combine, reference-allude

These dependencies can be represented as a form of formal context. Table 2.6 presents the dependencies of the above example.

|         | building | reference | allude |
|---------|----------|-----------|--------|
| house   | 1        | 0         | 1      |
| combine | 1        | 0         | 0      |
| reference | 0      | 0         | 1      |

Table 2.6: The concept lattice of the context in Figure 2.1

Then, the concept lattice can be derived from the above formal context. In this sense, Figure 2.4 shows the concept lattice of formal context in Table 2.5

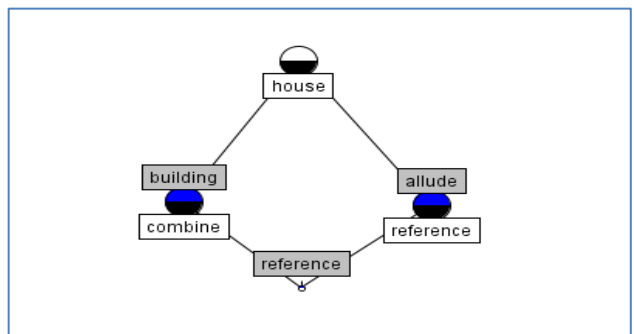

Figure 2.4: The concept lattice of the context in Figure 2.1

**2.5.4 The Concept Explorer**

The Concept Explorer is an open source project which was firstly developed as a part of a master thesis. It implements the basic ideas needed for the research and study of Formal Concept Analysis (FCA) and lattice compaction [37]. This software provides the mathematical theory and belongings to algebra it is a branch of lattice theory. In addition, the Concept Explorer can be used to analyse attribute-object tables which are called formal context in Formal Concept Analysis, draw the corresponding lattice and to explore different dependencies that may exist between attributes. Besides, the Concept Explorer provides the following tools:

- Building and editing formal contexts and concepts
- Building concept lattices from a formal context
- Finding a base for implications and association rules that are true in formal context
- Performing attribute exploration

**2.6 Summary**

In this chapter, a theory, background, and the state of art of ontology and ontology learning have introduced. The relevant techniques of natural language processing and information extraction have presented. First, a detail has given regarding ontologies, and then automatic ontology generation lifecycle has explained. Second, ontology learning including ontology learning input sources and methodologies and ontology learning approaches has presented. Last but not least, an overview has given in regard with information extraction and text mining that composes of Natural Language Processing, Stanford Natural Language Processing framework, and WordNet. Last, the basic idea of Formal Concept Analysis and lattice compaction for the purpose of research and study of this project and dissertation has introduced with giving details about formal contexts and concepts, concept lattice, first example of Formal Concept Analysis, and the Concept Explorer as subsections.

# 3. Chapter Three: Automating Concept Hierarchies

In the previous chapter, a theory and background of ontology and ontology learning have introduced and the relevant techniques of natural language processing and information extraction have presented. It has focused on Stanford Natural Language Processing framework, the Concept Explorer as open source software for implementing Formal Concept Analysis and Lattice Compaction, WordNet as a lexical database for the English language, and particularly an approach has presented based on Formal Design Analysis based on order theory and mainly used for data analysis. These could be presented together to propose a novel approach for deriving concept lattice and could be converted into a special type of partial order constituting a concept and concept hierarchy accordingly. On the basis of chapter two, this chapter describes design phase of the overall concept hierarchy process and a method for discovering concepts and concept hierarchies in a free text by using Formal Concept Analysis and WordNet. At the first glance, the needed three steps of automatic ontology generation lifecycle are focused. Afterwards, the overall process of how to learn concepts and concept hierarchies from free text by ten steps is explained. Then, Formal Concept Analysis as one of the ontology learning techniques is defined and detailed by focusing on formal contexts, concepts and concept lattices, and giving an example about applying Formal Concept Analysis on free text. Finally, WordNet-based and Frequency-based techniques as two distinct techniques that can be applied on formal context are demonstrated and detailed via examples

## 3.1 Required Steps in Automatic Generation Lifecycle

The main task of this project and dissertation can be represented by the first steps of automatic generation lifecycle. These three steps start from building ontologies from an existing corpus source.

**1. Extraction:** this provides the needed information acquisition to generate the ontology by starting from an existing corpus source. Unstructured data, such as free text is used as input resource in this step.

**2. Analysis:** this stage focuses on alignment and/or the matching of retrieved information of two or more existing ontologies. This step needs: the used techniques in the first step; a semantic analysis to detect homonyms , synonyms,  and other relations of this type; an analysis of concepts' structure to find hierarchical relationships and identify common attributes; techniques based upon reasoners in order to detect inconsistencies and induced relations. For this step, Stanford Natural Language Processing framework, WordNet are used.

**3. Generation:** this step is about the formalisation of the meta-model used by the tool in a more general formalism which is interpretable by other applications. The implementation section and the Concept Explorer as open source software for implementing Formal Concept Analysis and lattice compaction represent this step.

## 3.2 Overall Process of Concept Hierarchies

The overall process of automating concept hierarchies by using Formal Concept Analysis includes a set of steps. Firstly, the document is part-of-speech (POS) tagged, and then parsed to yield a parse tree for each sentence. Next, verb/object, verb/subject, and verb/prepositional phrase dependencies are extracted from the parse trees. In particular, pairs are extracted that contain verb and the head of the object, subject or prepositional phrase. Then, the verb and the head are lemmatised, i.e. assigned to their base form. Nonetheless, the output of the parser may be erroneous. That is, not all of the derived pairs are correct or interesting. We will investigate two techniques for eliminating the erroneous and uninteresting pairs and their terms: firstly, the pairs are compared with the WordNet as a lexical database for English language; secondly, the pairs are weighted based on some statistical measure and thereby only the pairs over a certain threshold are transformed into a formal context to which Formal Concept Analysis is applied. Finally, lattice of formal concepts is computed and results in a traditional concept hierarchy. This approach is evaluated by comparing the resulting concept lattices with the same one that the two techniques have not applied to it. The algorithm below represents the overall picture of framework.

Constructing concept hierarchy (D,T)

/* construct a hierarchy for the terms in T on the basis of the documents in D */

1: Tokenises = parse (POS-tag(D))
2: SynDeps = tgrep(Parses)
3: Lemmatise (SynDeps);
4: WordPairs (SynDeps);
5: Pruning (SynDeps);
6: Filtering (SynDeps);
7: K = getFormalContext (T,SynDeps);
8: () = computeLattice (K);
9: return (C", ≤");

The main steps of concept hierarchies depict in Figure 3.1. The overall project design starts from selecting a text corpus as a source of ontology learning.

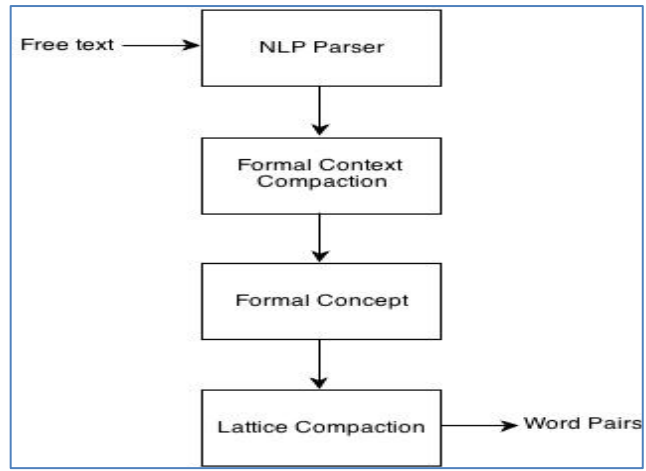

Figure 3.1: The overall process for concept hierarchy retrieval

### 3.2.1 Natural Language Processing Parser

Natural Language Processing Parser is a set of natural language analysis tools that can take free English language text as an input and give the based forms of words and their parts of speeches including tokenisation, part-of-speech tagging, named entity recognition, coreference, and parsing. This could be done by using Stanford Natural Language Processing framework. Figure 3.2 gives the components that can be used for Natural Language Processing parser. It begins with splitting free text and extracting word pairs at the end. This is explained in the below five points:

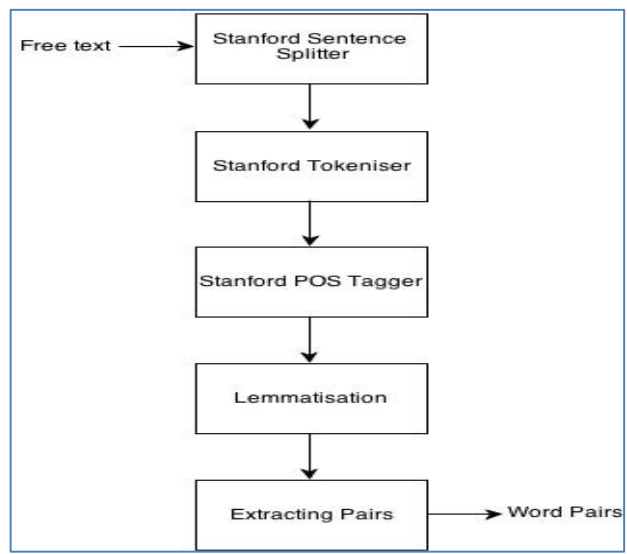

Figure 3.2: Natural language processing components

**1. Sentence Splitter:** Sentence splitter is on the first steps of text processing. It tries to find the sentence boundaries from a given chunk of text. Sentences are usually marked by punctuation marks or periods, but these characters can serve for other purposes.

**2. Tokenisation:** Tokenisation is the process of breaking a stream of texts or a sentence into units called tokens [38]. These tokens are either words, symbols, phrases, numbers, or other meaningful elements in a sentence. Tokenisation typically occurs at the word level. English as a Latin alphabet based language uses a straightforward approach to define its words called inter-word spaces. Unlike Chinese language, this approach is not difficult in English language because it have word boundaries. Furthermore, phrases are detected as a token from the elements of a sentence. Languages have constraints in word order and thereby words in languages are grouped and organised in units called phrases, such as noun and verb phrases [38]. Noun phrase (NP) is a syntactic unit of a sentence where information about the noun is gathered. Noun phrases usually consist of zero or more than one adjectives. On the contrary, verb phrase (VP) is a syntactic unit of a sentence where information about the verb is gathered. In a verb phrase, the elements of verb phrase are words that syntactically depend upon the verb. Hence, phrase is usually detected as a token and activated by rules which define patterns of Part-of-Speech Tags. Using statistical measures where a phrase is defined as a sequence of words which co-occur frequently is another approach.

**3. Part-of-Speech Tagger:** Determination of the part of speech (POS) of a word an essential precursor to several Natural Language Processing tasks. POS-Tagged is a process used as a reasonable compromise between accuracy and utility [39]. It consists of labelling each word in a sentence with its appropriate part of speech, such as nouns, verbs, adjectives, and prepositions. This process is based upon a predefined tagging scheme. Training the tagger to use an already tagged document collection or corpus is the most common scheme of tagging. This could resolve word ambiguity they may appear using an approach, such as looking up a word in a dictionary to find its part-of-speech since the word may belong to several morphological classes. Thus, statistical and rule-based models are the most common approaches for training a tagger.

**4. Lemmatisation:** Lemmatisation is the process of finding the normalised form of a word [40]. In other words, lemmatisation is the same as looking for a transformation to apply on a word to find the base form of word's lemma. This could be achieved by looking a given word in a lemma list or dictionary. For example, the lemma of the word "walked" is "walk". This is essential in text mining and information extraction because "walked" and "walk" would be considered as two different words without the lemmatisation process. In contrast, stemming process is to find a base or root form of a word by stripping off its affixes, suffix, or prefixes [41]. This process is usually done by using rules. For instance, both "manager" and "management" can be reduced to the stem "manage".

**5. Word Pairs Detection:** Finding grammatical relations and dependencies between words in a text is provided by Stanford Natural Language Processing framework in a package called Stanford Dependencies (SD). Stanford Dependencies provide a representation of grammatical relations between words in a sentence. Stanford dependencies have been developed to be effectively used and easily understood to extract textual relations. Stanford Dependencies are in triple form: relation name, dependent and governor. Similarly, word pairs can be defined as a two corresponding words which are similar in grammatical form or function in a sentence. the Stanford typed dependencies representation is designed to provide a straightforward description of the grammatical relations in a sentence that can easily be understood and effectively used by an user who could benefit from automatic text understanding without linguistic expertise who want to extract textual relations [42]. In other words, it represents sentence relations uniformly as typed dependency relations, such as "Bell distributes…" as nsubj(distributes-10, Bell-1). For such purposes of this project, the Stanford typed dependencies are suitable to pair the interested words in order for the formal context to be constructed. The here below is an example sentence:

*Bell, based in Los Angeles, makes and distributes electronic, computer and building products.*

For this sentence, the Stanford Dependencies (SD) representation is:

*nsubj(makes-8, Bell-1), nsubj(distributes-10, Bell-1), partmod(Bell-1, based-3), nn(Angeles-6, Los-5), prep in(based-3, Angeles-6), root(ROOT-0, makes-8), conj and(makes-8, distributes-10), amod(products-16, electronic-11), conj and(electronic-11, computer-13), amod(products-16, computer-13), conj and(electronic-11, building-15), amod(products-16, building-15), dobj(makes-8, products-16), dobj(distributes-10, products-16)*

These dependencies can also map onto a directed graph representation. Each word can be resented as a node and linking each on another by edge labels. Figure 3.3 gives the graph representation for the example sentence above.

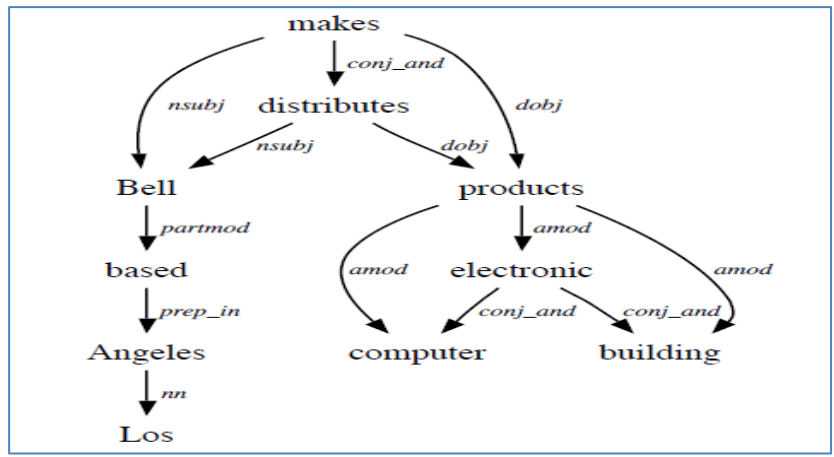

Figure 3.3: Graphical representation of the Stanford Dependencies for the example sentence

**3.2.2 Formal Context Construction**

This module mainly focuses on extracting the interested and wanted word pairs by filtering and pruning them to be used for constructing formal context. Figure 3.4 details the steps of constructing formal context the word pairs after filtering and pruning techniques. These steps are explained in the below three points:

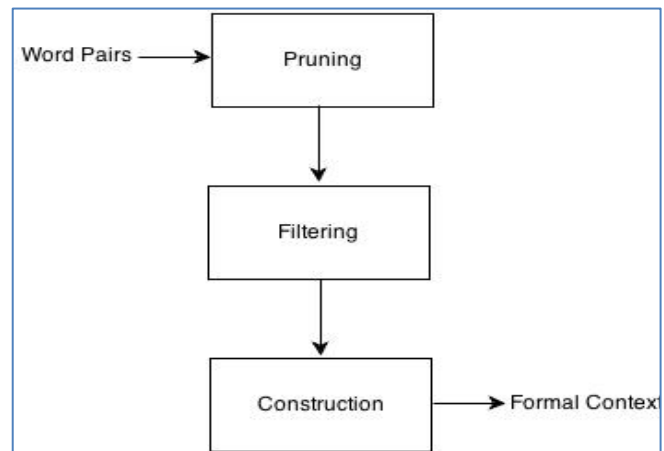

Figure 3.4: Components of formal context construction

**1. Pruning:** All of the pairs should be in the form of verb-noun, but the extracted pairs are sometimes not in that form. In this case, a technique is needed to opts to prune weights off and change them to the form of verb -noun. For example, driving - car pair might not be appropriately useful for to be represented in the next step and accordingly it should be pruned into a verb - noun format as drive - car.

**2. Filtering:** Some of the constructed pairs are erroneous and may not be interested and a technique is needed to filter the word pairs based on the necessity of this project and dissertation. The current Stanford Dependencies include approximately fifty grammatical relations [42]. These typed Stanford dependencies are used for representing the word pairs in the previous step. However, they need to be filtered in order to opt for the needed and interested grammatical relations for applying formal contexts and concepts, and context lattice are filtered. Table 3.1 lists the needed and interested Stanford typed dependencies used for this project and dissertation.

| Dependency type | Meaning |
| --- | --- |
| acomp | Adjectival complement |
| agent | Agent |
| conj | Conjunct |
| cop | Copula |
| csubj | Clausal subject |
| csubjpass | Clausal passive subject |
| dobj | Direct object |
| infmod | Infinitival modifier |
| nsubj | Nominal subject |
| nsubjpass | Passive nominal subject |
| parataxis | Parataxis |
| partmod | Participial modifier |
| prepc | Prepositional clausal modifier |
| purpcl | Purpose clause modifier |
| rcmod | Relative clause modifier |
| rel | Relative |
| tmod | Temporal modifier |
| xcomp | Open clausal component |
| xsubj | Controlling subject |

Table 3.1: Interested Stanford typed dependencies manual

**3. Formal Context Construction:** After filtering and pruning techniques on the extracted word pairs, the word pairs can be represented in a form of table as it is the simplest format for manipulation and writing down a formal context. The objects (nouns) can be corresponding to the rows of the table, while the attributes (verbs) can be corresponding to the columns of the table. The cells (x, y) of the table represent Booleans values wherever object x has value y.

### 3.2.3 Formal Concept and Lattice Compaction

The Concept Explorer can represent the structure of finite formal context and concept lattice. This is a graph with nodes and edges connecting them as well as ordered structure with a bottom and a top element. Figure 3.5 shows the how to derive concept lattice from formal context and this process is detailed in the below two points:

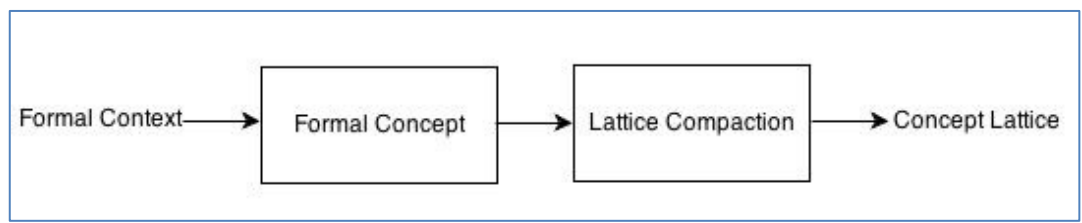

Figure 3.5: Components of formal concept and lattice compaction

**1. Working on Formal Concept:** The formal context includes information about how attributes and objects should be related and this can derive the set containing all attributes common to all objects from the set O and only these attributes. After constructing the formal context, the Concept Explorer can be used to yield formal concept and concept lattice accordingly. Each node of concept lattice corresponds to a formal concept. This can be denoted as a pair (O, A), where A is a subset of the attribute set and some additional properties are satisfied, and O is a subset of the object set.

**2. Concept Lattice and Line Diagram:** After building the formal contexts and concepts, the Concept Explorer can be used to draw the corresponding lattice and explore different dependencies that may exist between attributes from the formal contexts.

### 3.3 Formal Concept Analysis

The formal definition of Formal Concept Analysis can be illustrated by the below brief example:                                           From a corpus of text on tourism, for instance, some features can be derived and a vector representation of objects G = {apartment, car, bike, trip, excursion} can be constructed. The representation with features M := {book, …, join} is presented in table 3.2. This describes that all the given objects can be booked. Moreover, one can rent a car, a bike or an apartment and one can drive a car or a bike, but one may only ride a bike. Finally, it is derived from the corpus that one may join an excursion or a trip. Based on the binary representation of objects, Formal Concept Analysis constructs a lattice of formal concepts. The lattice corresponds to the running example of Figure 3.6. To be interpreted as taxonomy, the below two rules ought to be applied:

*People book hotels. The boy drove the bike along the beach.*

1. Applying Stanford typed dependencies representation: this is designed to provide a simple description of the grammatical relationships in a sentence that can effectively. The here below is an example sentence:

*People book hotels. Hotels can be reserved online. The boy drove the bike along the beach.*

For this sentence, the Stanford Dependencies representation is:

people :: Node   nsubj, nsubj: people – reserve, reserve :: Node   dep, dep: reserve – drive, hotels :: Node   dobj, dobj: hotel – reserve, the :: Node   det

2. After parsing these sentences, the following syntactic dependencies are extracted:

*book_subj(people), book_obj(hotels), drove_subj(boy), drove_obj(bike), drove_along(beach)*

And then, these above syntactic dependencies are lemmatised to get the following below pairs:

*book_subj(people), book_obj(hotel), drove_subj(boy), drove_obj(bike), drove_along(beach)*

In addition, the output of the parser can be erroneous and not all derived verb/object dependencies are correct and interesting in the sense that they will help to discriminate between the different objects. The verb/object dependencies in regard with a certain information measure are weighted. Those verb/object relations can be considered to be above a threshold *t*.

|  | book | reserve | rent | drive | ride | join |
|---|---|---|---|---|---|---|
| apartment | 1 | 1 | 1 | 0 | 0 | 0 |
| car | 1 | 0 | 1 | 1 | 0 | 0 |
| motor-bike | 1 | 0 | 1 | 1 | 1 | 0 |
| excursion | 1 | 0 | 0 | 0 | 0 | 1 |
| trip | 1 | 1 | 0 | 0 | 0 | 1 |

Table 3.2: Formal context representation

The formal context of Table 3.2 can be represented as concept lattice. Figure 3.6 presents the concept lattice of Table 3.2. Nonetheless, it seems that the hierarchy shown in Figure 3.6 is kind of odd because of the fact that the labels of abstract concepts are verbs rather than nouns as it is typically assumed. From the formal point of view, however, concept hierarchies do not have meaning at all such that we could have just named the concepts with some other arbitrary symbols.

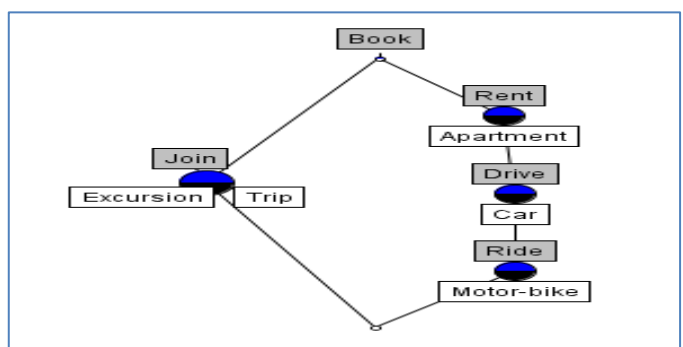

Figure 3.6: the lattice of formal concepts

From an intentional point of view, hypernym or synonym may exist with adequate intension to label a certain abstract concept so that using a verb-like identifier may even be more appropriate choice. For example, join, ride, and drive identifiers can be replaced by activity, two-wheeled vehicle by vehicle respectively. Table 3.2 presents the reduced formal context of Table 3.2. Nevertheless, this may be difficult to substitute these terms by some meaningful terms denoting the same extension without the help of a linguistic resource language, such as WordNet. This is called WordNet-based technique. The next section gives more details with example about this technique. Moreover, the task needs to be focused is can a concept hierarchy be derived form a given a certain number of terms referring to concepts from a text? In terms of formal concept, the objects can be given and we need to find their corresponding attributes so as to build an incidence formal context and concepts, and lattice compaction to be transformed into a corresponding concept hierarchy. Most importantly, after building the formal context, can the size of formal context be reduced by combining the hypernym of synonym objects and attributes? This question will be answered by conducting an experiment in chapter four.

|           | book | rent | activity |
|-----------|------|------|----------|
| apartment | 1    | 1    | 0        |
| car       | 1    | 1    | 0        |
| motor-bike| 1    | 1    | 0        |
| excursion | 1    | 0    | 1        |
| trip      | 1    | 0    | 1        |

Table 3.3: reduced formal context from Table 3.2

**3.4 Techniques Used in Formal Context**

There are two basic techniques that can be applied on formal context and used in an experiment as well: the first technique is a linguistic based technique called WordNet-based technique; the second technique is a statistical based technique called Frequency-based technique. The former one is used to compare the pairs with the WordNet as a lexical database for English language, whereas the latter one is used to weight the pairs based on some statistical measure and thereby only the pairs over a certain threshold are transformed into a formal context to which Formal Concept Analysis is applied. Figure 3.7 presents the overall process for concept hierarchy retrieval including the framework of the experiment. As it shown in the framework, formal concept and lattice compaction can be directly applied on the extended CEX file so as to extract the concept lattice 1.

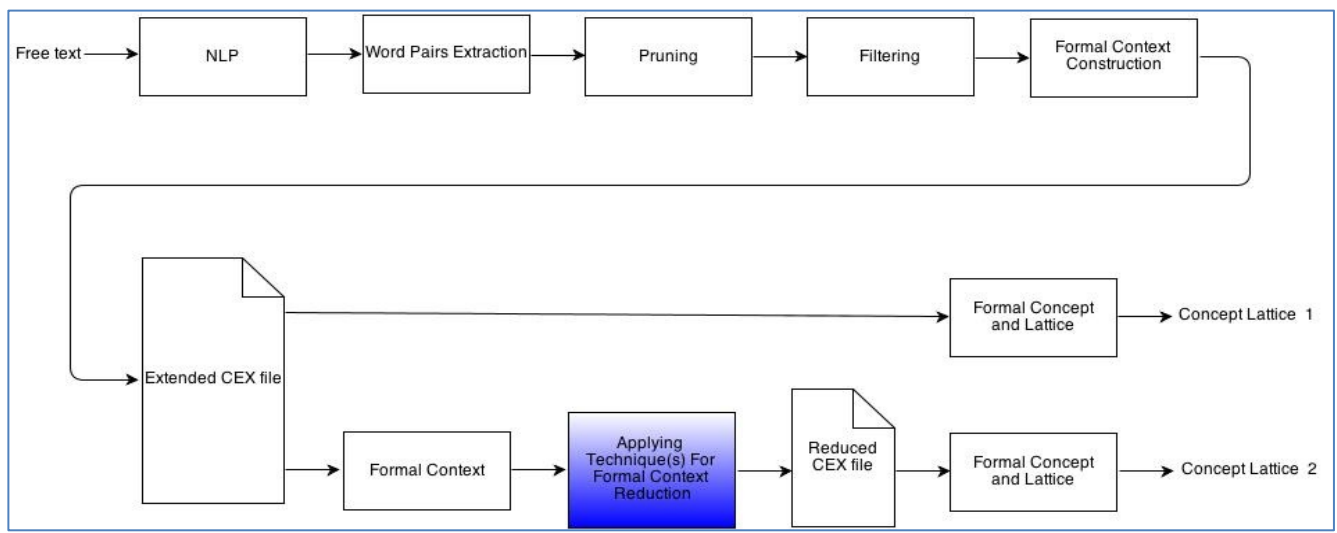

Figure 3.7: Overall process framework including the experiment

**3.4.1 WordNet-based Technique**

WordNet-based technique is a linguistic resource-based technique for the English language that can be used as a lexical database dictionary for comparing the objects and attributes of formal context to check whether they are synonyms or hypernyms of each other in order to be combined together. The objects and attributes of formal concept may have synonyms or hypernyms. For example, object A can be a synonym of object B. In this situation, both object A and B can be represented by one record. After choosing the most general terms between them, the attribute values of these objects can be merged together in one record. Table 3.4 is an example of formal context with its objects and attributes.

|   | W | X | Y | Z |
|---|---|---|---|---|
| A | 1 | 0 | 1 | 0 |
| B | 1 | 0 | 0 | 1 |
| C | 0 | 0 | 1 | 1 |
| D | 0 | 1 | 1 | 0 |

Table 3.4: An example of formal context

In the above formal concept table, if object A is a synonym or hypernym of object B, both object A and B can be merged and represented together in the form of A/B. Object A/B is literally the most general term between object A and B. If a specific attribute value of object A and B is the same, the attribute value of object A/B should be the same accordingly. That is, if the value of attribute x of object A and B is 1 or 0, the attribute value of object A/B should be 1 or 0 respectively. It is also assumed that if an attribute value of object A is 0 and the attribute value of object B is 1 or vice versa, the attribute value of object A, B would be 1. Table 3.5 shows the truth table for merging values of two objects or attributes.

| A | B | A/B |
|---|---|---|
| 0 | 0 | 0 |
| 0 | 1 | 1 |
| 1 | 1 | 1 |
| 1 | 1 | 1 |

Table 3.5: Truth table for merging A and B

The result of merging object A and B from Table 3.4 is shown in Table 3.6.

|  | W | X | Y | Z |
|---|---|---|---|---|
| A,B | 1 | 0 | 1 | 1 |
| C | 0 | 0 | 1 | 1 |
| D | 0 | 1 | 1 |  |

Table 3.6: The result of merging object A and B from Table 2.6

Likewise, two or more attributes can be merged when they are synonym or hypernym of one another and thereby their object values should be merged. From Table 3.6, if the attributes W and X are a synonym or hypernym of one another, they can be merged together by applying the truth table of Table 3.7 and one of the most general term of W and X would be the name of the new merged attribute. The result of merging attribute W and X from Table 3.6 is shown in Table 3.7.

|  | W,X | Y | Z |
|---|---|---|---|
| A | 1 | 0 | 0 |
| B | 1 | 0 | 1 |
| C | 0 | 1 | 1 |
| D | 0 | 1 | 0 |

Table 3.7: The result of merging attribute W and X from Table 3.6

Incidentally, there are two ways for checking synonyms and hypernyms of objects and attributes as described below:

**1. Single and Dual Methods:** This method checks the objects or attributes of formal context one with another and then combines together. For example, assume that A and B are objects or attributes and represented with their binary values as rows or columns in the formal context respectively. If A is synonym or hypernym of B or vice versa, A and B can be directly combined together irrespective of checking A with the rest of objects or attributes in the formal context. Then, A can be compared with the rest of objects or attributes one by one until it reaches the last object or attribute in the formal context. Figure 3.8 depicts single and dual methods for checking synonyms and hypernyms of objects and attributes in formal context.

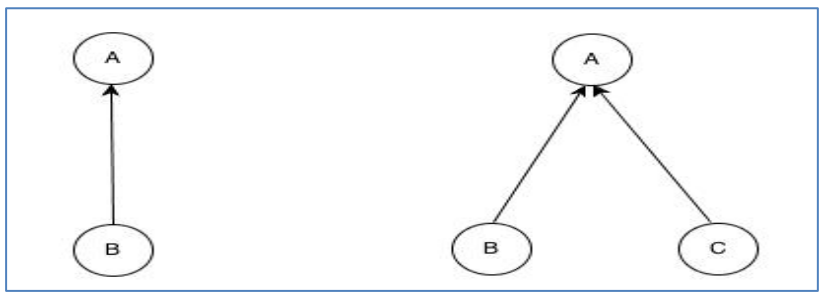

Figure 3.8: Single and dual method for finding synonyms and hypernyms of objects and attributes

For example, if this method is applied on the formal context of Table 3.4, the result would be as below by comparing its objects and attributes.

Comparing the objects with each other is like below:

- *A with B; A with C; B with C; B with D; and C with D*

Likewise, comparing the attributes with each other is like the below trace:

- *W with X; W with Y; W with Z; X with Y; X with Z; and Y with Z*

**2. Multidisciplinary method:** This method checks the objects or attributes of formal context one with all others in the formal context and then combines all together. Multidisciplinary method is more efficient than the single and dual methods. For instance, assume that A is one of the objects or attributes and represented with its binary values as a row or column in the formal context respectively. A can be compared with the rest of objects or attributes in the formal context to find its synonym or hypernym, and then all of them are combined with object or attribute A in one step. Figure 3.8 depicts single and dual methods for checking synonyms and hypernyms of objects and attributes in formal context. Hence, the used method of checking the synonyms and hypernyms of objects and attributes in this project and dissertation is multidisciplinary way because single or dual methods are not so efficient in comparison with the multidisciplinary method.

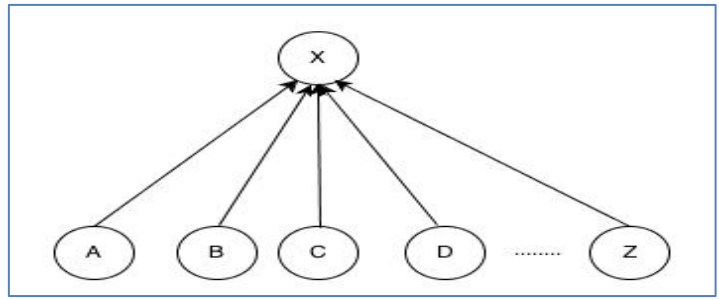

Figure 3.9: A multidisciplinary method for finding synonyms and hypernyms of objects and attributes

For example, if this method is applied on the formal context of Table 3.4, the result would be as below by comparing its objects and attributes.

Comparing the objects with each other is like below:

- *A with B, C, and D; B with C, and D; and C with D*

Similarly, comparing the attributes with each other is like the below trace:

- *W with X, Y, and Z; X with Y and Z; Y with Z*

### 3.4.2 Frequency-based Technique

This technique can be applied on the formal context after its construction to eliminate the objects or attributes that do not happen together with the existing attributes or objects frequently. For example, in table 3.8, A occurs with the attributes of the table three times and occurs with W, Y, and Z. That means, the frequency of A is 75%. Another similar example is the occurrence of attribute W against the objects of the table. Occurrence of attribute W is two times that is with object A and E. therefore, the frequency of attribute W is 40%.

|   | W | X | Y | Z | Frequency |
|---|---|---|---|---|---|
| **A** | 1 | 0 | 1 | 1 | 75% |
| **B** | 0 | 0 | 0 | 1 | 25% |
| **C** | 0 | 0 | 1 | 1 | 50% |
| **D** | 0 | 1 | 0 | 0 | 25% |
| **E** | 1 | 0 | 1 | 0 | 50% |
| Frequency | 40% | 20% | 60% | 60% | |

Table 3.8: An example of formal context with applying frequency-based technique

This technique can be applied, for instance, on the objects and attributes of formal context in Table 3.8 by specifying the threshold to 20%.

|   | W | Y | Z | Frequency |
|---|---|---|---|---|
| **A** | 1 | 1 | 1 | 75% |
| **C** | 0 | 1 | 1 | 50% |
| **D** | 0 | 0 | 0 | 25% |
| **E** | 1 | 1 | 0 | 50% |
| Frequency | 40% | 60% | 60% | |

Table 3.9: The result of Table 3.8 after applying frequency-based technique

## 3.5 Summary

This chapter has mainly focused on automating concept hierarchies from free text by using Formal Concept Analysis and WordNet. At the first section, the needed three steps of automatic ontology generation lifecycle are focused. Next, the overall process of how to learn concepts and concept hierarchies from free text by ten steps has explained. Then, Formal Concept Analysis as one of the ontology learning techniques has defined and detailed by focusing on formal contexts, concepts and concept lattices, and giving an example about applying Formal Concept Analysis on free text. Finally, WordNet-based and Frequency-based techniques as two distinct techniques that can be applied on formal context and used for conducting an experiment as well have demonstrated and detailed via examples.

# 4. Chapter Four: Implementation and Testing

In chapter three, a design phase of the overall process and a method for discovering concepts and concept hierarchies in a free text by using Formal Concept Analysis and WordNet. Likewise, WordNet-based technique and Frequency-based technique have used and investigated as two distinct techniques on formal context via conducting an experiment. Chapter four describes the implementation and testing of the baseline design from chapter three. At the first glance, an implementation of this project is introduced in chapter four. First and foremost, a novel approach is proposed to introduce a framework architecture for the overall process of this project. Second, the modules of this framework are implemented in order. It starts from Natural Language Processing parser, and then the word pairs are extracted and thereby a formal context can be constructed. Correspondingly, an experiment is conducted by using WordNet-based and frequency-based techniques that might result in eliminating uninterested and erroneous pairs and reducing the size of formal context to consume less time for deriving the concept lattice. Afterwards, miscellaneous module as a method of communication among the other modules is detailed. Next, formal concept and lattice compaction are applied by using the Concept Explorer software. Final, an example is applied to clarify the implementation step by step with the experiment in detail. Later in this chapter, the project is tested by using JUnit testing and Big Bang integrated testing.

## 4.1 Framework Architecture for the Overall Process

Figure 4.1 presents the architecture of the framework to be implemented. Implementing the modules begin from the top left-hand side to the right-hand side. These modules are described in the next sections in order.

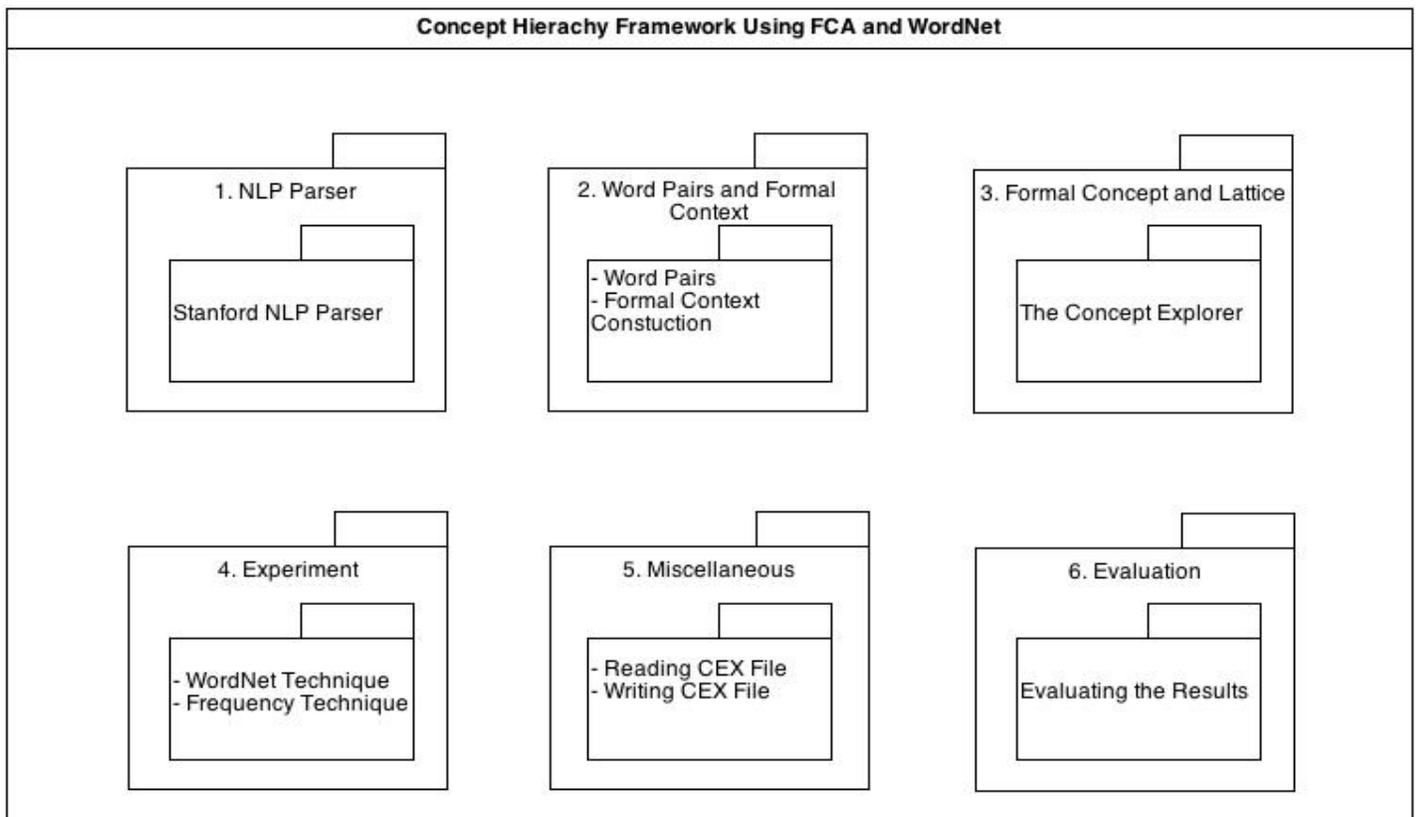

Figure 4.1: An overview of the implemented framework

## 4.2 Natural Language Processing Parser Module

The natural processing language module is used in conjunction with Stanford Natural Language Processing built-in components. The most important components of Stanford Natural Language Processing framework that are used in this project and dissertation are tokenisation, sentence splitter, Part-of-Speech tagger, and lemmatisation. The details are given in the below subsections in regard of these components. Figure 4.2 shows the Natural Language Processing framework used in this project and dissertation comprising of Stanford Natural Language Processing and pair extraction components.

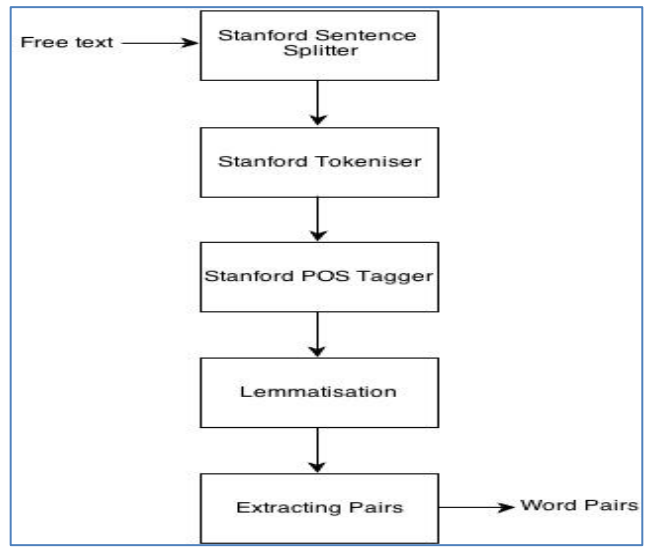
Figure 4.2: Natural Language processing framework used in this project

**4.2.1 Stanford Sentence Splitter**

Sentence splitter is on the first steps of text processing. It tries to find the sentence boundaries from a given chunk of text. Sentences are usually marked by punctuation marks or periods, but these characters can serve for other purposes. Figure 4.3 presents a pseudo code for sentence splitter.

```
1   for a chunk of text or a document
2       if there is a period
3           return the sentence
4       else if there exist full stop and the next token is capitalised
5           return the sentence
6       else if there is a new line and the next token is capitalised
7           return the sentence
8       else
9           do nothing
10      end if
11  end for
```
Figure 4.3: Pseudo code for sentence splitter

Indeed, Stanford sentence splitter splits sentences based on newlines only or treat each document as one sentence. However, standard vanilla approach is needed for use in this project since it has 95% accuracy for splitting sentences from free corpus [43]. This approach locates the end of a sentence and split the free corpus into sentences and a sentence can be ended in the following situations:

- If it is a period
- If there exist full stop and the next token is capitalised
- If there is a new line and the next token is capitalised

**4.2.2 Stanford Tokeniser**

Tokenisation is the process of breaking a stream of texts or a sentence into units called tokens [38]. These tokens are either words, symbols, phrases, numbers, or other meaningful elements in a sentence. Tokenisation typically occurs in word level. Moreover, phrases are detected as tokens from the elements of a sentence. Languages have constraints in word order and thereby words in languages are grouped and organised in units called phrases, such as noun and verb phrases [38]. Noun phrase (NP) is a syntactic unit of a sentence where information about the noun is gathered. Noun phrases usually consist of zero or more than one adjectives. On the contrary, verb phrase (VP) is a syntactic unit of a sentence where information about the verb is gathered. In a verb phrase, the elements of verb phrase are words that syntactically depend upon the verb. Hence, phrase is usually detected as a token and activated by rules which define patterns of Part-of-Speech Tags. Using statistical measures where a phrase is defined as a sequence of words which co-occur frequently is another approach. Figure 4.4 presents how Stanford tokeniser works.

```
1    for each sentence in a text corpus
2        for each token in the sentence
3            retrieve token and its type
4        end for
5    end for
```

Figure 4.4: Pseudo code for sentence tokenisation

**4.2.3 Stanford Part-of-Speech Tagger**

Determination of the Part-of-Speech of a word is an essential precursor to several Natural Language Processing (NLP) tasks. POS-Tagged is a process used as a reasonable compromise between accuracy and utility [39]. It consists of labelling each word in a sentence with its appropriate part of speech, such as nouns, verbs, adjectives, and prepositions. This process is based upon a predefined tagging scheme. Training the tagger to use an already tagged document collection or corpus is the most common scheme of tagging. This could resolve word ambiguity and it may appear using an approach, such as looking up a word in a dictionary to find its POS Tagger since the word may belong to several morphological classes. Thus, statistical and rule-based models are the most common approaches for training a tagger. Figure 4.5 shows how Stanford POS Tagger works.

```
1    for each word in a sentence or text
2        if the word is not phrase
3            put lable on the word
4        else if
5            put the phrase type on the word
6        end if
7    end for
```

Figure 4.5: Pseudo code for Part-of-Speech Tagger

**4.2.4 Lemmatisation**

After the Stanford POS Tagger, the lemmatization sub-module is responsible for finding the lemma of each word. The lemma of a word is usually found by looking up the word in a lemma list that includes a list of words and their associated lemma. This could be achieved by looking a given word in a lemma list or dictionary. For example, the lemma of the word "drove" is "drive". This is an essential in text mining and information extraction because "talked" and "talk" would be considered as two different words without the lemmatisation process. Alternatively, stemming process is essential. This process is usually done by using rules. For instance, both "manager" and "management" can be reduced to the stem "manage". Therefore, both lemmatisation and stemming may be preferable for in the use of this project and dissertation, but stemming is difficult to derive the original word from a stem and Stanford Natural Language Processing framework does not support stemming. Figure 4.6 depicts a sequence diagram of how EnglishLemmatiser class finds a lemma of a word.

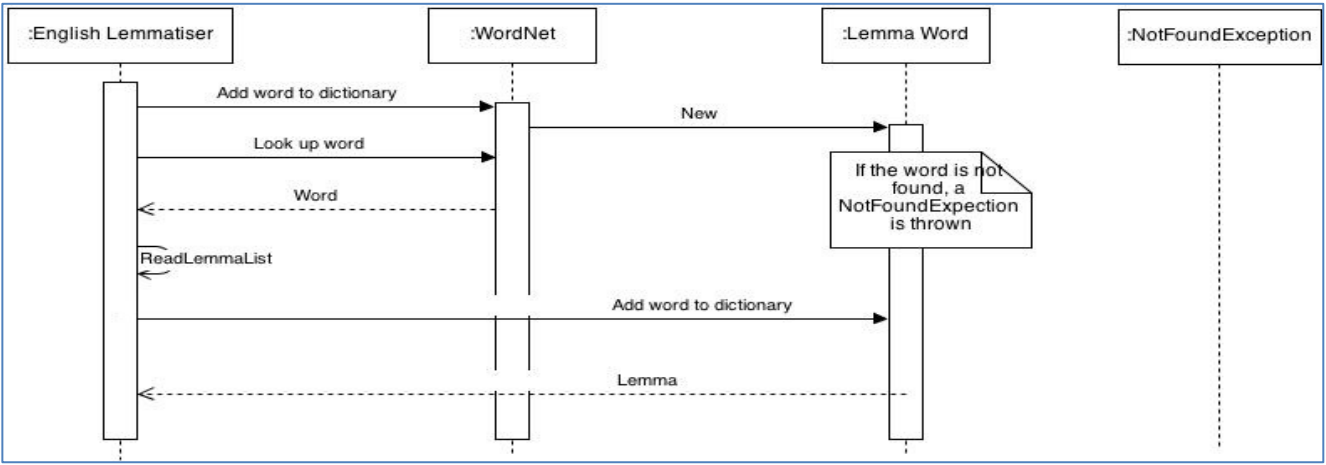

Figure 4.6: Sequence diagram for getting word's lemma

**4.2.5 Word Pairs Extraction**

After the lemmatisation step, the tokens can be paired together in a way that syntactic dependencies between the verbs appearing in the text collection and the heads of the subject, object and PP-components are used so as to derive context attributes describing the terms. In order for these dependencies to be extracted automatically, the text is parsed by using Stanford Natural Language Processing framework, a set of natural language analysis tool. The syntactic dependencies are extracted from the verbs as well as the head of the subject, object, and PP-component by looking up the lemma in the lexicon provided by that tool since lemmatisation maps a word to its root and used as a sort of normalisation of the text. Figure 4.7 shows a pseudo code of how extracting word pairs works.

```
1   for each term in the setence or text corpus
2       if it occurs with the another term
3           add it pairsList in the form of verb - noun (pair)
4       else
5           search if it occurs with the other terms
6       end if
7   end for
```

Figure 4.7: Pseudo code for extracting word pairs

The scenario of extracting pairs can be explained by the below example. Let us take the following below two sentences:

*A museum is an institution that cares for (conserves) a collection of artefacts and other objects of scientific, artistic, cultural, or historical importance and makes them available for public viewing through exhibits that may be permanent or temporary. Most large museums are located in major cities throughout the world and more local ones exist in smaller cities, towns and even the countryside. The continuing acceleration in the digitization of information, combined with the increasing capacity of digital information storage, is causing the traditional model of museums to expand to include virtual exhibits and high-resolution images of their collections for perusal, study, and exploration from any place with Internet. The city with the largest number of museums is Mexico City with over 128 museums. According to The World Museum Community, there are more than 55,000 museums in 202 countries.*

After parsing these two sentences, the following below dependencies are extracted in the form of verb - noun:

*museum-institution, be-institution, institution-care, institution-make, care-institution, object-artefact, artistic-scientific, cultural-scientific, importance-scientific, make-institution, make-care, they-view, view-make, exhibit-permanent, exhibit-temporary, be-permanent, permanent-exhibit, temporary-exhibit, temporary-permanent, museum-located, one-world, town-city, countryside-city, acceleration-cause, acceleration-expand, combine-acceleration, model-cause, expand-cause, include-expand, exhibit-include, image-include, image-exhibit, study-perusal, exploration-perusal, city-City, be-City, to-be, museum-be*

In Natural Language Processing module, *Lemmatizer and WordPair* Java Classes are used.

### 4.3 Word Pairs and Formal Context Construction Module

This module mainly focuses on extracting the interested and wanted word pairs by filtering and pruning them to be used for constructing formal context.

#### 4.3.1 Pruning Word Pairs

All of the pairs should be in the form of verb-noun, but the extracted pairs are sometimes not in that form. In this case, a technique is needed to opts to prune weights off and change them to the form of verb -noun. For example, driving - car pair might not be appropriately useful for to be represented in the next step and accordingly it should be pruned into a verb - noun format as drive - car. Figure 4.8 presents the pseudo code of how to prune word pairs.

```
1   for each word-pair of the extracted pairs from the text
2       if a word-pair is in the form of verb - noun root
3           remove their suffix
4           get their roots
5       else
6           do nothing
7       end if
8   end for
```

Figure 4.8: Pseudo code for pruning word pairs

Referring to the example in section 4.2.5, the derived pairs can be pruned if they need so. In that example, the pairs would remain the same as presented before.

### 4.3.2 Filtering Word Pairs

Some of the constructed pairs are erroneous and may not be interested and a technique is needed to filter the word pairs based on the necessity of this project and dissertation. The current Stanford Dependencies include approximately fifty grammatical relations [42]. These dependencies are binary relations; a grammatical relation holds between a governor and a dependent and can be used for representing the word pairs in the previous step. However, they need to be filtered in order to opt for the needed and interested grammatical relations for applying formal contexts and concepts, and context lattice are filtered. Figure 4.9 presents the pseudo code of how to filter word pairs.

```
1   for each word-pair of the extracted pairs from the text
2       if word-pair in the interested pair list
3           put on the interested pairs
4       else
5           ignore it
6       end if
7   end for
```

Figure 4.9: Pseudo code for filtering word pairs

Table 4.1 lists the needed and interested Stanford typed dependencies for applying Formal Concept Analysis used for this project and dissertation.

| Dependency type | Meaning | Description |
|---|---|---|
| acomp | Adjectival complement | An adjectival component of a verb |
| agent | Agent | a component of a passive verb |
| conj | Conjunct | A relation between two elements that connected by a coordination conjunction |
| cop | Copula | The relation between the components of a copular verb and the copular verb |
| csubj | Clausal subject | Clausal syntactic subject of a clause |
| csubjpass | Clausal passive subject | A clausal syntactic subject of a passive clause |
| dobj | Direct object | The direct object of a VP is the noun phrase which is the object of the verb |
| infmod | Infinitival modifier | An infinitival modifier of an NP is an infinitive that serves to modify the meaning of the NP |
| nsubj | Nominal subject | A noun phrase which is the syntactic subject of a clause |
| nsubjpass | Passive nominal subject | A noun phrase which is the syntactic subject of a passive clause |
| parataxis | Parataxis | A relation between the main verb of a clause and other sentential elements |
| partmod | Participial modifier | A Participial modifier of an NP or VP or sentence is a participial verb form that serves to modify the meaning of a noun phrase or sentence |
| prepc | Prepositional clausal modifier | Prepositional clausal modifier of a verb, noun, or adjective is a clause introduced by a preposition which serves to modify the meaning of the verb, noun , or adjective |
| purpcl | Purpose clause modifier | Purpose clause modifier of a VP is a clause headed by "(in order) to" specifying a purpose |
| rcmod | Relative clause modifier | A relative clause modifier of an NP is a relative clause modifying the NP |
| rel | Relative | A relative of a relative clause is the head word of the WH-phrase introducing it |
| tmod | Temporal modifier | A temporal modifier of a VP, NP, or ADJP is a bare noun phrase constituent that serves to modify the meaning of the constituent by specifying a time |
| xcomp | Open clausal component | An open clausal component of a VP or an ADJP is a clausal component without its own subject |
| xsubj | Controlling subject | The relation between the head of an open clausal component and the external subject of that clause |

Table 4.1: Interested Stanford typed dependencies manual

Referring to the example in section 4.2.5, the derived pairs can be filtered based on our needed pairs. As a result, the following below interested pairs are extracted:

*museum-institution, be-institution, institution-care, institution-make, care-institution, object-artefact, make-institution, make-care, view-make, exhibit-permanent, be-permanent, permanent-exhibit, one-world, town-city, countryside-city, acceleration-cause, combine-acceleration, model-cause, image-exhibit, study-perusal, exploration-perusal, city-City, be-City, museum-be*

### 4.3.3 Formal Context Construction

After filtering and pruning techniques on the extracted word pairs, the word pairs can be represented in a form of table as it is the simplest format for manipulation and writing down a formal context. The objects (nouns) can be corresponding to the rows of the formal context, while the attributes (verbs) can be corresponding to the columns of the formal context. The cells (x, y) of the formal context represent Booleans values wherever object x has value y. For the sake of clarity, an example is given below by referring the extracted pairs in section 4.3.2 and these pairs can be represented in a formal context as follows:

| A | B | C | D | E | F | G | H | I | J | K | L | M |
|---|---|---|---|---|---|---|---|---|---|---|---|---|
|   | be | care | object | make | view | exhibit | combine | model | expand | include | image | study |
| institution | X | X |   |   |   |   |   |   |   |   |   |   |
| care |   |   |   | X |   |   |   |   |   |   |   |   |
| make |   |   |   |   | X |   |   |   |   |   |   |   |
| artifact |   |   | X |   |   |   |   |   |   |   |   |   |
| view |   |   |   |   |   |   |   |   |   |   |   |   |
| permanent | X |   |   |   |   | X |   |   |   |   |   |   |
| exhibit |   |   |   |   |   |   |   |   |   |   | X |   |
| world |   |   |   |   |   |   |   |   |   |   |   |   |
| city |   |   |   |   |   |   |   |   |   |   |   |   |
| cause |   |   |   |   |   |   |   | X | X |   |   |   |
| acceleration |   |   |   |   |   |   | X |   |   |   |   |   |
| perusal |   |   |   |   |   |   |   |   |   |   |   | X |
| City | X |   |   |   |   |   |   |   |   |   |   |   |
| be |   |   |   |   |   |   |   |   |   |   |   |   |

Table 4.10: Formal context construction example

Incidentally, *Stemmer and WordPair* Java classes are used for the implementation of this module

### 4.4 Formal Concept and Lattice Compaction Module

This module includes the use of the Concept Explorer as open source software for implementing the basic ideas needed for the research and study of Formal Concept Analysis (FCA) and lattice compaction. In turn, this module is completely done by using the Concept Explorer by reopening and saving CEX file. Figure 4.11 depicts a sequence diagram of how the Concept Explorer uses word pairs for deriving concept lattice

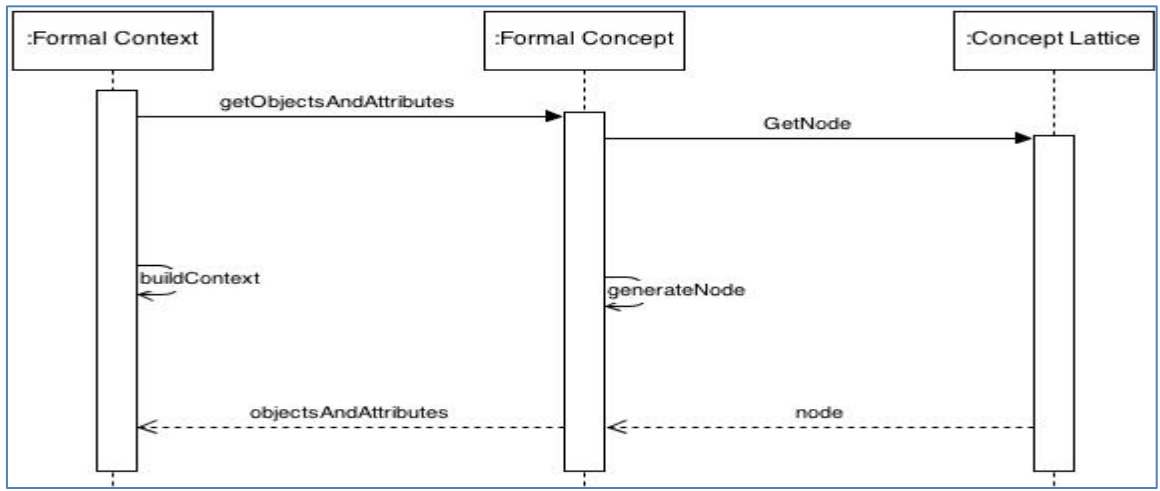

Figure 4.11: Sequence diagram of how the Concept Explorer uses word pairs for deriving concept lattice

#### 4.4.1 Working on Formal Concept

The Concept Explorer can represent the structure of finite formal context and concept lattice. This is a graph with nodes and edges connecting them as well as ordered structure with a bottom and a top element. Each node in a concept lattice corresponds to a formal concept. This can be denoted as a pair (O, A), where A is a subset of the attribute set and some additional properties are satisfied, and O is a subset of the object set. On the other hand, the Concept Explorer allows several different formats, but the recommended storage format by the Concept Explorer is native ConExp format, which is CEX. In addition, information about the context and the lattice line diagram can be stored in CEX file. Figure 4.8 depicts a sequence diagram of how the Concept Explorer uses formal context to build formal concepts and thereby its lattice. Further, the Concept Explorer supports undo/redo, changing the size of a context, compressed view of a context, visualisation of the arrow relations, and entering data into the context. On the basis of main operations with contexts, five operations can be performed on contexts: object and attribute clarification, object and attribute set reduction, context reduction, and transposition.

#### 4.4.2 Concept Lattices and Line Diagram

The formal context includes information about how attributes and objects should be related and this can derive the set containing all attributes common to all objects from the set O and only these attributes. For drawing the concept lattice, labelling is used so as to present information about extents and intents of formal context concisely. For example, each node x of the structure represents a formal concept (O, A) and the extent O of this concept can be represented a formal context (O, A) and the extent O of this concept can be received collecting all objects reachable attached to nodes by descending paths from node x to the bottom element of the lattice. Oppositely, the intent A of this concept can be received collecting all attributes attached to nodes reachable by ascending paths from this node x to the top element of the lattice. In the Concept Explorer, Build Lattice button can be used on the main toolbar so as to derive the concept lattice as a straight line diagram from the formal context. This process might be time consuming, but it depends on the complexity of the data. Figure 4.12 presents the concept lattice of formal context of Table 4.2.

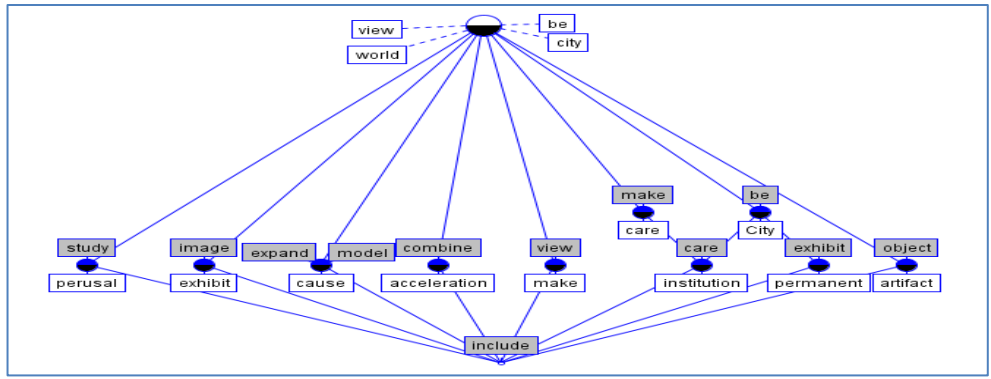

Figure 4.12: Lattice and line diagram

**4.5 Experiment Module**

After the construction of formal context, there may be still some erroneous and uninteresting pairs in it because the output of the parser can be erroneous and no all the derived pairs are interesting. In the meantime, the size of formal context might be substantially large due to constructing the formal context from a large free text corpus. In the Concept Explorer, deriving the concept lattice from the formal context might be a time consuming processing and it depends on the size and complexity of the formal context data [44]. In that sense, this experiment might result in eliminating uninterested and erroneous pairs and reducing the size of formal context to consume less time for deriving concept lattice. In turn, the following questions need to be asked:

- Does reducing the size of formal context require linguistic resources, such as WordNet or/and statistical approaches?
- Are linguistic resources more effective than statistical approaches in reducing the size of formal context or vice versa?
- If both linguistic resources and statistical approaches need to reduce the size of formal context? In which order should they apply on the formal context to be more effective?
- Can the size of formal context be reduced without affecting the quality of the result?

To answer these questions and deal with these issues, this experiment is basically conducted to deal with reducing the size of the formal context as well as eliminating erroneous and uninterested pairs in the formal context by using two distinct techniques which are WordNet-based technique and Frequency-based technique. These two techniques are investigated for eliminating the erroneous and uninteresting pairs and their terms. The First technique deals with comparing the pairs with the WordNet as a lexical database for English language, whereas the second technique weights the pairs based on some statistical measure and thereby only the pairs over a certain threshold are transformed into a formal context to which Formal Concept Analysis is applied. In addition, WordNet-based technique is a linguistic resource-based technique for the English language that can be used as a lexical database for comparing the objects and attributes of formal context to check whether they are synonyms or hypernyms of each other in order to be combined together. On the contrary, frequency-based technique is a statistical based technique than can be used to eliminate the least frequently objects or attributes with their corresponding attributes and objects respectively. As shown in the framework of Figure 4.13, formal concept and lattice compaction can be directly applied on the extended CEX file so as to extract the concept lattice 1. On the other hand, to extract the concept lattice 2, the formal context should be read from the extended CEX file, and then the two techniques are applied on it, and then it should be written on a CEX file. Finally, the formal concept and Lattice could be applied on it so as to extract the concept lattice 2. Meanwhile, CEX file can be used in the Concept Explorer to yield formal contexts and concepts, and concept lattice. As a result of this experiment, the questions might be answered as the resulted concept lattices are compared and evaluated to answer that does the size of the formal context affect the quality of the result and the other above questions as well.

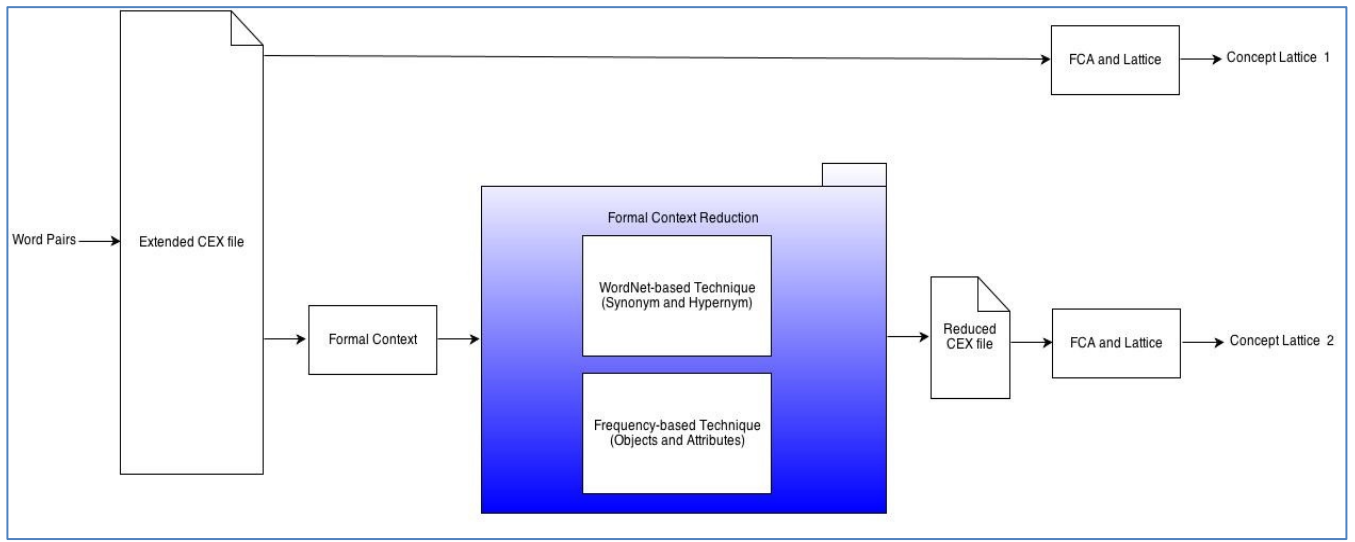
Figure 4.13: Framework of the experiment

In regard with the order of applying these techniques on the formal context, it is not clear yet whether applying WordNet-based technique or Frequency-based technique at first time is more efficient in reducing the size of formal context. WordNet-based technique for finding synonyms and hyponyms to reduce the size of the formal context can be applied first, and then the frequency-based technique. The order of applying these two techniques might be due to the fact that some of the least frequently pairs in the formal context might be a synonym or hyponym of some other pairs and they could be combined with the most frequently pairs accordingly. In addition, applying the WordNet technique might substantially reduce the formal context size. Therefore, applying the frequency-based technique after the WordNet techniques can be more efficient than applying it before the WordNet technique. Further, WordNet-based technique may help the frequency-based technique in the sense that to what extent these pairs are intuitively usable. On the other hand, Frequency-based technique can be applied first, and then the WordNet-based technique. The order of applying these two techniques might be due to the fact that Frequency-based technique can eliminate the erroneous and uninteresting pairs in the formal context, and then applying WordNet-based technique might be more efficient. In brief, the order of applying these techniques has not practiced and justified yet. On that basis, this is investigated in the next chapter.

**4.5.1 WordNet-based Technique**
WordNet-based technique is a linguistic resource-based technique for the English language that can be used as a lexical database dictionary for comparing the objects and attributes of formal context to check whether they are synonyms or hypernyms of each other in order to be combined together. This technique deals with comparing each object or attribute with all of the other objects or attributes respectively in the formal context to check whether are synonym or hypernym of each other or not. Figure 4.14 gives a pseudo code about applying this technique on the formal context.

```
1   for each object(attribute) in the formal context
2       compare the object(attribute) with the rest of objects(attributes) in the formal context
3       if the object(attribute) has synonyms
4           merge their rows(colums) together in the formal context
5       else if the object(attribute) has hypernyms
6           choose the most general word among the hypernyms objects((attributes)
7           merge their rows(columns) together in the formal context
8       else
9           do nothing
10      end if
11  end for
```
Figure 4.14: Pseudo code for WordNet-based technique

In the formal context, there may be a number of objects or attributes that are hypernym of each other, but depth of their semantic relations could be minimum or maximum. If there semantic relations are minimum, their type-of relationships are closer than the maximum semantic relations. For example, carmine, scarlet, and vermilion are all hyponyms of red, which is, in turn, a hyponym of colour. Accordingly, the depth between carmine and red is closer than the depth between carmine and colour. Practically, the hypernym depth between two objects and attributes are applied in this project and dissertation. The depth can be specified between two hypernym words, and then the most general term is retained. Referring to the formal context shown in the table 4.2, WordNet-based technique can be applied on both its objects and attributes. When the depth of objects and attributes is specified by 4, the result is as shown in Table 4.3.

|   | A | B | C | D | E | F |
|---|---|---|---|---|---|---|
|   |   | care | object | exhibit | image | study |
| make |   |   | × |   |   |   |
| artifact |   | × | × | × | × |   |
| view |   |   |   |   |   |   |
| permanent |   |   | × | × |   |   |
| exhibit |   |   |   |   | × |   |
| city |   |   | × |   |   |   |
| acceleration |   |   |   | × |   |   |
| perusal |   |   |   |   |   | × |
| be |   |   |   |   |   |   |

Table 4.3: Formal context construction after applying WordNet-based technique

**4.5.2 Frequency-based Technique**

Frequency-based technique is a statistical based technique that can be used to eliminate the least frequent objects or attributes with their corresponding attributes and objects respectively. This technique deals with the frequency of each object and attribute in occurrence with each attribute and object respectively and can be applied on the formal context after its construction to eliminate the objects or attributes that do not occur together with the existing attributes or objects frequently. The least frequent objects and attributes can be eliminated in the formal context based on a threshold. The threshold is about specifying the percentage of how frequently and object or attribute occur in correspondence with its values of attributes or objects respectively. Figure 4.15 presents the pseudo code of frequency-based technique that explains the way of eliminating objects and attributes in the formal context which are less than or equal to the threshold.

```
1    for each object(attribute) in the formal context
2        if the occurence of object(attribute) is less than or equal the threshold
3            merge the row(colum) of that object(attribute) in the formal context
4        else
5            do nothing
6    end for
```

Figure 4.15: Pseudo code for Frequency-based technique

Referring to the formal context shown at Table 4.2, Frequency-based technique can be applied on both its objects and attributes with their occurrence of corresponding attributes and objects respectively. When threshold is specified by 5%, the result shown in Table 4.4 can be presented as concept lattice accordingly as shown in Figure 4.16 (right).

|   | A | B | C | D | E | F | G | H | I | J | K | L |
|---|---|---|---|---|---|---|---|---|---|---|---|---|
|   | be | care | object | make | view | exhibit | combine | model | expand | image | study |
| institution | × | × |   |   |   |   |   |   |   |   |   |   |
| care |   |   |   | × |   |   |   |   |   |   |   |   |
| make |   |   |   | × |   |   |   |   |   |   |   |   |
| artifact |   |   |   |   | × |   |   |   |   |   |   |   |
| permanent | × |   |   |   |   |   | × |   |   |   |   |   |
| exhibit |   |   |   |   |   |   |   |   |   | × |   |   |
| cause |   |   |   |   |   |   |   | × | × |   |   |   |
| acceleration |   |   |   |   |   |   | × |   |   |   |   |   |
| perusal |   |   |   |   |   |   |   |   |   |   |   | × |
| City |   | × |   |   |   |   |   |   |   |   |   |   |

Table 4.4: Formal context construction after applying Frequency-based technique

Alternatively, if both WordNet-based and Frequency-based techniques are applied on the formal context of table 4.2 by specifying 5% as the threshold and 4 as the depth of hypernym, the result would be as shown in Table 4.5 and the lattice shown in Figure 4.16 (left) is produced accordingly.

|   | A | B | C | D | E | F |
|---|---|---|---|---|---|---|
|   |   | care | object | exhibit | image | study |
| make |   |   | × |   |   |   |
| artifact |   | × | × | × | × |   |
| permanent |   |   | × | × |   |   |
| exhibit |   |   |   |   | × |   |
| city |   |   | × |   |   |   |
| acceleration |   |   |   | × |   |   |
| perusal |   |   |   |   |   | × |

Table 4.5: Formal context construction after applying both techniques

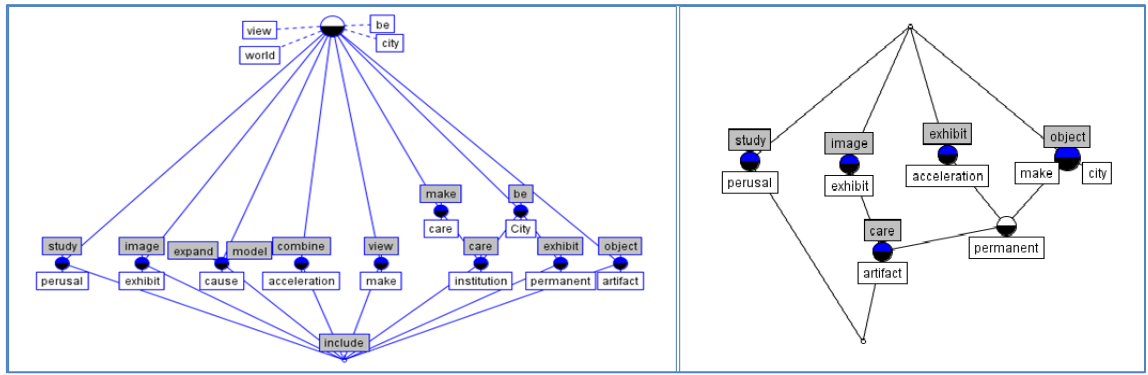

Figure 4.16: Concept lattice of Table 4.4 (left) and Table 4.5 (right)

The following Java classes are contained in the experiment module for both WordNet-based and Frequency-based techniques:

*FreqBasedTech, WordNetBasedTech, CEXReaderTest, CEXWriterTest, WordNetHelper, SynReplace, TwoDimentionalArrayList, TwoDimentionalArrayListUtil, Lemmatiser, Stemmer, and WordPair*

## 4.6 Miscellaneous Module

This module is an XML-based representation of formal context used as a method of communication among the other modules. It uses CEX file as an XML-based format as a widely used representation of arbitrary data structures among the other modules to emphasise generality, simplicity, and usability over all of the modules. Figure 4.17 gives the details of how this module as a middleware interconnects the other modules of the proposed framework. The Concept Explorer allows several different formats, but the recommended storage format by the Concept Explorer is native ConExp format, which is CEX. CEX is an XML-based format that stores information about the context and the lattice line diagram. In addition, the Concept Explorer allows saving and re-opening documents by selecting on the items in the sub-menu files and information about the context and the lattice line diagram are stored in both extended CEX and CEX files. Meanwhile, CEX file can be read by the Concept Explorer in order to analyse attribute-object tables which are called formal context in Formal Concept Analysis, draw the corresponding lattice and to explore different dependencies that may exist between attributes.

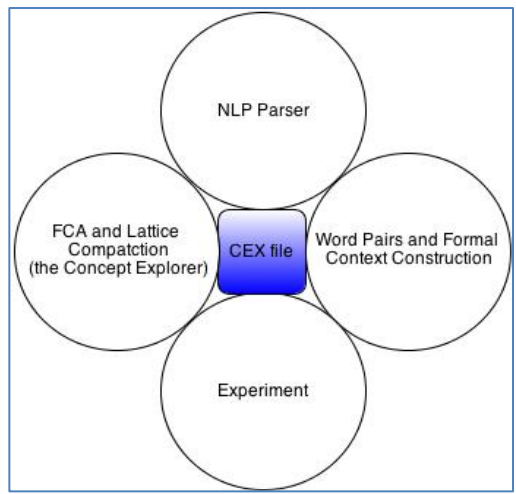

Figure 4.17: CEX file format as a middleware among the modules

The modules of the framework can be connected to each other systematically by using CEX as a middleware among all of them. This standard file format can be deployed by all of the modules because CEX is an XML-based format, but not always all CEX-based formats look the same as XML-based format in. This is also a way for connecting Natural Processing Language parser, and FCA and lattice compaction modules to the Concept Explorer and acts as a base-step for conducting the experiment. Further, the Concept Explorer allows CEX file formats as the native and recommended storage format by the Concept Explorer is native ConExp format, which is CEX. CEX is an XML-based format that stores information about the context and the lattice line diagram. It also needs to mention that the Concept Explorer allows saving and re-opening documents by selecting on the items in the sub-menu files. Figure 4.18 shows a sample of CEX file by XML syntax.

```xml
<?xml version="1.0" encoding="UTF-8" standalone="no"?><ConceptualSystem><Version MajorNumber="1" MinorNumber="0"/>
<Contexts><Context Identifier="0" Type="Binary">
<Attributes>
<Attribute Frequency="1" Identifier="0"><Name Identifier="0">building</Name></Attribute>
<Attribute Frequency="0" Identifier="1"><Name Identifier="1">reference</Name></Attribute>
<Attribute Frequency="1" Identifier="2"><Name Identifier="2">allude</Name></Attribute>
</Attributes>
<Objects>
<Object Frequency="0"><Name>hold</Name><Intent/></Object>
<Object Frequency="1"><Name>combine</Name><Intent><HasAttribute AttributeIdentifier="0"/></Intent></Object>
<Object Frequency="1"><Name>reference</Name><Intent><HasAttribute AttributeIdentifier="2"/></Intent></Object>
</Objects>
</Context></Contexts><RecalculationPolicy Value="Clear"/><Lattices/></ConceptualSystem>
```

Figure 4.18: Sample of CEX file by XML syntax

There are two versions of CEX file in the framework. The first one can be called as extended CEX file and the second one is the one that the two techniques applied on it. Incidentally, *CEXReader* and *CEXWriter* Java classes are used in the miscellaneous module.

**4.7 Applied Example**

To clarify the implementation steps with the experiment in detail, this example is taken. In this example, threshold is assigned to 5% and the depth of hypernym is assigned to 4. This example can be explained step by step by the following below five points:

**1. Text Corpus as Input Source:** A free text is used in this example. Let us take the following below text corpus which is about British Museum. This is saved in a text file.

*The British Museum is a museum in London dedicated to human history and culture. Its permanent collection, numbering some 8 million works, is among the largest and most comprehensive in existence and originates from all continents, illustrating and documenting the story of human culture from its beginnings to the present. The British Museum was established in 1753, largely based on the collections of the physician and scientist Sir Hans Sloane. The museum first opened to the public on 15 January 1759 in Montagu House in Bloomsbury, on the site of the current museum building. Its expansion over the following two and a half centuries was largely a result of an expanding British colonial footprint and has resulted in the creation of several branch institutions, the first being the British Museum (Natural History) in South Kensington in 1887. Some objects in the collection, most notably the Elgin Marbles from the Parthenon, are the objects of intense controversy and of calls for restitution to their countries of origin. Until 1997, when the British Library (previously centred on the Round Reading Room) moved to a new site, the British Museum was unique in that it housed both a national museum of antiquities and a national library in the same building. The museum is a non-departmental public body sponsored by the Department for Culture, Media and Sport, and as with all other national museums in the United Kingdom it charges no admission fee, except for loan exhibitions. Since 2002 the director of the museum has been Neil MacGregor.*

**2. Natural Processing Language Parser:** After parsing the sentences of the above text and thereby extracting the syntactic dependencies, the following below word pairs are extracted:

*museum-museum, be-museum, dedicate-London, culture-history, collection-be, most-largest, originate-existence, illustrate-largest, document-largest, document-illustrate, story-illustrate, museum-establish, on-establish, sloane-physician, museum-open, January-public, expansion-result, half-two, be-result, result-footprint, be-museum, museum-result, object-object, be-object, of-object, Library-move, museum-unique, be-unique, it-house, house-unique, museum-house, library-house, library-museum, museum-body, be-body, sponsor-body, Department-sponsor, Media-culture, Sport-culture, it-charge, charge-body, fee-charge, for-charge, director-MacGregor, be-MacGregor, MacGregor-2002*

**3. Constructing Formal Context:** Firstly, the extracted pairs can be pruned and filtered. The following below syntactic dependencies are the filtered and pruned pairs:

*be-museum, dedicate-London, originate-existence, be-result, result-footprint, be-museum, object-object, be-object, be-body, sponsor-body, Sport-culture, charge-body, fee-charge, be-MacGregor*

Subsequently, formal context is extracted from the filtered and pruned word pairs as shown in Table 4.6

|   | A | B | C | D | E | F | G | H | I | J | K | L | M |
|---|---|---|---|---|---|---|---|---|---|---|---|---|---|
|   |   | be | dedicate | originate | illustrate | document | result | object | house | sponsor | Sport | charge | fee |
| museum |   | X |   |   |   |   |   |   |   |   |   |   |   |
| London |   |   | X |   |   |   |   |   |   |   |   |   |   |
| history |   |   |   |   |   |   |   |   |   |   |   |   |   |
| be |   |   |   |   |   |   |   |   |   |   |   |   |   |
| existence |   |   |   | X |   |   |   |   |   |   |   |   |   |
| physician |   |   |   |   |   |   |   |   |   |   |   |   |   |
| open |   |   |   |   |   |   |   |   |   |   |   |   |   |
| public |   |   |   |   |   |   |   |   |   |   |   |   |   |
| result |   | X |   |   |   |   |   |   |   |   |   |   |   |
| two |   |   |   |   |   |   |   |   |   |   |   |   |   |
| footprint |   |   |   |   |   |   |   | X |   |   |   |   |   |
| object |   | X |   |   |   |   |   |   | X |   |   |   |   |
| move |   |   |   |   |   |   |   |   |   |   |   |   |   |
| house |   |   |   |   |   |   |   |   |   |   |   |   |   |
| body |   | X |   |   |   |   |   |   |   |   | X |   | X |
| sponsor |   |   |   |   |   |   |   |   |   |   | X |   |   |
| culture |   |   |   |   |   |   |   |   |   |   |   |   |   |
| charge |   |   |   |   |   |   |   |   |   |   |   |   | X |
| MacGregor |   | X |   |   |   |   |   |   |   |   |   |   |   |

Table 4.6: Extracted formal context from

**4. Formal Concept and Lattice Compaction:** After extracting the formal context from the word pairs, a lattice diagram can be derived from it as shown in Figure 4.19.

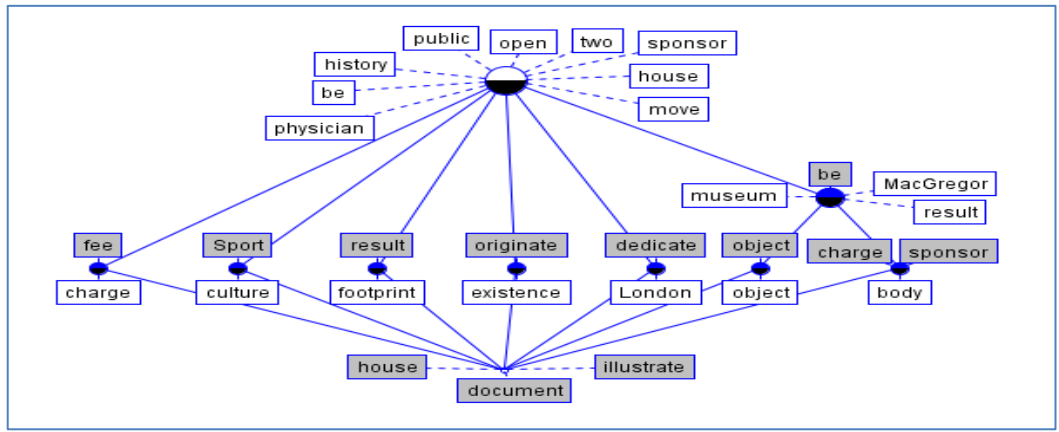

Figure 4.19: Concept lattice of Table 4.6

**5. Experiment:** This experiment basically aims to reduce the size of the formal context and eliminate erroneous and uninterested pairs in the formal context by using an investigating two distinct techniques together and separately as well. By doing so, the reduced concept lattice of Figure 4.20 is resulted from applying both WordNet-based and Frequency-based techniques.

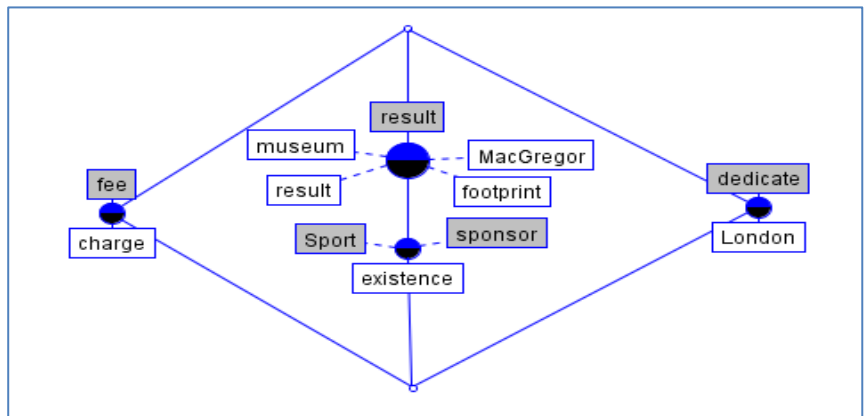

Figure 4.20: Reduced Lattice by applying both techniques

On the other hand, Figure 4.21 (left) is resulted from applying only WordNet-based technique, while Figure 4.21 (right) is resulted from applying Frequency-based technique.

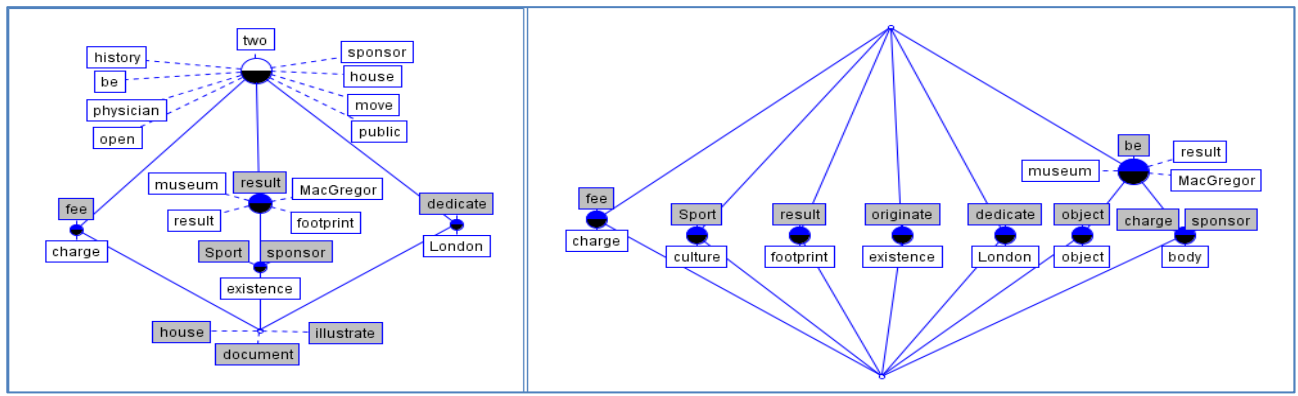
Figure 4.21: Reduced concept lattices by applying WordNet-based technique (left) and Frequency-based technique (right)

## 4.8 Testing

Testing an application or a project is an integral part of the development cycle. With the use of this project and dissertation, a set of programs together with associated control data, operating procedures, and usage procedures have determined to ensure that they are fit with it. In this project and dissertation, two types of testing have used: firstly, unit testing has used to test each components of the framework; secondly, an integrated testing has used to test the whole framework. These two types of testing and the way of applying them on the system are detailed below two sub-sections:

### 4.8.1 Unit Testing

Writing and maintaining unit tests can help in ensuring about individual classes and methods in the project work feasibly correct [45] [46].The integration of Netbeans IDE with the JUnit unit testing framework enables creating JUnit tests and test suites easily and quickly. JUnit as an instance of the xUnit architecture for unit testing frameworks is a framework to write repeatable tests that deals with modules of the framework individually and tests and test suites for the java classes and methods of the project have created. Moreover, there are two versions of JUnit test, which are JUnit 3 and JUnit 4. Both of them suits for testing Java class library project. Netbeans IDE as a used tool in this project provides JUnit 3 and 4 tests. For this project and dissertation, all of the modules of the framework have tested in JUnit 4 by using NetBeans IDE.

Incidentally, the following JUnit tests classes are used for testing the classes and methods of this project:

> *CEX ReaderTest, CEX WriterTest, WordNetBasedTechTest, FreqBasedTechTest, WordNetHelperTest, SynReplaceTest, TwoDimentionalArrayListTest, TwoDimentionalArrayListUtilTest, LemmatiserTest, FrameworkComponentsTest, StemmerTest, and WordPair*

### 4.8.2 Integrated Testing

The modules of the framework have combined and tested as a group to verify functional, reliability requirements, and performance to test the whole program at once. There are several different types of integrated testing. Big Bang integrated testing is the most proper one among the others because, in this approach, almost all the developed modules can be coupled together to form the complete software system, and then used for integration testing. Big Bang integrated testing is successful for saving time and small to medium size project [47] [48]. This method of integrated testing is highly effective for saving time in the integrated testing process, but the test result should be recorded properly. Otherwise, the entire integrated test process will probably be complicated and may prevent achieving the goal of integration testing. As already mentioned earlier, Big Bang testing is useful for small to medium size programs as other approaches require a lot more overload work to be able to test a program in small subsections. Even though the other approaches may provide more insight as to where do exactly the bugs and faults occur, they most likely out ways their benefits on a small or medium project. Thus, Big Bang testing is an appropriate type of integrated testing for this project and dissertation. The Big Bang integrated testing approach on this project and dissertation is outlined in Table 4.7. This testing does not require Graphical User Interface (GUI). In the appendix G, the input and expected output of the module of this project and dissertation are outlined.

| Modules | Test to Perform | Expected Result |
|---|---|---|
| CEXReader | To read CEX XML-based format file properly | Pass |
| CEXWriter | To write CEX XML-based format file properly | Pass |
| WordNetBasedTech | To implement WordNet-based technique properly | Pass |
| FreqBasedTech | To implement Frequency-based technique properly | Pass |
| WordNetHelper | To help in implementing WordNet-based technique properly | Pass |
| SynReplace | To help in implementing WordNet-based technique properly | Pass |
| TwoDimentionalArrayList | To help in implementing WordNet-based and Frequency-based techniques properly | Pass |
| TwoDimentionalArrayListUtil | To help in implementing WordNet-based and Frequency-based techniques properly | Pass |
| Lemmatiser | To implement word lemmatisation properly | Pass |
| FrameworkComponents | To integrate all of the modules together properly | Pass |
| Stemmer | To transform a word into its root form. | Fail |
| WordPair | To read text corpus, construct formal context | Pass |

Table 4.7: Big Bang integrated testing on this project

## 4.9 Summary

This chapter has introduced the implementation and testing of the design phase from chapter three. At the first glance, an implementation of this project is introduced in chapter four. First and foremost, a novel approach is proposed to introduce framework architecture for the overall process of this project. Second, the modules of this framework are implemented which are Natural Language Processing parser, word pairs and formal context construction, conducting an experiment by using WordNet-based and frequency-based techniques, miscellaneous module, and formal concept and lattice compaction. Further, an example is applied to clarify the implementation steps with the experiment in detail. At the end of this chapter, the project is tested by using JUnit testing and Big Bang integrated testing.

# 5. Chapter Five: Evaluation and Result

In chapter four, the deriving concept lattices from free text has implemented and an experiment has conducted by using WordNet-based technique and frequency-based technique. Several tools, framework, and software application have used to do so. Chapter five is basically about evaluating the used frameworks and tools, and the resulted concept lattices as well as discussing the results and achievements from the described method of evaluation for the concept lattices found by the system. It begins with evaluating the used frameworks and tools in this project and dissertation including Stanford Natural Language Processing frame, the Concept Explorer, and WordNet. Afterwards, the methods of evaluating concept lattices are presented and discussed, and then the evaluation strategy that composes of a case study and its specification are presented. At the end of this chapter, the results, achievements, and limitations from the described method of evaluation for the concept lattices found by the system are discussed.

## 5.1 Frameworks and Tools

One of the objectives of this project and dissertation is to evaluate how well suited these tools and packages as a framework for ontology learning because a common framework for ontology is essential as:

- Several different processing components should be utilised for ontology learning.
- Different processing components are required by different situations.
- An analysis often includes several nested processing components together.

### 5.1.1 Stanford Natural Language Processing Framework

The experience of using Stanford Natural Language Processing as a framework has shown that it is highly customisable and scalable. It has several components that make it useful for ontology learning and the framework ought to not only be evaluated and tested with a small document, but also with a large document collection. However, it has some shortcomings. First and foremost, the Stanford Natural Language Processing framework is sometimes unable to determine a more precise dependency relation between two words. This is probably due to a weird grammatical construction, a parser error, a limitation of the Stanford Dependency conversion software, or due to unresolved long distance dependency. When the system is not able to determine the exact dependencies relation between two words, a special dependency type is shown and labelled as dep. For example, the dependency of the below text is *dep (show, if)*.

> "Them, as if to show that he could,…"

Second, Stanford sentence splitter splits sentences based on only newlines only or treat each document as one sentence without splitting it at all, yet these two situations for splitting sentences in that way are very unlikely and it does not really work in practical applications because they are not based on standard vanilla approach for sentence splitting to locate the end of a sentence and split a text corpus into sentences [43]. Third, stemming as the process for reducing inflected word to their stems, base or root form is an essential part of Natural Language Processing and information retrieval, but word stemming is not adopted by Stanford Natural Language Processing framework yet. Last but not least, syntactic dependencies tend to do not identify the dependency of those words that are both noun and verb as those words might be appeared as objects and attributes in the formal context. For example, "make" is both noun and verb. Apparently, this word should appear as either noun or verb in the formal context depending on its syntactic dependency in the sentence, yet it usually appears as both object and attribute in the formal context. Therefore, it seems that Part-of-Speech Tagger and Stanford dependency parser cannot syntactically find these kind of words' dependency and thereby all possible dependencies may be allocated to them inexpertly. Last, the system sometimes gets wrong where words are used as noun and verb. For example, "building" can be noun and gerund as well. This may be considered as noun by Stanford Natural Language Processing framework, but when it is used with WordNet, it may be interpreted as gerund. That makes a kind of inconsistency between the components of the system. This ambiguity is either related to the malfunction of Stanford's Part-of-Speech Tagger, WordNet, or both of them.

### 5.1.2 The Concept Explorer

The Concept Explorer is an open source project that is still on extension and development used to implement the basic ideas needed for the research and study of Formal Concept Analysis (FCA) and lattice compaction. The Concept Explorer can be used to analyse attribute-object tables which are called formal context in Formal Concept Analysis, draw the corresponding lattice, and to explore different dependencies that may exist between attributes. However, it has the following drawbacks:

1. The Concept Explorer is not entirely stable as it is still in development stage.
2. The Concept Explorer does not directly support XML file, but it supports XML syntax via CEX file format.
3. By using the Concept Explorer, deriving the concept lattice from the formal context might be a time consuming processing depending upon the complexity and size of the data. Thus, lattice-layout tools are time consuming and lattice drawings consist of only one node can appear. After some time, the whole structure will follow.
4. This open source application ought to be tested properly.

### 5.1.3 WordNet

WordNet as a lexical electronic database for the English language is considered to be the most important resource that is available to researchers in computational linguistics [32, 33] to support automatic text analysis and artificial intelligence applications. WordNet is also known as one of the most used manually compiled electronic dictionaries that is not restricted to any specific domain and covers almost all English verbs, nouns, adjectives, and adverbs. Even though there are similar products, such as CYC, Cycorp, and Roget's International Thesaurus, WordNet is the most successful and growing one and it has been used in several applications over the last ten years [12]. However, it has some problems and limitation as pointed out below [12]:

1. WordNet does not facilitate changing the word forms. For example, from nouns to verbs or from adjectives to nouns and so on so forth.
2. WordNet is relatively designed for manual consultation rather than for processing natural language texts automatically. As a result, there is not any particular emphasis to differentiate between the various involved concepts automatically.
3. WordNet still has polysemy issue as this may prompt flawed operation in several Natural Language Processing systems because processing is often done with specific records or sublanguages. Sometimes, finding synonyms or hypernyms of a word is not exactly accurate for those words that have more than one meaning or have different meaning in different context. For example, a word may come with a specific meaning in the free text, but WordNet may find different meaning of it when it searches for its synonym and hypernym.
4. Classification is conducted manually in WordNet. That is, the depth and reasons of classification may not be consistent.
5. As mentioned in the previous sections, the system sometimes gets wrong where words are used as noun and verb and this leads to a kind of inconsistency between the components of the system. This ambiguity is either related to the malfunction of Stanford's Part-of-Speech Tagger, WordNet, or both of them.

### 5.2 Evaluation Methods

It is needed to assess how good the concept lattices and concept hierarchies reflect a given domain so as to evaluate our approach. There are several methods for evaluating lattice graphs. Firstly, one possible approach is to compute how similar the automatically concept hierarchy against a given hierarchy for the domain in question, but the crucial question here is how to define similarity between concept hierarchies. Even though there is a substantial amount of work on how to compute similarity between trees in the community of Artificial Intelligence about concept lattices, plain graphs, and conceptual graphs by [49-51], it is not clear how to measure these similarities and translate to concept hierarchies as well. Secondly, another interesting study is presented by [52] in which ontologies can be compared along with different levels. These levels are semiotic, pragmatic and syntactic. Particularly, the measure to compare the taxonomic and lexical overlap between two ontologies is presented. In addition, an interesting study is presented I which different subjects were asked to model ontology. As a result, the ontologies are compared in regard with the defined similarity measures and thereby the agreement of different subject on the task of modelling ontology is yielded. Due the fact that concept lattice is a specific kind of structure homomorphism, evaluating the concept lattices by finding a relationship between the lattice graphs is another method for the evaluation. In graph theory, two graphs are homeomorphism if there is an isomorphism from subdivision of first graph to some subdivision of the second one [53, 54]. For example, graph G and H are homeomorphism if there is an isomorphism from subdivision of graph G to some subdivision of the graph H. Further, subdivision of a graph in graph theory can be defined as a graph that results from the subdivision of edges in itself [53]. For instance, the subdivision of some edge e with endpoints {w, z} produces a graph that contains one new vertex x and with an edge set replacing e by two new edges, {w, x} and {x, z}. This way of evaluation seems more systematic and formal way because it is theoretically clear to find two graphs are whether homeomorphism or not. However, it is not clear yet how to measure the homeomorphism of two lattice graphs in spite of substantial work about that in the community of Artificial Intelligence. Thirdly, one more possible evaluation would be to compute how many concepts in the automatically learning ontology are correct. This has been done by [55] and [56]. This way could be adapted for use in this project and dissertation by comparing concept count, edge count, lattice height, and lattice width estimation before and after applying the WordNet-based technique or/and Frequency-based technique.

## 5.3 Evaluation Strategy

As a result of the implementation and experiment, there are two types of concept lattices to evaluate. The first concept lattice is directly derived from a free text without applying either WordNet-based techniques or Frequency-based technique. The second type is derived from a free text after applying the WordNet-based technique or/and Frequency-based technique. The second concept lattice type is further divided into four concept lattices based on applying the type and/or the order of applied techniques. This includes applying WordNet-based technique, Frequency-based technique, WordNet-based and Frequency-based techniques, and Frequency-based and WordNet-based techniques. Figure 5.1 shows the mechanism of deriving the first concept lattice and the second one including the sub-types of the latter one.

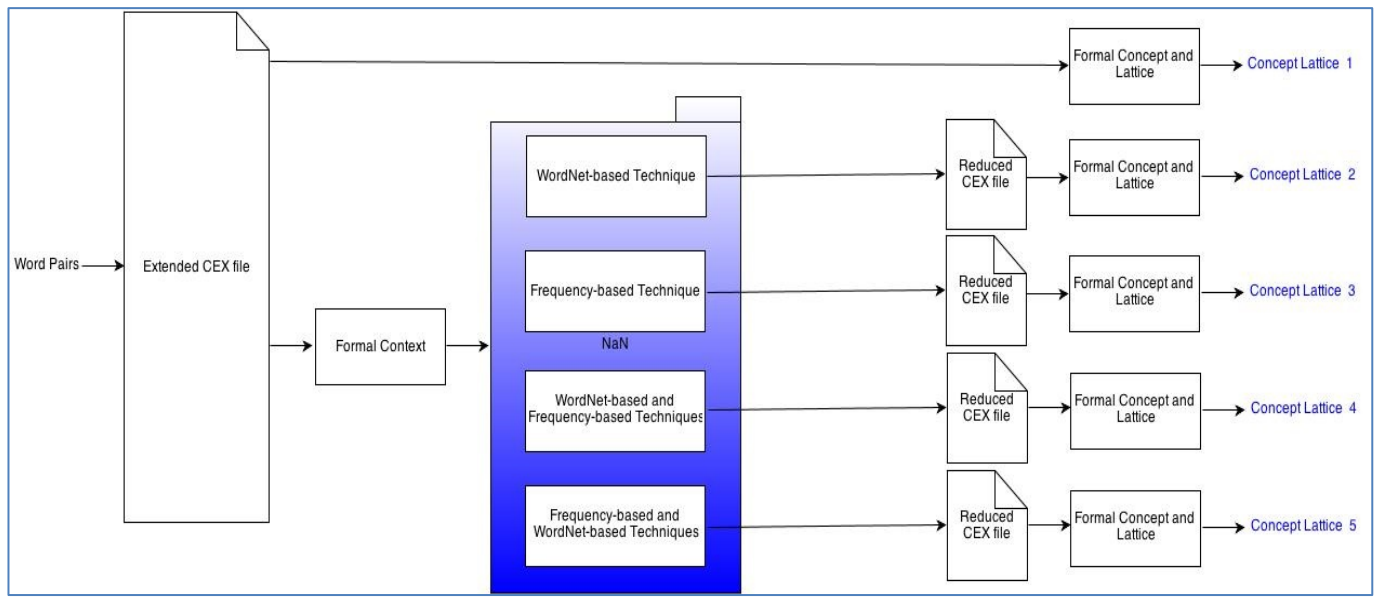

Figure 5.1: Statistics of concept lattice 1 and concept lattice2

In addition, one possible used evaluation method in this project and dissertation would be to compute how the statistical data of the resulted concept lattices and compare them. The statistical data of lattices mainly include concept count, edge count, lattice height, and lattice width estimation. This evaluation method is explained in detail as a case study in the following sub-sections.

### 5.3.1 Case Study

To evaluate the results of concept lattices, a case study as a descriptive, exploratory or explanatory analysis of the result is used to explore causation and find underlying principles about the resulted concept lattices in the implementation and the experiment as well. This is also used to find the impact of applying the order of WordNet-based technique and/or Frequency-based technique. This case study is prospective in which criteria are established and cases fitting the criteria are included as they become available. Table 5.1 gives a summary about this case study.

| Summary | To evaluate the result of concept lattices in order to explore causation and find underlying principles about the resulted concept lattices |
|---|---|
| Issue to be resolved | To evaluate the results |
| Specifications | Described in detail in the next section |
| Result | To find underlying principles about the resulted concept lattices |
| Next step | To find the achievements and limitations of this system |

Table 5.1: Summary of the case study

The case study includes the following five steps:

1. Applying twenty different types of free texts on the system. These text corpuses are taken from Wikipedia
2. Each text corpus is applied on the system to derive five different types of concept lattices. Figure 5.1 details that.

3. Some statistical data, such as concept count, edge count, lattice height and, lattice with estimation are calculated from each concept lattice
4. Standard deviations are calculated for the statistical data of the concept lattices.
5. Finally, the result can be presented visually.

**5.3.2 Case Study Specifications**

As mentioned in the previous section in brief, this case study takes the below five steps:

**1. Text Corpus as Input Source:** The text corpuses that are taken from Wikipedia have relatively the same length of word and all of the words in these text corpuses do not need stemming. Both WordNet-based and Frequency-based techniques are applied on these text corpuses together in different order and each of them is applied on them separately as well. The free text titles, their length with a brief description are presented in the appendix D. The topics of the text corpuses are as follows:

*British Empire, Traffic Law, Winston Churchill, Financial Times, Chemistry, British Museum, Adult Psychology, CCTV, Corporate Governance, Agriculture, Biological Agent – Virus, English Language, East Enders, BBC, Lion, UK Weather, Astronomy, Football, FBI, and Public Transportation*

**2. Applying the System:** After producing formal context from each free text, as it is shown in Figure 5.1, the technique(s) is applied on each formal context to derive five different types of concept lattices. The concept lattice results are related to the order and type of applying the techniques. For example, no techniques are applied on the formal context of concept lattice 1, whereas on or two techniques are applied on the other formal concepts of lattice 2, 3, 4, and 5. Table 5.2 presents the order and type of techniques on applying different formal contexts. As a result of applying these four different ways of techniques, there are five varieties of lattice graphs for the for the second concept lattice.

| Lattice ID | Lattice Name | Applied Technique(s) | Hypernym Depth | Threshold |
|---|---|---|---|---|
| 1 | Concept lattice 1 | - | - | - |
| 2 | Concept lattice 2 | WordNet-based technique | 4 | - |
| 3 | Concept lattice 3 | Frequency-based technique | - | 2% |
| 4 | Concept lattice 4 | WordNet-based and Frequency-based techniques | 4 | 2% |
| 5 | Concept lattice 5 | Frequency-based and WordNet-based techniques | 4 | 2% |

Table 5.2: The order of applying techniques on formal context

**3. Statistical Data:** After deriving the concept lattices, some statistical data from each of them are recorded. These statistical data are concept count, edge count, lattice height, and lattice width. Table 5.3 details statistical data of concept lattice.

| Statistical data | Detail |
|---|---|
| Concept count | Number of concepts in a concept lattice |
| Edge count | Number of edges in a concept lattice |
| Height | The height of a concept lattice |
| Width | The estimation width of a concept lattice |

Table 5.3: Statistics of concept lattices

**4. Standard Deviations:** Then, several standard deviation functions are applied on these statistical data. These functions are shown in Table 5.4.

| Statistical function | Detail |
|---|---|
| Mean | Returns the average of its arguments |
| Median | Returns the median, or number in the middle of the set of given numbers |
| Sum | Adds all the numbers in a range of cells |
| Max | Returns the largest value in a set of values, ignores logical values and text |
| Min | Returns the smallest value in a set of values, ignores logical values and text |
| STDV P | Calculates the standard deviation based on the entire population given as arguments |
| STDV S | Estimates standard deviation based on a sample |
| STDEVA | Estimates standard deviation based on a sample, including logical values and text |

Table 5.4: Applied standard deviation functions on concept lattices

**5. Presenting Results:** At the final step, the standard deviations of the statistical data for the techniques can be retrieved. Then, these results can be visually presented to be analysed clearly and concisely. This is done in the next section.

## 5.4 Results and Achievements

As already mentioned above, this approach is evaluated by applying twenty distinct free texts to get some statistical data from their resulted concept trees. In the below sub-sections, these results are analysed and the achievements are discussed as well.

### 5.4.1 Result Analysis

To analyse the results, we have created an excel sheet to record the results for the five concept lattices of all the twenty text corpuses and add the results of concept count, edge count, lattice height, and lattice width estimation. Then, standard deviations for these statistical data are calculated. Finally, the results can be presentenced visually. For example, the mean of concepts of concept lattice 1 is 43.2, whereas it is 30.2 for concept lattice 3. Table 5.5 gives the standard deviation values of concept count and edge count from concept lattice 1 to 5. In the appendix F, the additional results is given for presenting the standard deviation values for concept count, edge count, height, and width estimation for all the concept lattices of text corpuses.

| Text | 1. Original Text - Lattice | | 2. WordNet-based – Lattice | | 3. Frequency-based Lattice | | 4. WordNet and Frequency Lattice | | 5. Frequency and WordNet Lattice | |
|---|---|---|---|---|---|---|---|---|---|---|
| | Concept | Edge | Concept | Edge | Concept | Edge | Concept | Edge | Concept | Edge |
| 1 | 47 | 86 | 38 | 65 | 17 | 28 | 38 | 65 | 15 | 23 |
| 2 | 55 | 101 | 50 | 95 | 31 | 57 | 50 | 95 | 47 | 99 |
| 3 | 21 | 35 | 21 | 35 | 35 | 64 | 21 | 35 | 21 | 35 |
| 4 | 31 | 56 | 34 | 68 | 31 | 56 | 34 | 68 | 34 | 68 |
| 5 | 40 | 71 | 32 | 54 | 40 | 71 | 32 | 54 | 32 | 54 |
| 6 | 44 | 81 | 38 | 68 | 23 | 40 | 38 | 68 | 21 | 36 |
| 7 | 51 | 91 | 59 | 114 | 21 | 34 | 59 | 114 | 26 | 48 |
| 8 | 59 | 106 | 34 | 62 | 36 | 63 | 34 | 62 | 24 | 40 |
| 9 | 48 | 86 | 18 | 26 | 13 | 21 | 18 | 26 | 17 | 27 |
| 10 | 54 | 99 | 48 | 96 | 32 | 57 | 48 | 96 | 28 | 48 |
| 11 | 58 | 110 | 33 | 55 | 32 | 59 | 33 | 55 | 23 | 40 |
| 12 | 22 | 37 | 12 | 16 | 22 | 37 | 12 | 16 | 12 | 16 |
| 13 | 50 | 94 | 70 | 146 | 26 | 46 | 70 | 146 | 30 | 52 |
| 14 | 36 | 63 | 29 | 47 | 36 | 63 | 29 | 47 | 29 | 47 |
| 15 | 24 | 43 | 17 | 26 | 24 | 43 | 17 | 26 | 17 | 26 |
| 16 | 33 | 63 | 21 | 32 | 33 | 63 | 21 | 32 | 21 | 32 |
| 17 | 45 | 83 | 20 | 31 | 45 | 83 | 20 | 31 | 20 | 31 |
| 18 | 49 | 92 | 79 | 185 | 25 | 46 | 79 | 185 | 28 | 52 |
| 19 | 32 | 58 | 26 | 44 | 32 | 58 | 26 | 44 | 26 | 44 |
| 20 | 65 | 131 | 41 | 76 | 50 | 101 | 41 | 76 | 51 | 106 |
| Mean | 43.2 | 79.3 | 36 | 67.05 | 30.2 | 54.5 | 36 | 67.05 | 26.1 | 46.2 |
| Median | 46 | 84.5 | 33.5 | 58.5 | 31.5 | 57 | 33.5 | 58.5 | 25 | 42 |
| Sum | 864 | 1586 | 720 | 1341 | 604 | 1090 | 720 | 1341 | 522 | 924 |
| Max | 65 | 131 | 79 | 185 | 50 | 101 | 79 | 185 | 51 | 106 |
| Min | 21 | 35 | 12 | 16 | 13 | 21 | 12 | 16 | 12 | 16 |
| STDV P | 12.51 | 24.97 | 17.4 | 41.74 | 8.86 | 18.21 | 17.4 | 41.74 | 9.49 | 22.29 |
| STDV S | 12.84 | 26.47 | 17.85 | 44.74 | 9.09 | 18.62 | 17.85 | 44.74 | 9.73 | 19.65 |
| STDEVA | 12.84 | 25.62 | 17.85 | 42.83 | 9.09 | 18.69 | 17.85 | 42.83 | 9.73 | 22.87 |

Table 5.5: Evaluation results for concept and edge

To further clarify the results, these values can be presented in bar chart as well. Figure 5.2 depicts the concept count mean (left) and sum (right) of concept lattices from the twenty text corpuses.

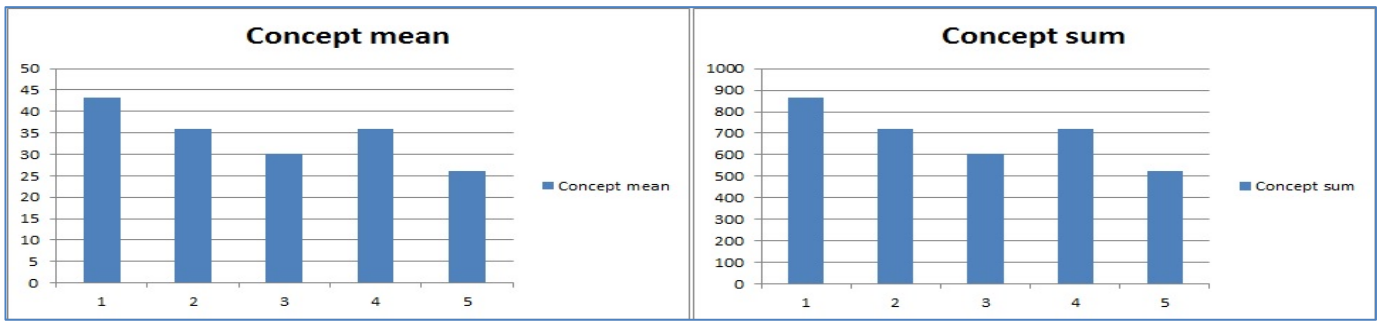

Figure 5.2: Mean of concept count (left) and sum of concept count (right)

Where:

1: number of concepts without applying any technique

2: number of concepts with applying WordNet-based technique

3: number of concepts with applying Frequency-based technique

4: number of concepts with applying WordNet-based technique and Frequency-based technique

5: number of concepts with applying Frequency-based technique and WordNet-based technique

Meanwhile, Figure 5.3 depicts the edge count mean (left) and sum (right) of concept lattices from the twenty text corpuses. Additional graphical results are shown in the appendix F.

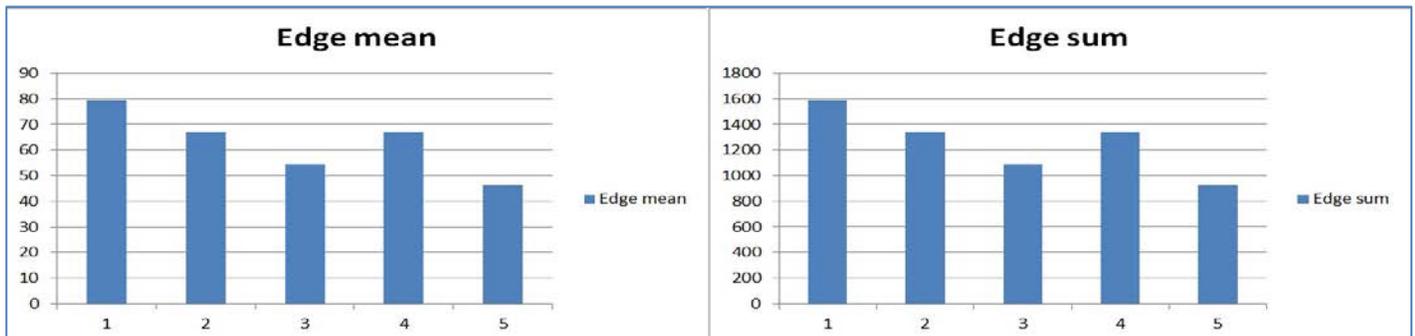

Figure 5.3: Mean of concept count (left) and sum of concept count (right)

Based on the given standard deviation values, the concept count, edge count, height, and width estimation are generally decreased from concept lattice 1, 2, 3, and 4 in comparison with the concept lattice 1, but the decrease rate is varied from one concept lattice to another. For example, the decease rate of concept count mean is the maximum in concept lattice 3 and it is the minimum in concept lattice 2 and 4 in comparison with the concept lattice 1. Different ways of applying the techniques on the formal concept of concept lattice 1 might substantially or insubstantially affect the decease rate of the concept lattice 1, 2, 3, and 4. On the contrary, the lattice height of concept lattice 2, 3, 4, and 5 are decreased in considerably in comparison with the concept lattice 1. There are some other factors that might substantially or insubstantially affect the result:

- The threshold for the Frequency-based technique
- The depth of hypernym for the WordNet-based technique
- In some situations, the type of free text
- It is not really clear and justified whether the length of free text might affect or not because the used free texts here have relatively the same length.

In spite of reducing the size of formal context, it can be seen that the resulted concepts of lattice 2, 3, 4, and 5 are still a subset of the concepts of lattice concept 1 and the most generalised names are used in their formal contexts. However, there are some concepts in the concept lattice 1 that do not exist in the concept lattice 2, 3, 4, and 5. The objects and attributes of these un-existed concepts in the concept lattices 2, 3, 4, and 5 are the least used one or they might be uninteresting and erroneous, but there is not any empirical evidence to justify it and this has not evaluated and confirmed practically yet. As a consequence, it can be seen that the reduced concept lattices have relatively the same quality of results as the concept lattice 1 thanks to the common concepts between them, but this should be confirmed and justified yet. Figure 5.3 presents a Venn diagram as the relationship between the concept of concept lattice 1 and the reduced concept lattices.

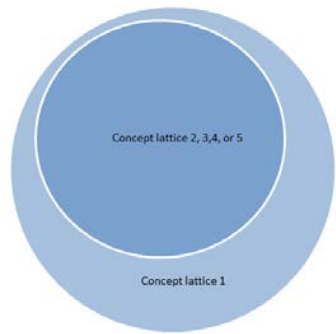

Figure 5.3: Venn diagram of concept lattices

**5.4.2 Achievements**

Via the lens of the evaluations and the result analysis, the main findings and achievements from this project and dissertation are as follows:

1. A framework has introduced for deriving concept lattice from text corpuses by Formal Concept Analysis and WordNet, and then it can be converted into a special type of partial order constituting a concept and concept hierarchy.

2. A novel approach has proposed for conducting an experiment by using two distinct techniques which are WordNet-based and Frequency-based techniques to eliminate the erroneous and uninteresting pairs in the formal context and reduce the size of formal context to result in less time consuming when concept lattice derives from it. Through the lens of that experiment, the quality of concept lattices can be investigated.

3. In the experiment, the concepts between the result of concept lattice and the reduced concept lattice tend to be similar.

4. The eliminated word pairs in the formal context by using WordNet-based technique and Frequency-based technique can be considered as erroneous and uninteresting pairs, but it has not evaluated and confirmed yet.

5. The size of formal context can be reduced by using any of linguistic resources or statistical approaches, but the impact of using statistical approaches is more than the impact of using linguistic resources.

6. The impact of using each WordNet-based technique and/or Frequency-based technique is investigated in terms of reducing the size of formal context. In regard with using one of these techniques, the impact of using WordNet-based technique is not as efficient as the impact of using Frequency-based technique. In regard with using both of these techniques, on the other hand, the impact of using Frequency-based technique and WordNet-based technique is more efficient than the impact of using WordNet-based technique and Frequency-based technique. It has found that the order of applying these techniques in an efficient way is as follows:
    - Applying Frequency-based technique at the first time, and then applying WordNet-based technique
    - Applying Frequency-based technique only
    - Applying WordNet-based technique firstly, and then Frequency-based technique
    - Applying WordNet-based technique

7. The impact of applying WordNet-based technique is slightly varied depending on the type and length of text corpus.

8. Specifying the threshold for Frequency-based technique may considerably affect the result of reduced the formal context. Likewise, the depth of hypernym may substantially or insubstantially affect the result of reduced formal context. In other words, the effectiveness is Frequency-based technique is varied from a free text to another depending on specifying a proper threshold for type of free text and the effectiveness of WordNet-based technique is varied from one free text to another depending upon specifying a suitable depth of hypernym between the words.

9. The related concepts found by both concept lattices tend to be good quality, but it has not evaluated and confirmed yet.

10. Learning concept hierarchies for the other languages, particularly Latin alphabet based languages can make use of the basic architecture of project and dissertation and their baseline can be adapted to these languages for deriving concept hierarchies.

### 5.4.3 Limitations

Certainly, this project and dissertation are not out of problems. There are a few issues that should be raised so as to be tackled in the future. These issues are listed as follows:

1. The system is evaluated, but it could be further evaluated by using a more algorithmic to find relationships between the lattice graphs.
2. The system could be evaluated by using larger text corpuses.
3. In the pruning step, the word pairs need to be stemmed to reduce the inflected word pairs to their stem, base or root form, but this system is not currently able to do so.
4. The experiment of this project and dissertation has conducted on the English language. Nevertheless, there are some minor issues between British and American English. For instance, according to the Longman Dictionary, in American English it is also possible to ride vehicles in general. So, there may be some mismatching if the free corpus is considered as British English and treated as American English when its terms are compared with WordNet to find a proper pairs.
5. Sometimes, it is not clear what depth between word do we need to specify hypernym.
6. There is still polysemy issue in this system as the systems does not have the capacity for a sign or signs to have multiple related meanings. Because the system is not capable for signing the specific meaning of multiple related meaning words, a word is usually regarded as distinct from homonymy, in which the multiple meanings of a word may be unconnected or unrelated.
7. Text length of the applied text corpuses is relatively static and this may affect the result.
8. Syntactic dependencies tend to do not identify the dependency of those words that are both noun and verb as those words might be appeared as objects and attributes in the formal context. Hence, it seems that Part-of-Speech tagger and Stanford dependency parser cannot syntactically find these kind of words' dependency and thereby all possible dependencies may be allocated to them inexpertly.
9. The system sometimes gets wrong where words are used as noun and verb. For example, "building" can be noun and gerund as well. This may be considered as noun by Stanford Natural Language Processing framework, but when it is used with WordNet, it may be interpreted as gerund. That makes a kind of inconsistency between the components of the system. This ambiguity is either related to the malfunction of Stanford's Part-of-Speech tagger or WordNet.
10. Classification as a way of identifying text corpuses into set of categories or sub-categories has not taken into account in this project and dissertation because this may help yielding a better result quality and making an easier decision to assign a proper value of hypernym depth and the threshold of Frequency-based technique.

### 5.5 Summary

This chapter has basically evaluated the used frameworks and tools, and the resulted concept lattices as well as discussed the results and achievements from the described method of evaluation for the concept lattices found by the system. It has begun with evaluating the used frameworks and tools in this project and dissertation including Stanford Natural Language Processing frame, the Concept Explorer, and WordNet. Afterwards, the methods of evaluating concept lattices have presented and discussed, and then the evaluation strategy that composes of a case study and its specification have presented. At the end of this chapter, the results, achievements, and limitations from the described method of evaluation for the concept lattices found by the system have discussed.

# 6. Chapter Six: Conclusion and Future Work

This chapter concludes the project and dissertation by first concluding the research, and matching it up to the objectives and research questions to the state of art that we laid out in chapter one. Following that, different ways of extending this project and dissertation in the future are discussed.

## 6.1 Conclusion

To achieve the objectives and answer the research questions of this work as a part and contribution of the on-going research in the field of ontology learning, this project and dissertation have basically proposed a framework for learning concept hierarchies from free texts by using Formal Concept Analysis and WordNet. In particular, this study has focused on the following: Firstly, a framework has introduced for deriving concept lattice from text corpuses, and then it can be converted into a special type of partial order constituting a concept and concept hierarchy; Secondly, a novel approach has proposed for conducting an experiment by using and investigating two distinct techniques which are WordNet-based and Frequency-based techniques to eliminate the erroneous and uninteresting pairs in the formal context and reduce the size of formal context in order to result in less time consuming when concept lattice derives from it; Thirdly, the proposed framework, and the experiment with its resulted concept lattices have evaluated as well as discussing the results, achievements, and limitations from the described method of evaluation for the concept lattices found by the system. In other words, this project and dissertation have provided a system as a baseline of basic engineering principles for deriving concept hierarchies by using Formal Concept Analysis and WordNet with involving the use of Stanford Natural Language Processing framework and the Concept Explorer. The system, however, has its uncertainty factors and limitations. Evidently, the system may be hindered by these limitations and uncertainty factors to draw a logical conclusion. Still, the results indicate, considering the uncertainty factor into account, the following:

1. The size of formal context can be reduced by using any of linguistic resources or statistical approaches, but the impact of using statistical approaches is more efficient than the impact of using linguistic resources. In the meantime, the size of formal context can be reduced by using any of linguistic resources or statistical approaches or both of them can be used in different order. It has found that the order of applying these techniques in an efficient way. In addition, the impact of applying WordNet-based technique is slightly varied depending on the type and length of text corpus.

2. The impact of using each WordNet-based technique and/or Frequency-based technique is investigated in terms of reducing the size of formal context. The size of formal context can be reduced by using any of linguistic resources or statistical approaches or both of them can be used in different order. It has found that the order of applying these techniques in an efficient way is as follows:
    - Applying Frequency-based technique at the first time, and then applying WordNet-based technique
    - Applying Frequency-based technique only
    - Applying WordNet-based technique firstly, and then Frequency-based technique
    - Applying WordNet-based technique

3. In the experiment, the concepts between the result of concept lattice and the reduced concept lattice tend to be similar as there are similarities between the concepts of the reduced concept lattice and the original one. Hence, it seems that the size of formal context can be reduced without a substantial effect on the quality of the result and the reduced concept lattice could probably be used as an alternative of the original one, but this has not confirmed and justified yet.

4. The impact of applying WordNet-based technique is slightly varied depending on the type and length of text corpus.

By doing so, the value of hypernym depth and threshold may affect the result of applying WordNet-based technique and Frequency-based technique respectively on the formal context.

5. The eliminated word pairs in the formal context by using WordNet-based technique and Frequency-based technique can be considered as erroneous and uninteresting pairs, but it has not evaluated and confirmed yet.

As the essence of this project and dissertation, a framework has proposed for learning concept hierarchies from free texts and an experiment has conducted on that basis. This study has generally dealt with some research questions as part of the on-going research on ontology learning in regard with reducing the size of formal context by applying different techniques. In the meantime, it has shown that the experience with using the frameworks and tools for ontology learning are customisable and scalable as well. They have several built-in components that make it useful for ontology learning, but they have several limitations that ought to be fixed in the future. Consequently, the results are potential and the study tend to make a contribution in the field of ontology learning and its objective has also been to establish a baseline for further research because this study could be a step towards the goal of finding more principles and answering more research questions completely with a proper improvement, and a more systematic and proper evaluation with the involvement of a domain expert to check the quality of the concept lattice and the reduced ones. In turn, further research on the baseline of this study will probably be a potential contribution to the field on ontology learning.

**6.2 Future Work**

As mentioned earlier, his project and dissertation have provided a framework provided a framework for learning concept hierarchies from free texts by using Formal Concept Analysis and WordNet and an experiment has conducted on this basis. Although this study indicates several potential results, there are possibilities for improvements in it by the use of Formal Concept Analysis and WordNet. These improvements not only lie in the development of the project and dissertation, but also lie in the identifying their issues for further research in the future. Due to the very short time limit of this project and dissertation, some limitations of their components, and a few inconsistencies among them, there are some issues that should be tackled in the future. Accordingly, further works in the future are mostly through the lens of these issues. Paving the way for further research, the following below points can be done in the future:

1. The system is evaluated on twenty text corpuses, but it could be further evaluated by finding a relationship between the lattice graphs whether they are homeomorphism or not. This is a more algorithmic approach for the evaluating the original concept lattice and the reduced concept lattice. Further evaluation in the future includes evaluating the quality and similarity of the concept lattice with the reduced concept lattice.
2. The system could be evaluated more concisely by using larger text corpuses with different length of words.
3. According to the result of concept count and edge count in the evaluation, it can be hypothesised in the future whether the reduced concept lattice has relatively the same quality of results as the original concept lattice or not.
4. In WordNet, multi-disciplinary is still an issue that prompts flawed operation in several Natural Language Processing systems because processing is often done with specific records or sublanguages. As a result, there is still polysemy issue in this system as the systems does not have the capacity for a sign or signs to have multiple related meanings. This issue can be addressed in the future by comparing the syntactic word dependencies with WordNet or classifying the category or sub-category of the text corpuses before applying to the system.
5. The system sometimes gets wrong where words are used as noun and verb. That makes a kind of inconsistency between the components of the system. This ambiguity is either related to the malfunction of Stanford's Part-of-Speech Tagger or WordNet. This issue should be addressed in a way that makes the distinct components of the system to be more consistent together.
6. Syntactic dependencies of Stanford Natural Language Processing framework tend to do not identify the dependency of those words that are both noun and verb as those words might be appeared as objects and attributes in the formal context. Therefore, it seems that Part-of-Speech tagger and Stanford dependency parser cannot syntactically find these kind of words' dependency and thereby all possible dependencies may be allocated to them inexpertly. This framework should be improved or an extra component can be built upon it as a remedy of that issue.
7. Stanford sentence splitter splits sentences based on only newlines only or treat each document as one sentence without splitting at all, yet these two situations for splitting sentences are very unlikely and they do not really work in practical applications because they are not based on standard vanilla approach for sentence splitting to locate the end of a sentence and split a text corpus into sentences [43]. This has done in the current system, but it could be integrated with the Stanford Natural Language Processing framework in the future.
8. In the pruning step, the word pairs need to be stemmed to reduce the inflected word pairs to their stem, base or root form, but this system is not currently able to do so. Stemming as the process for reducing inflected word to their stems, base or root form is an essential part of Natural Language Processing and information retrieval, but Stanford Natural Language Processing framework is not adopted word stemming yet. This could be fixed in the future and integrate it with Stanford Natural Language Processing framework.
9. The experiment of this project and dissertation has conducted on the English language. Nevertheless, there are some minor issues between British and American English. For instance, according to the Longman Dictionary, in American English it is also possible to ride vehicles in general. So, there may be some mismatching if the free corpus is considered as British English and treated as American English when its terms are compared with WordNet to find a proper pairs. Thus, these accents of the English language could be categorised in the future work such that the Stanford's Part-of-Speech Tagger and syntactic dependency could extract the pairs on the basis of the selected accent.

10. The Stanford Natural Language Processing framework is sometimes unable to determine a more precise dependency relation between two words. This is probably due to a weird grammatical construction, a parser error, a limitation of the Stanford Dependency conversion software, or due to unresolved long distance dependency. When the system is not able to determine the exact dependencies relation between two words, a special dependency type is shown and labelled as dep. This is one of the issues of Stanford Natural Language Processing framework and it could be pointed out in the future work.

11. Classification as a way of identifying text corpuses into set of categories or sub-categories has not taken into account in this project and dissertation. Hence, this could be considered to work on it in the future as a way to help yielding a better result quality and making an easier decision to assign a proper value of hypernym depth and the threshold of Frequency-based technique.

12. Sometimes, it is not clear how to specify the values of hypernym depth for WordNet-based technique and threshold for Frequency-based technique. There should be a mechanism to tune the value of hypernym depth and threshold to have a relatively good result.

13. This system can be used to provide a software framework to be a universal and reusable software platform for developing further work on the basis of Formal Concept Hierarchies and WordNet.

14. Learning concept hierarchies for the other languages, particularly Latin alphabet based languages can make use of the basic architecture of project and dissertation and their baseline can be adapted to these languages for deriving concept hierarchies.

15. Even though this approach is fully automatic, we will explore in the future how users can be involved in the process by making use of hashtag with the text corpuses that can provide a means of grouping such messages in a text corpus.

16. An intuitive application program with Graphical User Interface (GUI) based can be designed for the system to have a friendly interface and work like users would expect. This could be either based on the Concept Explorer or worked independently.

# Appendixes

## Appendix A: The Project Work Plan and Milestones

Within the three months of my project and dissertation, it was important that I planned my time carefully and anticipate any issue that may have prevented me bringing the project and dissertation to a satisfactory conclusion. In the Gantt chart below, the work plan and milestones of my project and dissertation are provided.

| Week number<br>Week beginning | 1<br>3/6 | 2<br>10/6 | 3<br>17/6 | 4<br>24/6 | 5<br>1/7 | 6<br>8/7 | 7<br>15/7 | 8<br>22/7 | 9<br>29/7 | 10<br>5/8 | 11<br>12/8 | 12<br>19/8 | 13<br>26/8 | 14<br>2/9 |
|---|---|---|---|---|---|---|---|---|---|---|---|---|---|---|
| Reading | | ■ | | | | | | | | | | | | |
| Proposal | | ■ | ■ | | | | | | | | | | | |
| Project Brief | | | | ■ | | | | | | | | | | |
| Background and Literature Review | | | | ■ | ■ | ■ | | | | | | | | |
| Implementation | | | | | | ■ | ■ | ■ | ■ | | | | | |
| Evaluation and Result Analysis | | | | | | | | | | ■ | ■ | | | |
| Reading | | ■ | ■ | ■ | ■ | ■ | ■ | ■ | ■ | ■ | ■ | | | |
| Conclusion | | | | | | | | | | | ■ | | | |
| Writing a draft | | | | ■ | ■ | ■ | ■ | ■ | ■ | ■ | ■ | | | |
| Writing-up | | | | | | | | | | | | ■ | ■ | ■ |
| Milestone – demonstrate to supervisor/examiner | | | | | | | | | | | | × | | |
| Milestone – dissertation draft complete | | | | | | | | | | | | × | | |
| Final corrections | | | | | | | | | | | | | ■ | ■ |
| Milestone – Hand-in | | | | | | | | | | | | | | × |

Table A.1: Gantt char of the project work plan and milestones

## Appendix B: Acronyms and Abbreviations

- API: Application Programming Interface
- NLP: Natural Language Processing
- NP: Noun Phrase
- VP: Verb Phrase
- FCA: Formal Concept Analysis
- XML: Extensible Mark-up Language
- CEX: Caddie Exchange File
- GUI: Graphical User Interface

## Appendix C: Digital Appendix

This project and dissertation are attached with a ZIP file that contains the following:

- Java classes and source codes of the implementation
- Classes of the automated JUnit testing
- The twenty free texts in TXT files
- The formal context results of the twenty free texts in CEX files and their concept lattices can be derived from them accordingly
- The evaluation results including concept count, edge count, height, and lattice width estimation for the original concept lattice and the other four reduced concept lattices
- The result of standard deviations including mean, median, sum, max, min, STDV P, STDV S, and STDEVA for the original concept lattice and the other four reduced concept lattices
- The graph results of the evaluation section

## Appendix D: Case Study

Table D.1 presents the free text titles, their length with a brief description in the evaluation.

| No | Original free text | Text length | Description |
|---|---|---|---|
| 1 | British Empire | 1005 | The British Empire comprised the dominions, colonies, protectorates, mandates and other territories ruled or administered by the United Kingdom |
| 2 | Traffic Law | 1203 | Traffic on roads may consist of pedestrians, ridden or herded animals, vehicles, streetcars and other conveyances, either singly or together, while using the public way for purposes of travel. |
| 3 | Winston Churchill | 949 | Sir Winston Leonard Spencer-Churchill was a British politician who was Prime Minister of the United Kingdom from 1940 to 1945 and again from 1951 to 1955 |
| 4 | Financial Times | 804 | The FT was launched as the London Financial Guide on 9 January 1888, renaming itself the Financial Times on 13 February the same year |
| 5 | Chemistry | 766 | Chemistry, a branch of physical science, is the study of the composition, properties and behaviour of matter. |
| 6 | British Museum | 1063 | The British Museum is a museum in London dedicated to human history and culture |
| 7 | Adult Psychology | 978 | A young or prime adult is generally a person in the age range of 20 to 40, whereas an adolescent is a person aging from 13 to 19, although definitions and opinions vary. |
| 8 | CCTV | 1194 | Closed-circuit television (CCTV) is the use of video cameras to transmit a signal to a specific place, on a limited set of monitors. |
| 9 | Corporate Governance | 1056 | Corporate governance refers to the system by which corporations are directed and controlled |
| 10 | Agriculture | 991 | Agriculture, also called farming or husbandry, is the cultivation of animals, plants, fungi, and other life forms for food, fiber, biofuel, drugs and other products used to sustain and enhance human life |
| 11 | Biological Agent – Virus | 1429 | A virus is a small infectious agent that replicates only inside the living cells of other organisms. |
| 12 | English Language | 981 | English is a West Germanic language that was first spoken in early medieval England and is now the most widely used language in the world. |
| 13 | East Enders | 1346 | EastEnders is a British television soap opera, first broadcast in the United Kingdom on BBC One on 19 February 1985. |
| 14 | BBC | 838 | The BBC is a semi-autonomous public service broadcaster[6] that operates under a Royal Charter[7] and a Licence and Agreement from the Home Secretary.[ |
| 15 | Lion | 905 | The lion is one of the four big cats in the genus Panthera and a member of the family Felidae |
| 16 | UK Weather |  | The United Kingdom straddles the geographic mid-latitudes between 49–60 N. |
| 17 | Astronomy | 1038 | Astronomy is a natural science that is the study of celestial objects, the physics, chemistry,mathematics, and evolution of such objects, and phenomena that originate outside the atmosphere of Earth, including supernovae explosions, gamma ray bursts, and cosmic background radiation. |
| 18 | Football | 1017 | Football refers to a number of sports that involve, to varying degrees, kicking a ball with the foot to score a goal |
| 19 | FBI | 905 | FBI is a governmental agency belonging to the United States Department of Justice that serves as both a federal criminal investigative body and an internal intelligence agency. |
| 20 | Public Transportation | 1046 | Public transport is a shared passenger transport service which is available for use by the general public, as distinct from modes such as taxicab, car-pooling or hired buses which are not shared by strangers without private arrangement. |

Table D.1: Twenty text corpuses used for the case study

## Appendix E: The Used Tools and Frameworks

The following below tools and techniques are used in this project and dissertation:

- Netbeans IDE
- Eclipse
- The Concept Explorer
- Notepad++
- WordNet
- Netbeans IDE JUnit testing
- Microsoft Visio and Draw.io
- Stanford Natural Language Processing framework

# Appendix F: Additional Results

In this appendix, the complete results with graphs that are mentioned in. Particularly, the next section presents a table of the evaluation results of applying twenty free texts by using WordNet-based technique and/or Frequency-based technique. This evaluation result includes the statistical data of the reduced lattices and the original one, and then standard deviations are calculated for the statistical data of the concept lattices. Section F.2 presents graphs of evaluation results.

## F.1 Evaluation Results

This section presents the evaluation results in Table F1.1.

| text | 1.Original Text - Lattice | | | | 2.WordNet-based – Lattice | | | | 3.Frequency-based Lattice | | | | 4.WordNet and Frequency Lattice | | | | 5.Frequency and WordNet Lattice | | | |
|---|---|---|---|---|---|---|---|---|---|---|---|---|---|---|---|---|---|---|---|---|
| | Concept | Edge | Height | Width | Concept | Edge | Height | Width | Concept | Edge | Height | Width | Concept | Edge | Height | Width | Concept | Edge | Height | Width |
| 1 | 47 | 86 | 3 | [39,44] | 38 | 65 | 11 | [5,27] | 17 | 28 | 3 | [11,14] | 38 | 65 | 11 | [5,27] | 15 | 23 | 5 | [4,10] |
| 2 | 55 | 101 | 4 | [35,51] | 50 | 95 | 11 | [7,39] | 31 | 57 | 4 | [18,27] | 50 | 95 | 11 | [7,39] | 47 | 99 | 8 | [8,39] |
| 3 | 21 | 35 | 5 | [7,16] | 21 | 35 | 5 | [7,16] | 35 | 64 | 4 | [27,31] | 21 | 35 | 5 | [7,16] | 21 | 35 | 5 | [7,16] |
| 4 | 31 | 56 | 3 | [26,28] | 34 | 68 | 6 | [10,28] | 31 | 56 | 3 | [26,28] | 34 | 68 | 6 | [10,28] | 34 | 68 | 6 | [10,28] |
| 5 | 40 | 71 | 4 | [27,36] | 32 | 54 | 10 | [5,22] | 40 | 71 | 4 | [27,36] | 32 | 54 | 10 | [5,22] | 32 | 54 | 10 | [5,22] |
| 6 | 44 | 81 | 4 | [35,40] | 38 | 68 | 10 | [7,28] | 23 | 40 | 4 | [16,19] | 38 | 68 | 10 | [7,28] | 21 | 36 | 7 | [7,14] |
| 7 | 51 | 91 | 3 | [39,48] | 59 | 114 | 11 | [9,48] | 21 | 34 | 3 | [15,18] | 59 | 114 | 11 | [9,48] | 26 | 48 | 6 | [8,20] |
| 8 | 59 | 106 | 3 | [41,56] | 34 | 62 | 11 | [7,23] | 36 | 63 | 3 | [25,33] | 34 | 62 | 11 | [7,33] | 24 | 40 | 8 | [4,16] |
| 9 | 48 | 86 | 3 | [38,45] | 18 | 26 | 8 | [4,10] | 13 | 21 | 3 | [8,10] | 18 | 26 | 8 | [4,10] | 17 | 27 | 6 | [4,11] |
| 10 | 54 | 99 | 4 | [36,50] | 48 | 96 | 10 | [9,38] | 32 | 57 | 4 | [18,28] | 48 | 96 | 10 | [9,28] | 28 | 48 | 9 | [6,19] |
| 11 | 58 | 110 | 3 | [43,55] | 33 | 55 | 12 | [4,21] | 32 | 59 | 3 | [22,29] | 33 | 55 | 12 | [4,21] | 23 | 40 | 7 | [6,16] |
| 12 | 22 | 37 | 3 | [17,19] | 12 | 16 | 6 | [3,6] | 22 | 37 | 3 | [17,19] | 12 | 16 | 6 | [3,6] | 12 | 16 | 6 | [3,6] |
| 13 | 50 | 94 | 3 | [38,47] | 70 | 146 | 11 | [11,59] | 26 | 46 | 3 | [17,23] | 70 | 146 | 11 | [11,59] | 30 | 52 | 9 | [5,21] |
| 14 | 36 | 63 | 4 | [25,32] | 29 | 47 | 11 | [5,18] | 36 | 63 | 4 | [25,32] | 29 | 47 | 11 | [5,18] | 29 | 47 | 11 | [5,18] |
| 15 | 24 | 43 | 3 | [20,21] | 17 | 26 | 6 | [4,11] | 24 | 43 | 3 | [20,21] | 17 | 26 | 6 | [4,11] | 17 | 26 | 6 | [4,11] |
| 16 | 33 | 63 | 4 | [23,29] | 21 | 32 | 8 | [4,13] | 33 | 63 | 4 | [23,29] | 21 | 32 | 8 | [4,13] | 21 | 32 | 8 | [4,13] |
| 17 | 45 | 83 | 3 | [35,42] | 20 | 31 | 8 | [4,12] | 45 | 83 | 3 | [35,42] | 20 | 31 | 8 | [4,12] | 20 | 31 | 8 | [4,12] |
| 18 | 49 | 92 | 4 | [34,45] | 79 | 185 | 12 | [13,65] | 25 | 46 | 4 | [14,21] | 79 | 185 | 12 | [13,67] | 28 | 52 | 7 | [6,21] |
| 19 | 32 | 58 | 3 | [27,29] | 26 | 44 | 8 | [5,18] | 32 | 58 | 3 | [27,29] | 26 | 44 | 8 | [5,18] | 26 | 44 | 8 | [5,18] |
| 20 | 65 | 131 | 5 | [38,60] | 41 | 76 | 11 | [6,30] | 50 | 101 | 5 | [26,45] | 41 | 76 | 11 | [6,30] | 51 | 106 | 10 | [9,41] |
| Mean | 43.2 | 79.3 | 3.55 | [31,40] | 36 | 67.05 | 9.3 | [8,27] | 30.2 | 54.5 | 3.5 | [21,27] | 36 | 67.05 | 9.3 | [6,27] | 26.1 | 46.2 | 7.5 | [6,19] |
| Median | 46 | 84.5 | 3 | [35,43] | 33.5 | 58.5 | 10 | [7,23] | 31.5 | 57 | 3 | [7,23] | 33.5 | 58.5 | 10 | [6,25] | 25 | 42 | 7.5 | [5,17] |
| Sum | 864 | 1586 | 71 | [623,793] | 720 | 1341 | 186 | [164,535] | 604 | 1090 | 70 | [417,534] | 720 | 1341 | 186 | [129,535] | 522 | 924 | 150 | [114,372] |
| Max | 65 | 131 | 5 | [43,60] | 79 | 185 | 12 | [28,65] | 50 | 101 | 5 | [35,45] | 79 | 185 | 12 | [13,67] | 51 | 106 | 11 | [10,41] |
| Min | 21 | 35 | 3 | [7,16] | 12 | 16 | 5 | [3,6] | 13 | 21 | 3 | [8,10] | 12 | 16 | 5 | [3,6] | 12 | 16 | 5 | [3,6] |
| STDV P | 12.51 | 24.97 | 0.66 | [9,13] | 17.4 | 41.74 | 2.17 | [6,16] | 8.86 | 18.21 | 2.17 | [6,9] | 17.4 | 41.74 | 2.17 | [3,16] | 9.49 | 22.29 | 1.65 | [2,9] |
| STDV S | 12.84 | 26.47 | 0.7 | [9,13] | 17.85 | 44.74 | 2.27 | [6,16] | 9.09 | 18.62 | 0.61 | [7,9] | 17.85 | 44.74 | 2.27 | [3,16] | 9.73 | 19.65 | 1.68 | [2,9] |
| STDEVA | 12.84 | 25.62 | 0.068 | [9,13] | 17.85 | 42.83 | 2.22 | [6,16] | 9.09 | 18.69 | 0.6 | [7,9] | 17.85 | 42.83 | 2.22 | [3,16] | 9.73 | 22.87 | 1.7 | [2,9] |

Table F.1: Evaluation results

## F.2 Graphs of Evaluation Results

This section presents some graphs of evaluation results.

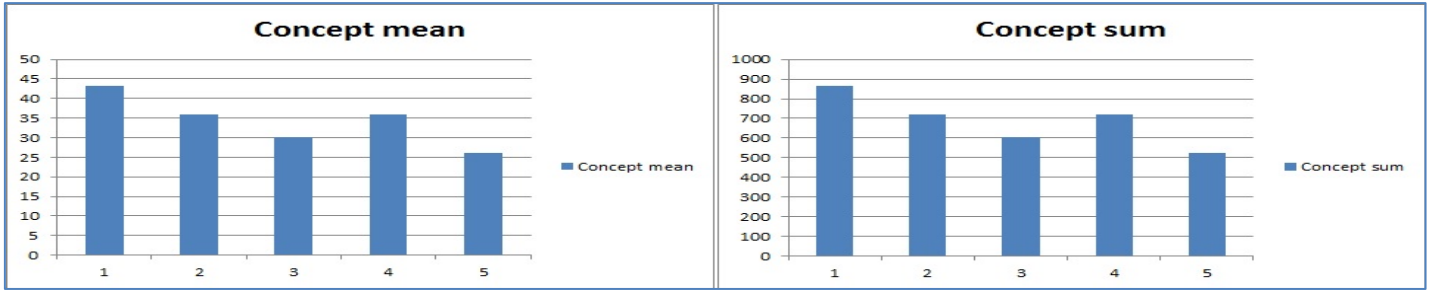

Figure F.2: Mean of concept count (left) and sum of concept count (right)

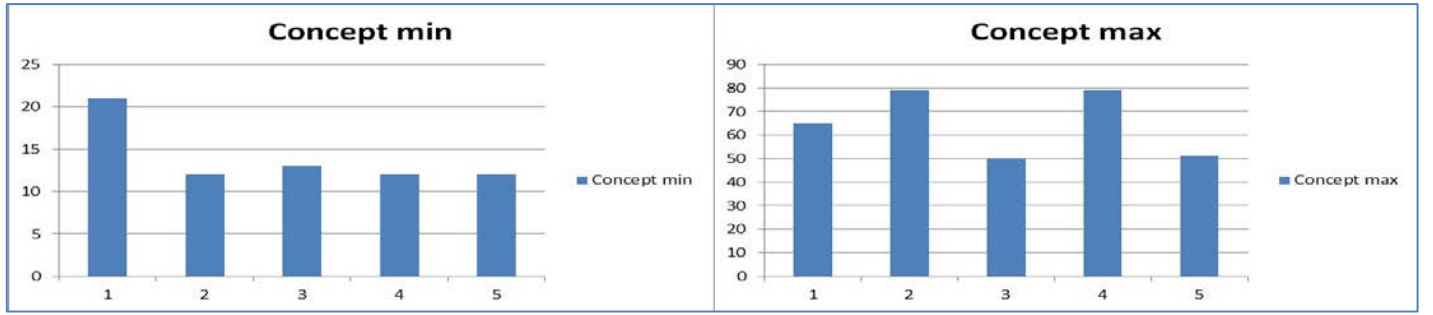

Table F.3: Min of concept count (left) and max of concept count (right)

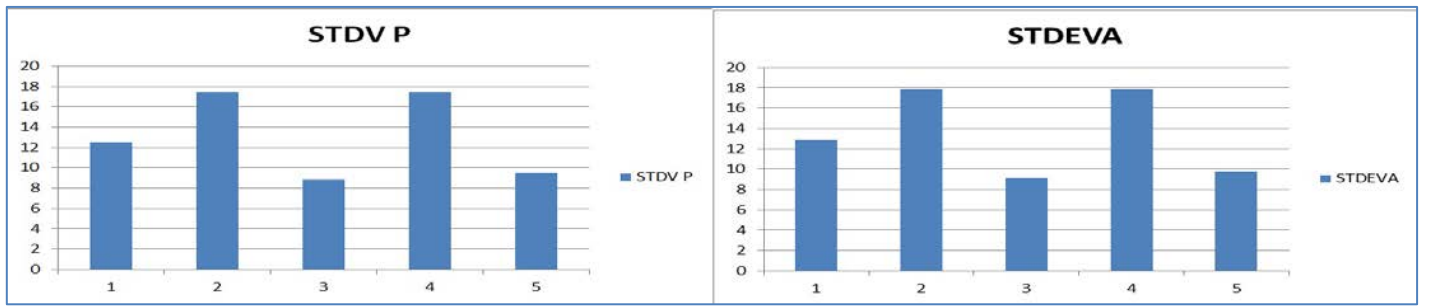

Table F.4: STDV P of concept count (left) and STDEVA of concept count (right)

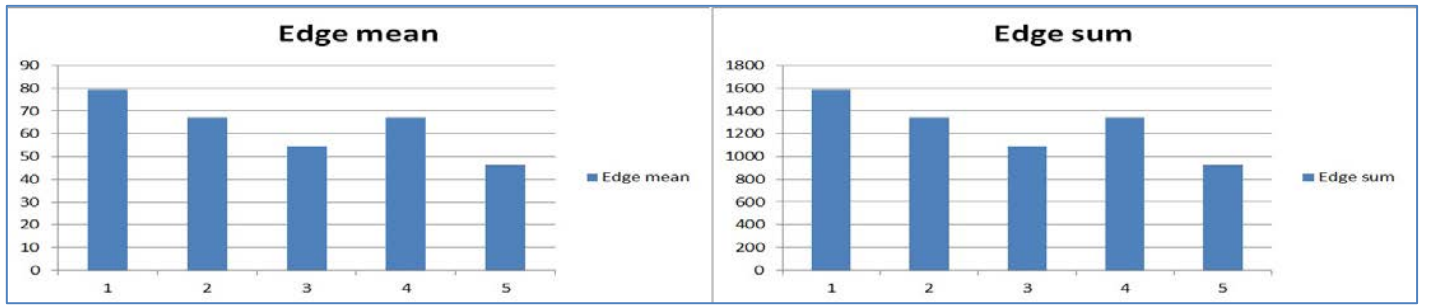

Table F.5: Mean of edge count (left) and sum of edge count (right)

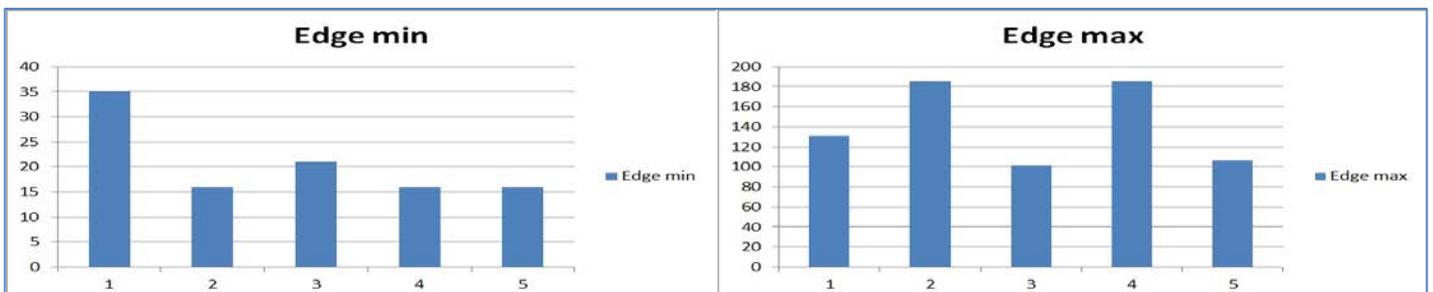

Table F.6: Min of edge count (left) and max of edge count (right)

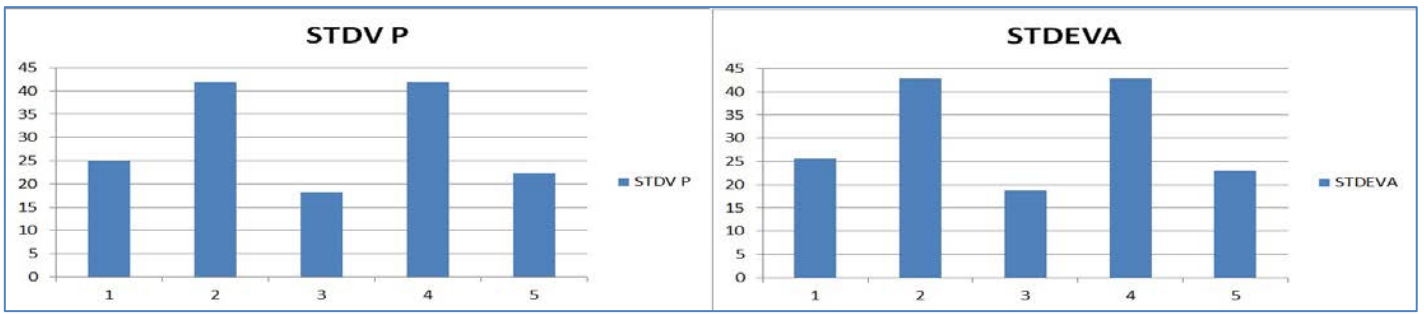
Table F.7: STD P of edge count (left) and STDEVA of edge count (right)

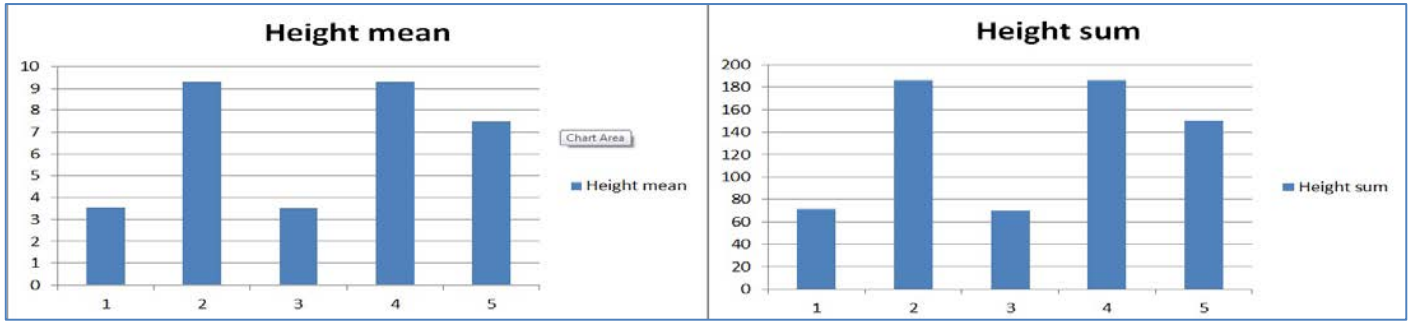
Table F.8: Mean of lattice height (left) and max of lattice height (right)

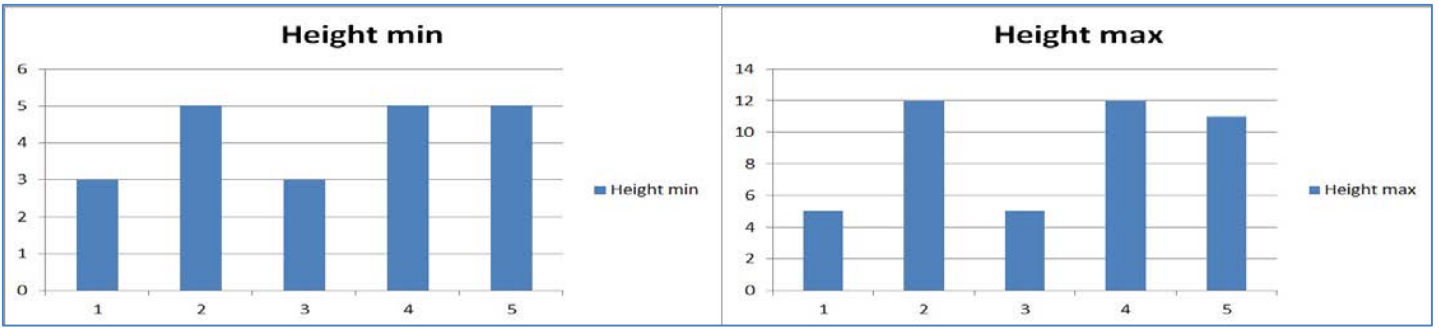
Table F.9: Min of lattice height (left) and max of lattice height (right)

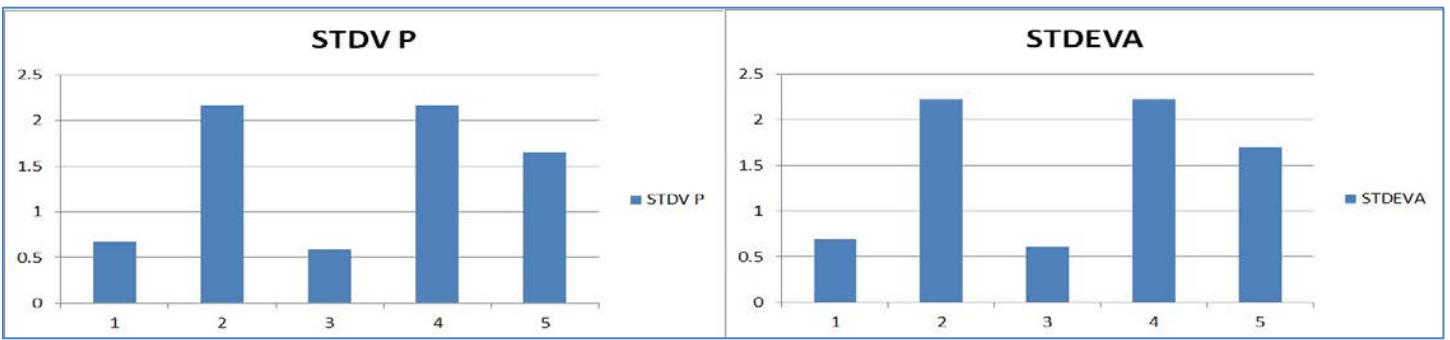
Table F.10: STDV P of lattice height (left) and STDEVA of lattice height (right)

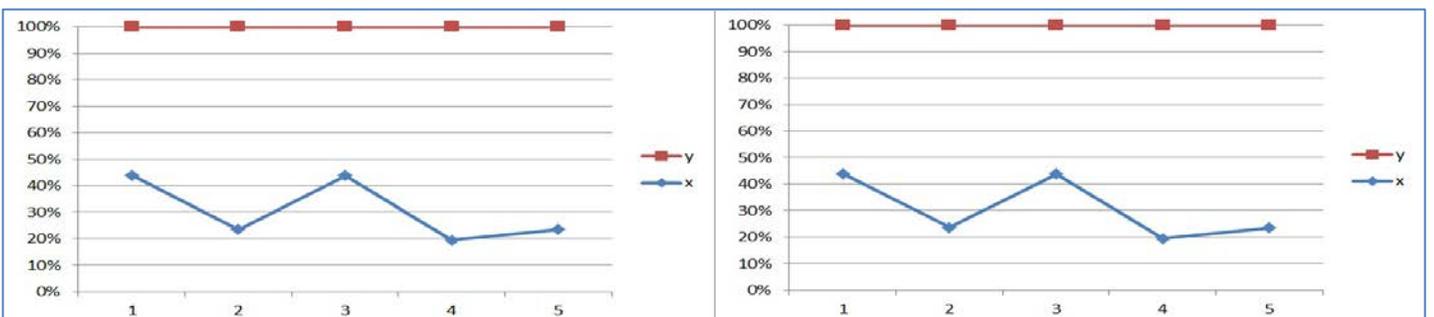
Table F.11: Mean of lattice width (left) and sum of lattice width (right)

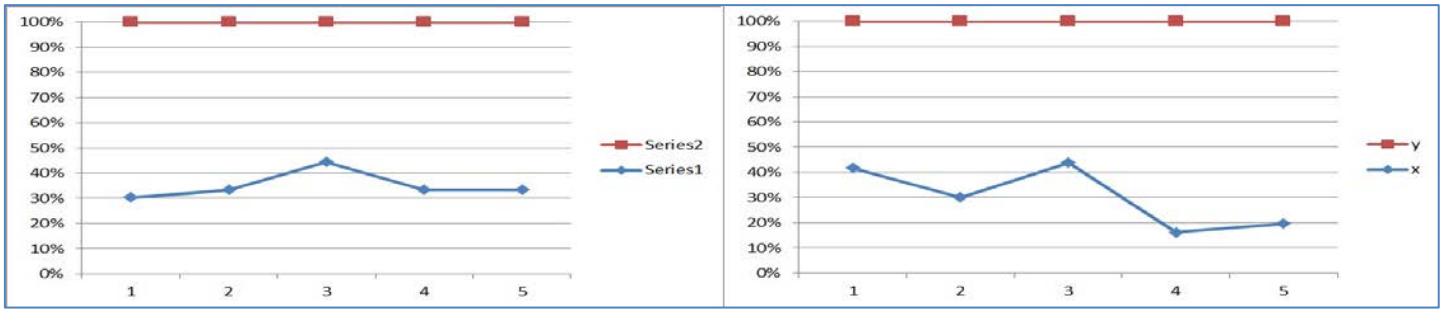
Table F.12: Min of lattice width (left) and max of lattice width (right)

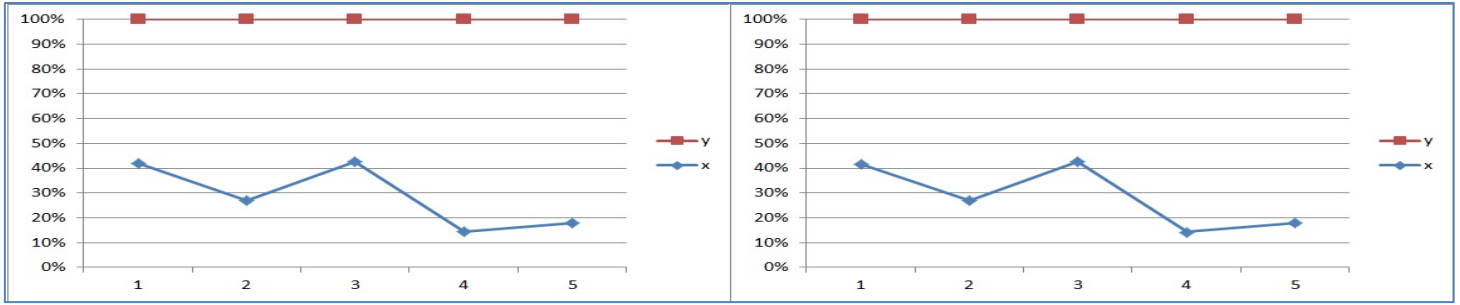
Table F.13: STDV P of lattice width (left) and STDEVA of lattice width (right)

**Appendix G: Bing Bang Integrated Testing**

By using Big Bang integrated testing approach, the input and expected output of the module of this project and dissertation are outlined in Table G.1.

| Modules | Input | Expected output |
|---|---|---|
| CEXReader | A sample XML file is used to read | The XML is read successfully |
| CEXWriter | A sample formatted data is used to write on an XML file | The formatted data is written into the XML file successfully |
| WordNetBasedTech | A formal context is given as input | The formal context is reduced by applying this technique successfully |
| FreqBasedTech | A formal context is given as input | The formal context is reduced by applying this technique successfully |
| WordNetHelper | Some words are given to test its methods | The output of the words are expected and worked successfully |
| SynReplace | Some words are given to test its methods | The output of the words are expected and worked successfully |
| TwoDimentionalArrayList | A two dimensional arrayList is used to test. | The two dimensional arrayList is tested and worked successfully |
| TwoDimentionalArrayListUtil | A two dimensional array is used to test and manipulated with it. | The two dimensional array is worked and manipulated with it successfully |
| Lemmatiser | Some un-lemmatised words are given to lemmatise | The words are lemmatised successfully |
| FrameworkComponents | All of the module is called here to work together to form the framework | The framework is formed and worked successfully |
| Stemmer | Some un-stemmed words are given to stem | The words are not stemmed successfully. For example, "beautiful" is stemmed as "beauty" |
| WordPair | A sample TXT file is given to read it to construct formal context | The file is read, and a formal context is constructed |

Table G.1: Big Bang integrated testing inputs and expected outputs